%% file: 4061.tex
\documentclass[runningheads]{llncs}
\usepackage{graphicx}
\usepackage{comment}
\usepackage{multirow}
\usepackage{amssymb}
\usepackage{subfigure}
\usepackage{epsfig}
\usepackage{amsmath,amssymb} 
\usepackage{color}

\usepackage{array}
\graphicspath{{figures/}}
\input{macrosjvg}

\begin{document}
\pagestyle{headings}
\mainmatter
\def\ECCVSubNumber{4061}  

\title{Deep Hough-Transform Line Priors} 

\titlerunning{Deep Hough-Transform Line Priors}
%
\author{Yancong Lin \and Silvia L. Pintea \and Jan C. van Gemert}
\authorrunning{Y. Lin, S.L. Pintea, and J.C. van Gemert}
%
\institute{Computer Vision Lab\\
Delft University of Technology, the Netherlands \\
}

\maketitle
\input{abstract}
\input{intro}
\input{related}
\input{method}

\input{experiments}
\input{conclusion}
%
%
{
    \small
    \bibliographystyle{splncs04}
    \bibliography{ref}{}
}

\newpage
\input{appendix}

\end{document}

%% file: macrosjvg.tex
\usepackage{xcolor}

\usepackage{wrapfig}

\usepackage{graphicx}
\usepackage{amsmath}
\usepackage{amssymb}
\usepackage{booktabs}
\usepackage{comment}
\usepackage{hyperref}

\usepackage{xspace}

\newcommand{\model}{HT-IHT block\xspace}

%% file: abstract.tex
\begin{abstract}
Classical work on line segment detection  is knowledge-based; it uses carefully designed geometric priors using either image gradients, pixel groupings, or Hough transform variants.
Instead, current deep learning methods do away with all prior knowledge and replace priors by training deep networks on large manually annotated  datasets.
Here, we reduce the dependency on labeled data by building on the classic knowledge-based priors while using deep networks to learn features. 
We add line priors through a trainable Hough transform block into a deep network. 
Hough transform provides the prior knowledge about global line parameterizations, while the convolutional layers can learn the local gradient-like line features. 
On the  Wireframe (ShanghaiTech) and York Urban datasets we show that adding prior knowledge improves data efficiency as line priors no longer need to be learned from data.
\keywords{Hough transform; global line prior, line segment detection.}
\end{abstract}

%% file: intro.tex
\section{Introduction}
\begin{figure}[t]
    \centering
    \begin{tabular}{cccc}
        \includegraphics[width=0.23\textwidth]{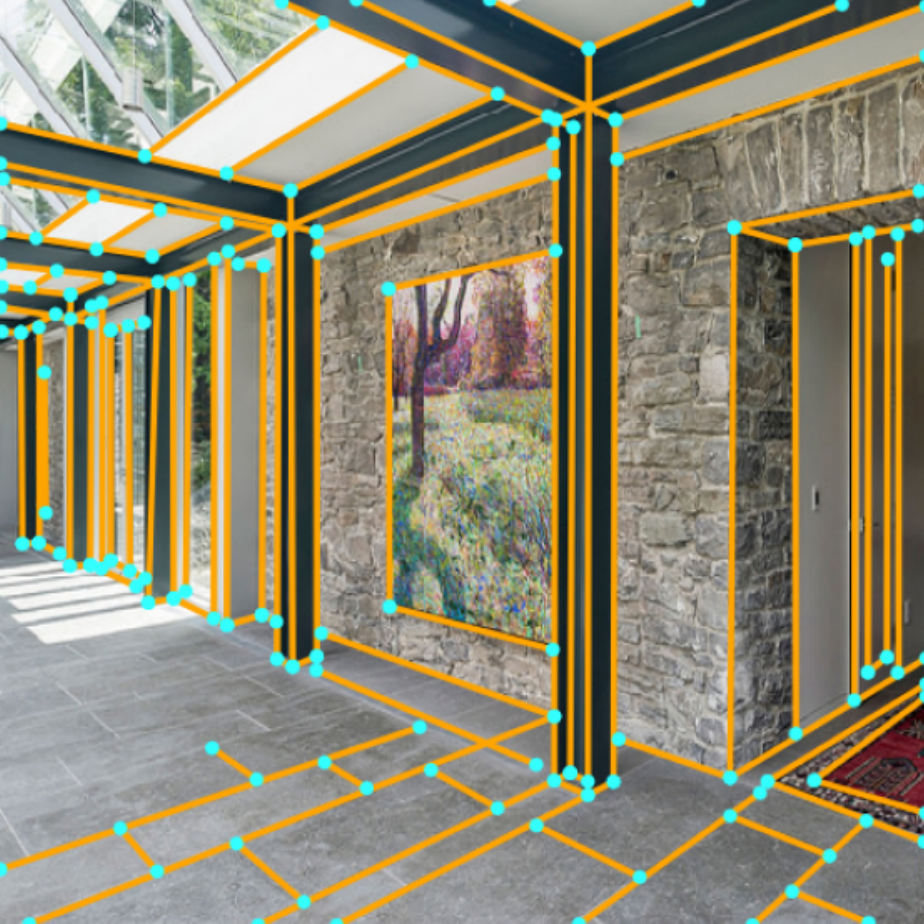} &
        \includegraphics[width=0.23\textwidth]{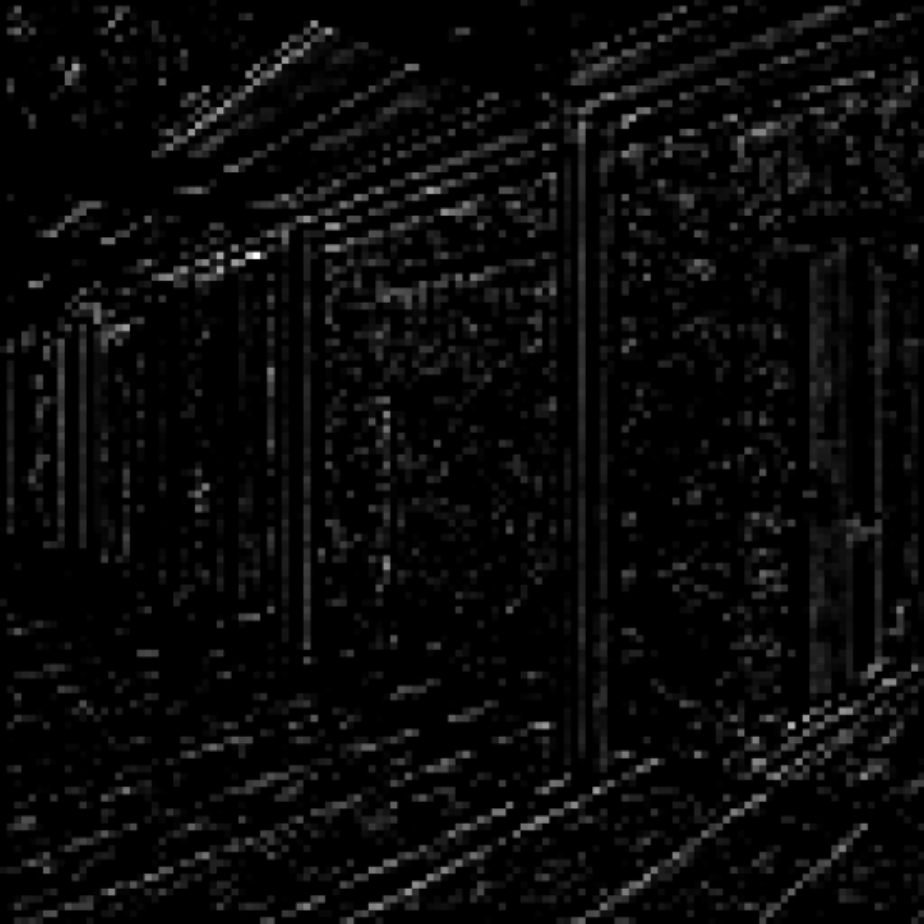} &
        \includegraphics[width=0.23\textwidth]{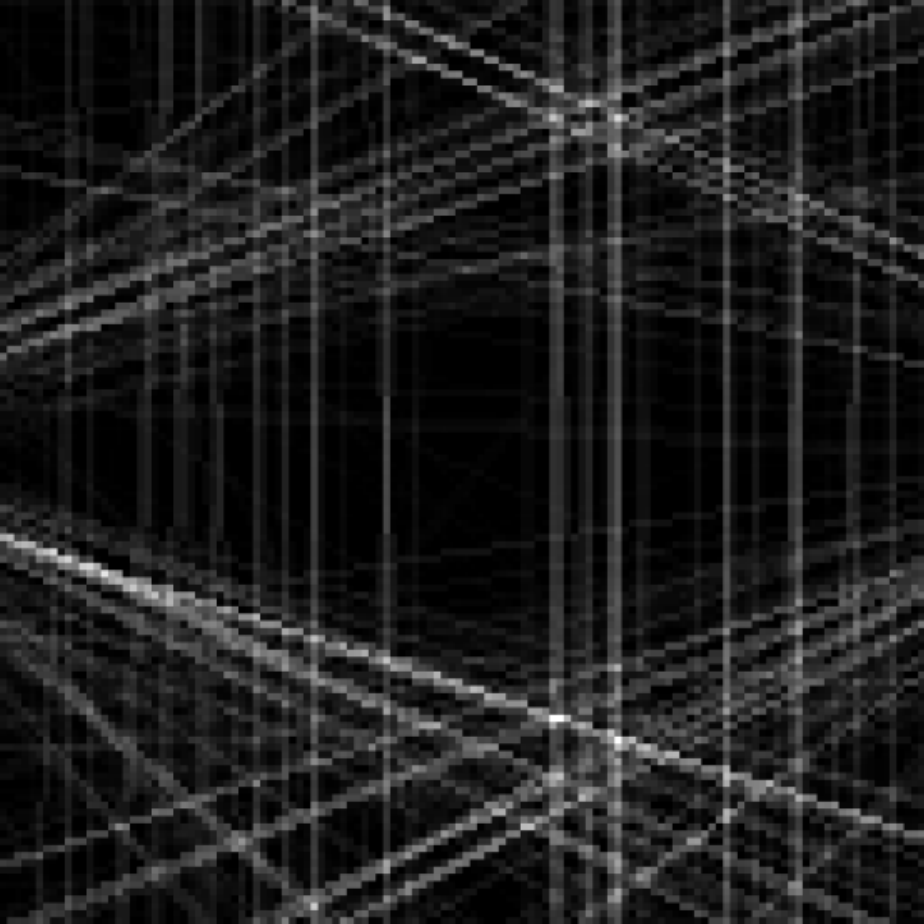} & 
        \includegraphics[width=0.23\textwidth]{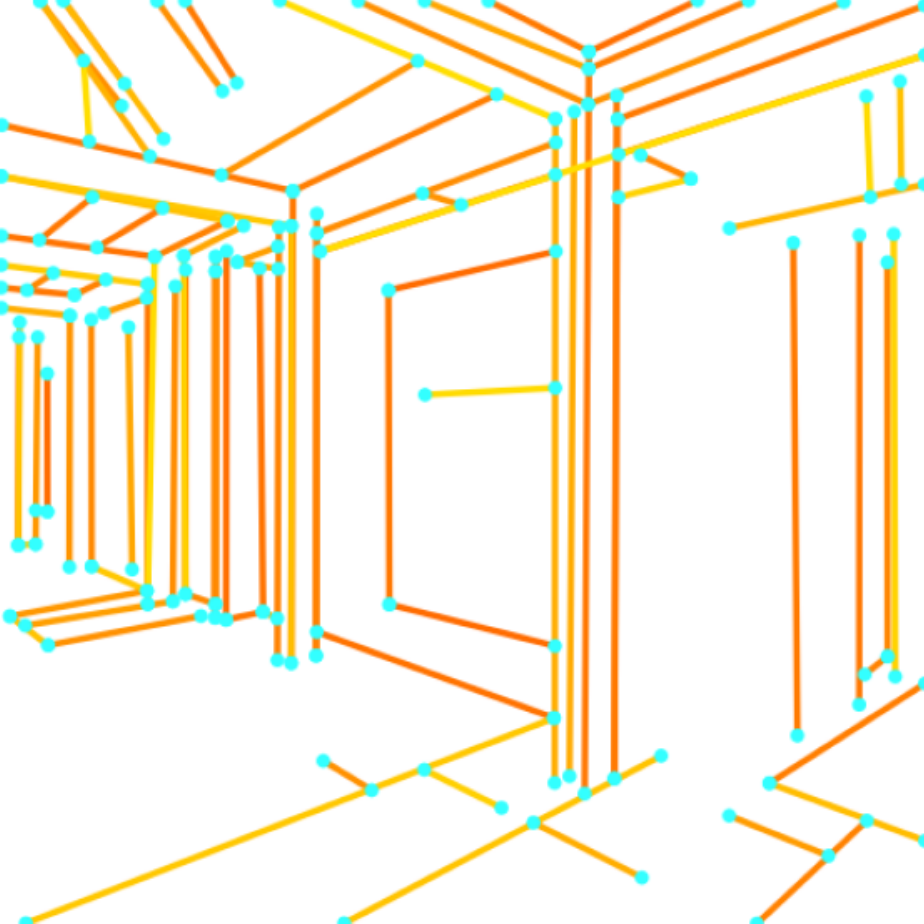} \\
        \footnotesize{Ground truth} & \footnotesize{Learned local features} & \footnotesize{Added line priors} & \footnotesize{Line predictions} \\
    \end{tabular}
    \caption{We add prior knowledge to deep networks for data efficient line detection. 
    We learn local deep features, which are combined with a global inductive line priors, using the Hough transform. Adding prior knowledge saves valuable training data.}
    \label{fig:global}
\end{figure}

Line segment detection is a classic Computer Vision task, with applications
such as road-line detection for autonomous driving \cite{hillel2014recent,lee2017vpgnet,niu2016robust,satzoda2014efficient}, wireframe detection for design in architecture \cite{huang2018learning,zhou2019end,zhou2019learning}, horizon line detection for assisted flying \cite{gershikov2013horizon,porzi2016deeply,simon2018contrario}, image vectorization \cite{sun2007image,zou2001cartoon}.
Such problems are currently solved by state-of-the-art line detection methods \cite{huang2018learning,zhou2019end,xue2019learning} by relying on deep learning models powered by huge, annotated, datasets.

Training deep networks demands large datasets \cite{barbu2019objectnet,ILSVRC15}, which are expensive to annotate. 
The amount of needed training data can be significantly reduced by adding prior knowledge to deep networks  \cite{bruna2013invariant,jacobsen2016structured,kayhan2020translation}.
Priors encode inductive solution biases: e.g. for image classification, objects can appear at any location and size in the input image. 
The convolution operation adds a translation-equivariance prior \cite{kayhan2020translation,urban2016deep}, and multi-scale filters add a scale-invariance prior \cite{shelhamer2019blurring,sosnovik2019scale}. Such priors offer a strong reduction in the amount of required data: built-in knowledge no longer has to be learned from data. Here, we study straight line detection which allows us to exploit the line equation. 

In this work we add geometric line priors into deep networks for improved data efficiency by relying on the Hough transform. 
The Hough transform has a long and successful history for line detection \cite{duda1972use,kamat1998complete,matas2000robust}.
It parameterizes lines in terms of two  geometric terms: an offset and an angle, describing the line equation in polar coordinates. 
This gives a global representation for every line in the image. 
As shown in figure~\ref{fig:global}, global information is essential to correctly locate lines, when the initial detections are noisy. In this work we do not exclusively rely on prior knowledge as in the classical approach~\cite{burns1986extracting,cho2017novel,puatruaucean2012parameterless,von2008lsd} nor do we learn everything in deep architectures~\cite{huang2018learning,xue2019learning,zhou2019end}. Instead, we take the best of both: we combine learned global shape priors with local learned appearance.

This paper makes the following contributions:
(1) we add global geometric line priors through Hough transform into deep networks; 
(2) we improve data efficiency of deep line detection models;  
(3) we propose a well-founded manner of adding the Hough transform into an end-to-end trainable deep network, with convolutions performed in the Hough domain over the space of all possible image-line parameterizations; 
(4) we experimentally show improved data efficiency and a reduction in parameters on two popular line segment detection datasets, Wireframe (ShanghaiTech) \cite{huang2018learning} and York Urban \cite{denis2008efficient}.

%% file: related.tex
\section{Related work}

\textbf{Image Gradients.} Lines are edges, therefore substantial work has focused on line segment detection using local image gradients followed by pixel grouping strategies such a region growing~\cite{puatruaucean2012parameterless,von2008lsd}, connected components~\cite{burns1986extracting},  probabilistic graphical models~\cite{cho2017novel}. Instead of knowledge-based approach for detecting local line features, we use deep networks to learn local appearance-based features, which we combine with a global Hough transform prior.  

\textbf{Hough transform.} The Hough transform is the most popular algorithm for image line detection where the offset-angle line parameterization was first used in 1972~\cite{duda1972use}.
Given its simplicity and effectiveness, subsequent line-detection work followed this approach~\cite{furukawa2003accurate,kamat1998complete,xu2014accurate}, by focusing on analyzing peaks in Hough space.
To overcome the sensitivity to noise, previous work proposed statistical analysis of Hough space~\cite{xu2015statistical}, and segment-set selection based on hypothesis testing~\cite{von2008straight}.
Similarly, a probabilistic Hough transform for line detection, followed by Markov Chain modelling of candidate lines is proposed in~\cite{almazan2017mcmlsd}, while
\cite{matas2000robust} creates a progressive probabilistic Hough transform, which is both faster and more robust to noise.
An extension of Hough transform with edge orientation is used in~\cite{guerreiro2012connectivity}. 
Though less common, the slope-intercept parameterization of Hough transform for detecting lines is considered in~\cite{sheshkus2019houghnet}. 
In~\cite{nikolaev2008hough} Hough transform is used for detecting page orientation for character recognition. 
In our work, we do not use hand-designed features, but exploit the line prior knowledge given by the Hough transform when included into a deep learning model, allowing it to behave as a global line-pooling unit.  

\textbf{Deep learning for line detection}
The deep network in~\cite{huang2018learning} uses two heads: one for junction prediction and one for line detection. This is extended in~\cite{zhou2019end}, by a line-proposal sub-network.
A segmentation-network backbone combined with an attraction field map, where pixels vote for their closest line is used in \cite{xue2019learning}.
Similarly, attraction field maps are also used in \cite{xue2020holistically} for generating line proposals in a deep architecture.  Applications of line prediction using a deep network include aircraft detection~\cite{wei2019x}, and  power-line detection~\cite{nguyen2019ls}. Moving from 2D to 3D,~\cite{zhou2019learning} predicts 3D wireframes from a single image by relying on the assumption that image scenes have an underlying Cartesian grid.
Another variation of the wireframe-prediction task is proposed in~\cite{xue2019learning} which creates a fisheye-distorted wireframe dataset and proposes a method to rectify it. A graph formulation~\cite{zhang2019ppgnet} can learn the association between end-points. 
The need for geometric priors for horizon line detection is investigated in~\cite{workman2016horizon}, concluding that CNNs (Convolutional Neural Networks) can learn without explicit geometric information.  
However, as the availability of labeled data is a bottleneck, we argue that prior geometric information offers improved data efficiency.

\textbf{Hough transform hybrids}
Using a vote accumulator for detecting image structure is used in~\cite{beltrametti2019geometry} for curve detection.
Deep Hough voting schemes are considered in~\cite{qi2019deep} for detecting object centroids on 3D point clouds, and for finding image correspondences~\cite{min2019hyperpixel}.
In our work, we also propose a Hough-inspired block that accumulates line votes from input featuremaps. The Radon transform is a continuous version of the Hough~transform~\cite{beatty2012radon,magnusson1993linogram,toft1996radon}.
Inverting the Radon transform back to the image domain is considered in~\cite{He2018RadonIV,rim2020exact}.
In~\cite{rim2020exact} an exact inversion from partial data is used, while \cite{He2018RadonIV} relies on a deep network for the inversion, however the backprojection details are missing.
Related to Radon transform, the ridgelet transform~\cite{do2003finite} maps points to lines, and the Funnel transform detects lines by accumulating votes using the slope-intercept line representation~\cite{wei2019funnel}.
Similar to these works, we take inspiration from the Radon transform and its inversion in defining our Hough transform block.

%% file: method.tex
\section{Hough transform block for global line priors}
\begin{figure}[t!]
    \centering
    \includegraphics[width=1.\textwidth]{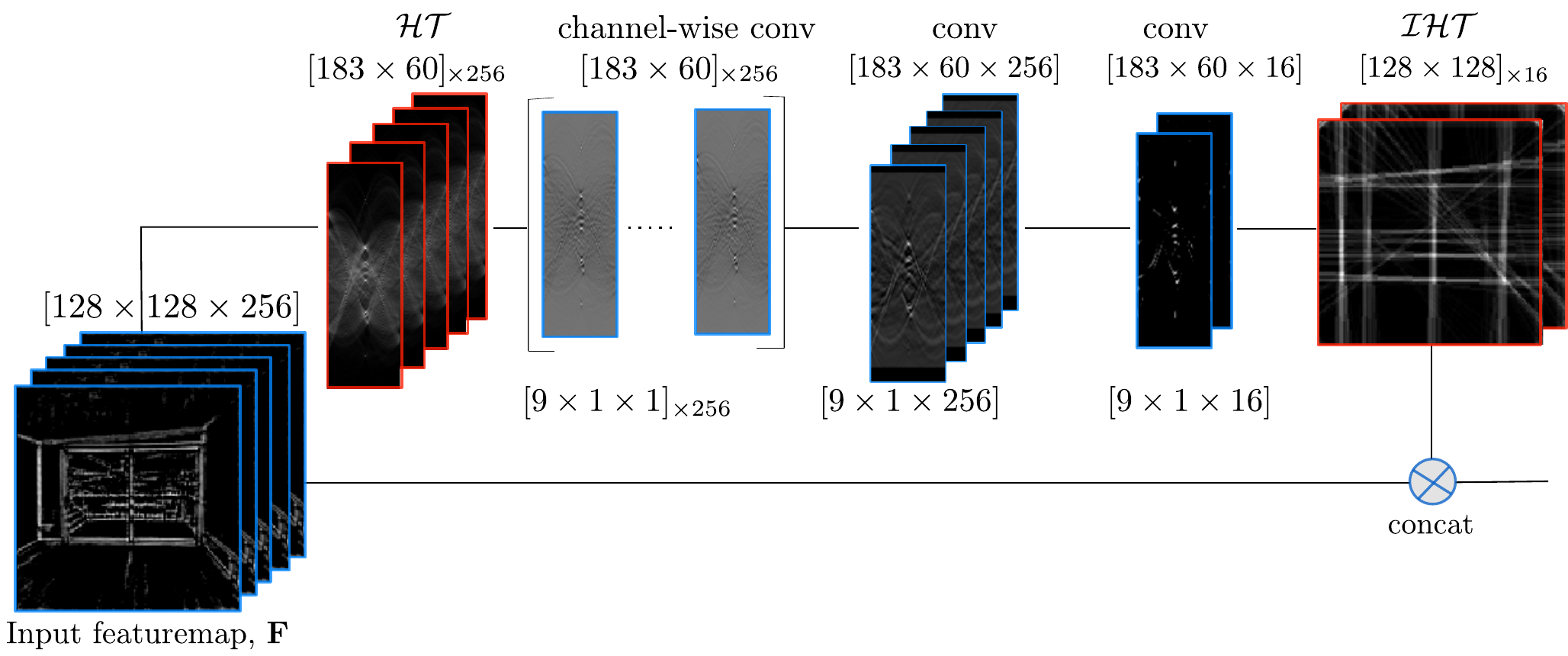}
    \caption{\textbf{\model:} The input featuremap, $\mathbf{F}$, coming from the previous convolutional layer, learns local edge information, and is combined on a residual branch with line candidates, detected in global Hough space.
    The input featuremap of $128\times 128 \times 256$ is transformed channel-wise to the Hough domain through the $\mathcal{HT}$ layer into multiple $183\times 60$ maps. 
    The result is filtered with 1$D$ channel-wise convolutions.
    Two subsequent 1$D$ convolutions are added for merging and reducing the channels.  
    The output is converted back to the image domain by the $\mathcal{IHT}$ layer.
    The two branches are concatenated together.
    Convolutional layers are shown in blue, and in red the $\mathcal{HT}$ and $\mathcal{IHT}$ layers.
    Our proposed \model can be used in any architecture.}
    \label{fig:block}
\end{figure}

Typically, the Hough transform parameterizes lines in polar coordinates as an offset $\rho$ and an angle, $\theta$. These two parameters are discretized in bins. Each pixel in the image votes in all line-parameter bins to which that pixel can belong. 
The binned parameter space is denoted the Hough space and its local extrema correspond to lines in the image. For details, see  figure~\ref{fig:forward}.(a,b) and \cite{duda1972use}.     

We present a Hough transform and inverse Hough transform  (\model) to combine local learned image features  with  global line priors.
We allow the network to combine information by defining the 
Hough transform on a separate residual branch. The $\mathcal{HT}$ layer inside the \model maps input featuremaps to the Hough domain.  
This is followed by a set of local convolutions in the Hough domain which are equivalent to global operations in the image domain. 
The result is then inverted back to the image domain using the $\mathcal{IHT}$ layer, and it is subsequently concatenated with the convolutional branch.
Figure~\ref{fig:block} depicts our proposed \model, which can be used in any architecture.  
To train the \model end-to-end, we must specify its forward and backward definitions. 

\subsection{$\mathcal{HT}$: From image domain to Hough domain}
\begin{figure}[t]
    \centering
    \begin{tabular}{cccc}
    \includegraphics[width=0.24\textwidth]{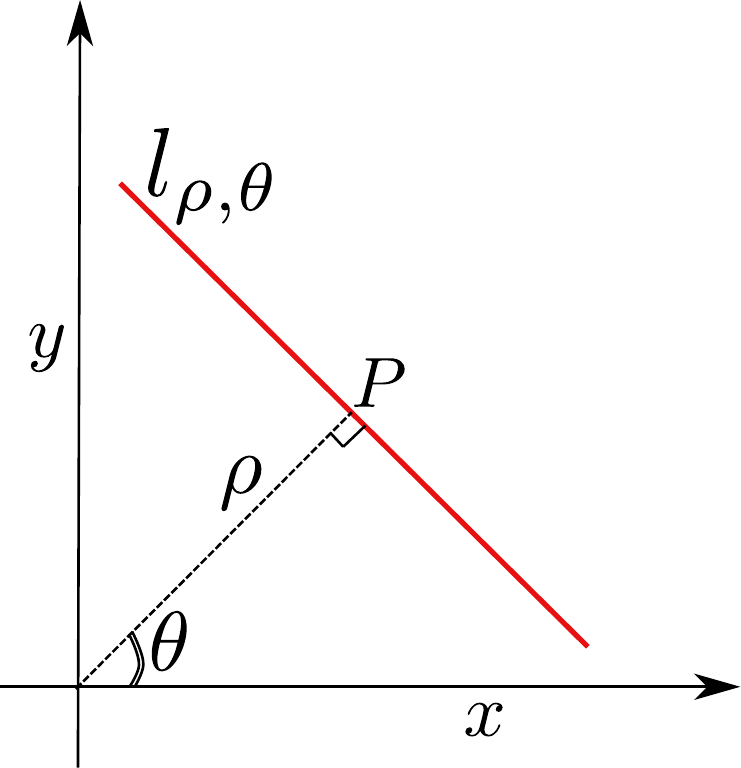} &
    \includegraphics[width=0.24\textwidth]{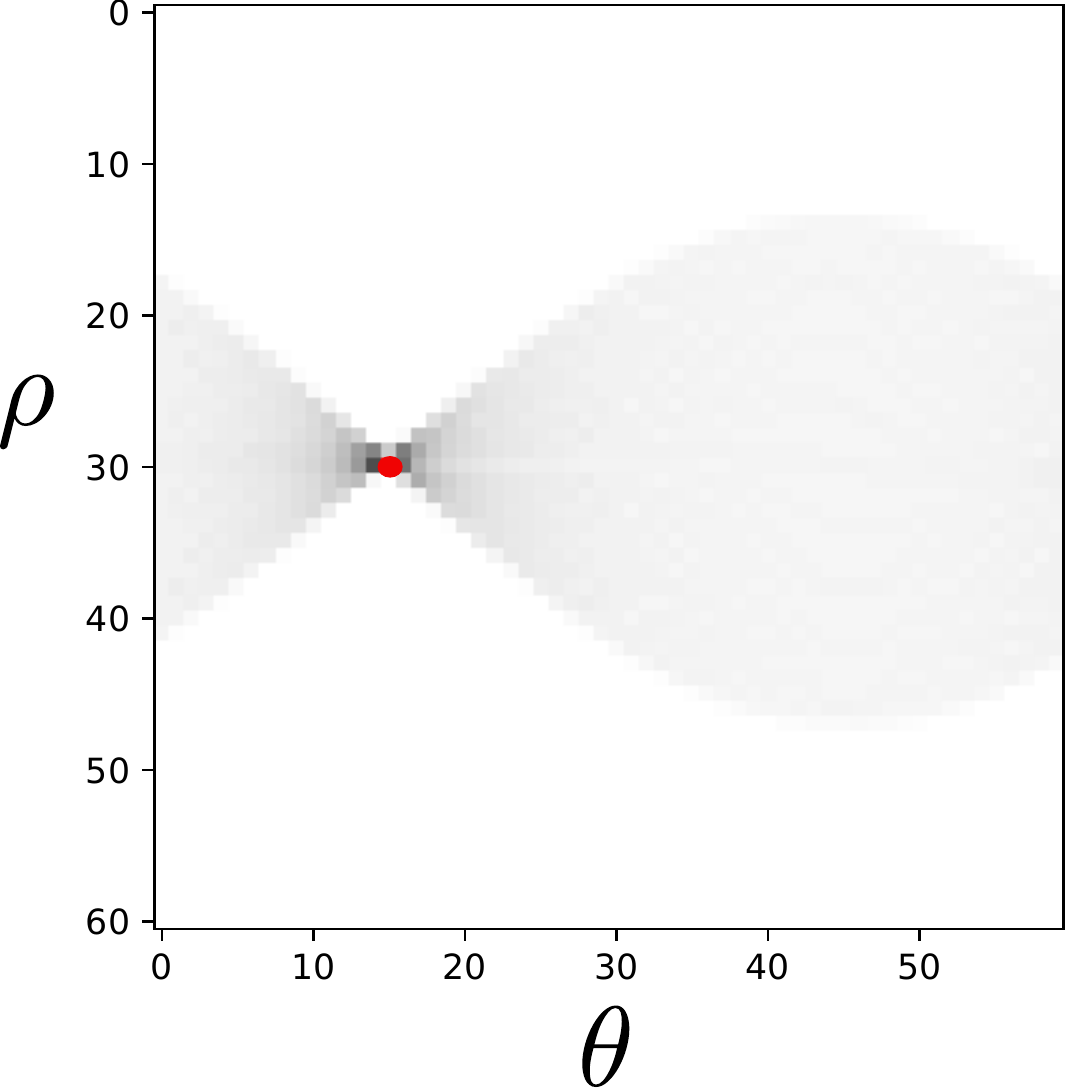} &
    \includegraphics[width=0.24\textwidth]{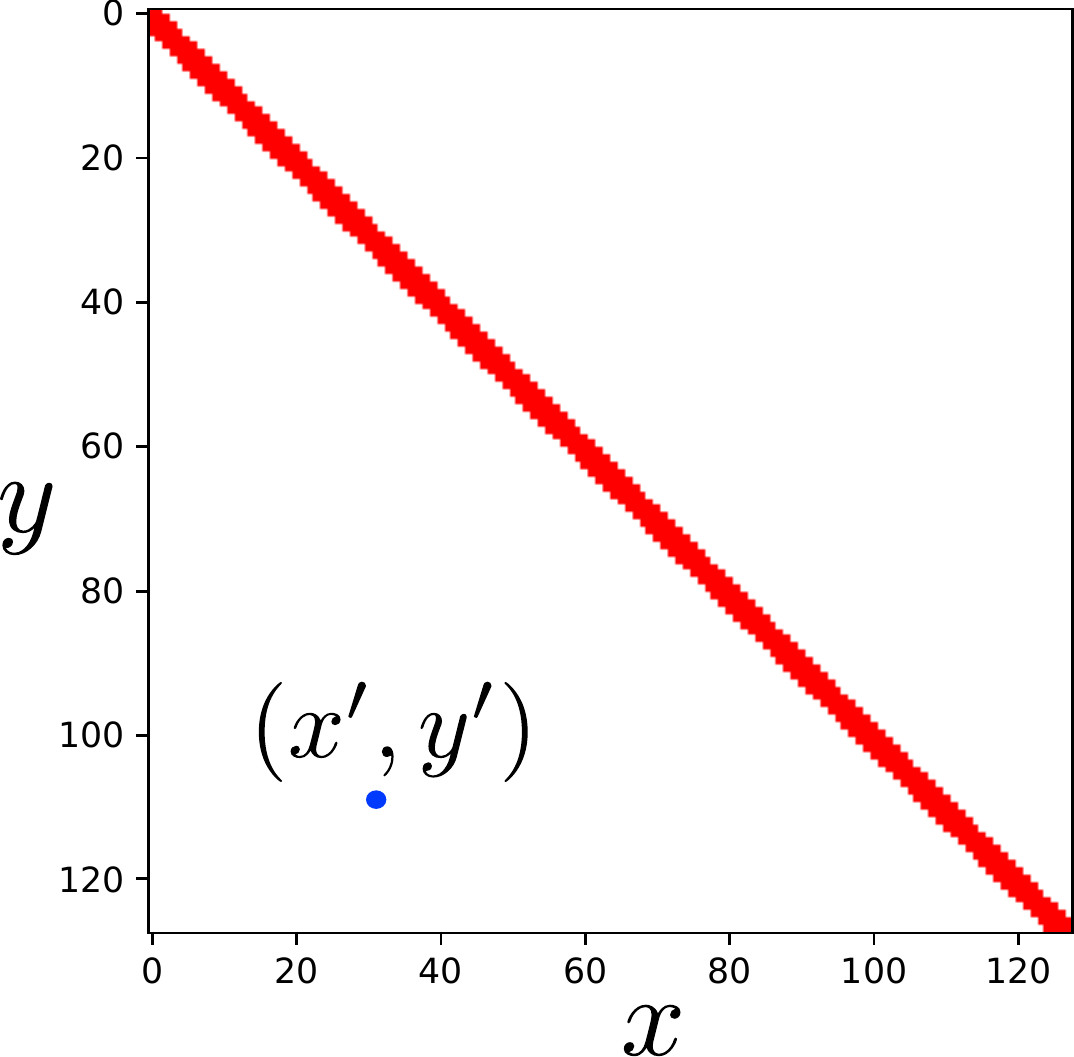} &
    \includegraphics[width=0.24\textwidth]{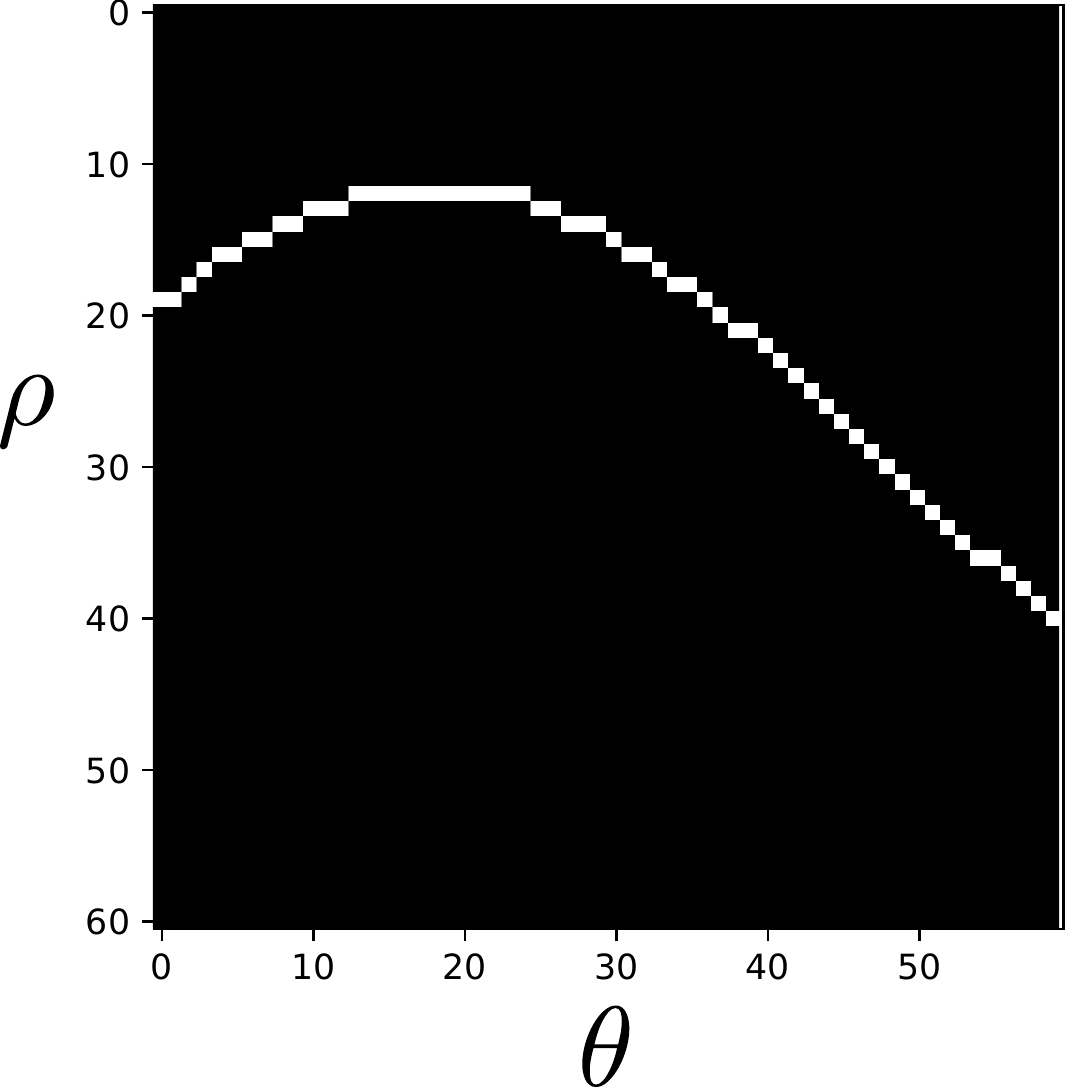} \\
    (a) Input line & (b) Line $\mathcal{HT}$ & (c) Line $\mathcal{IHT}$ & (d) Mask $\mathbf{B}(x^\prime,y^\prime)$\\
    \end{tabular}
    \caption{(a) A line together with its $(\rho,\theta)$ parameterization. 
    (b) The Hough transform ($\mathcal{HT}$) of the line.
    (c) The inverse Hough transform ($\mathcal{IHT}$) of the Hough map.
    (d) The binary mask $\mathbf{B}$, mapping the pixel location $(x^\prime,y^\prime)$ highlighted in blue in (c) to its corresponding set of bins in the Hough domain.}
    \label{fig:forward}
\end{figure}

Given an image line $l_{\rho, \theta}$ in polar coordinates, with an offset $\rho$ and angle $\theta$, as depicted in figure~\ref{fig:forward}.(a), 
for the point $P=(P_x, P_y)$ located at the intersection of the line with its normal, it holds that: $(P_x, P_y) = (\rho \cos \theta, \rho \sin \theta)$. A  point along this line $(x(i), y(i))$ is given by:
\begin{alignat}{1}
    (x(i), y(i)) &= (\rho \cos \theta - i \sin \theta, \rho \sin \theta + i \cos{} \theta),
\label{eq:line}    
\end{alignat}
where $x(\cdot)$ and $y(\cdot)$ define the infinite set of points along the line as functions of the index of the current point, $i$, where $i \in \mathbb{R}$ can take both positive and negative values.  
Since images are discrete, here $(x(i), y(i))$ refers to the pixel indexed by $i$ along an image direction. 

The traditional Hough transform \cite{duda1972use,matas2000robust} uses binary input where featuremaps are real valued. Instead of binarizing the featuremaps, we define the Hough transform similar to the Radon transform~\cite{beatty2012radon}.
Therefore for a certain $(\rho, \theta)$ bin, our Hough transform accumulates the featuremap activations $\mathbf{F}$ of the corresponding pixels residing on that image direction: 
\begin{alignat}{2}
    \mathcal{HT}(\rho,\theta) &= \sum_i \mathbf{F}_{\rho,\theta}(x(i), y(i)),
        \label{eq:ht}
\end{alignat}
where the relation between the pixel $(x(i),y(i))$ and bin $(\rho,\theta)$ is given in equation~(\ref{eq:line}), and $\mathbf{F}_{\rho,\theta}(x(i), y(i))$ is the featuremap value of the pixel indexed by $i$ along the $(\rho, \theta)$ line in the image.
The $\mathcal{HT}$ is computed channel-wise, but for simplicity, we ignore the channel dimension here. 
Figure~\ref{fig:forward}.(b) shows the Hough transform map for the input line in figure~\ref{fig:forward}.(a), where we highlight in red the bin corresponding to the line.

Note that in equation~(\ref{eq:ht}), there is a correspondence between the pixel $(x(i),y(i))$ and the bin $(\rho, \theta)$.
We store this correspondence in a binary matrix, so we do not need to recompute it.
For each featuremap pixel, we remember in which $\mathcal{HT}$ bins it votes, and generate a binary mask $\mathbf{B}$ of size: $[W,H,N_\rho,N_\theta]$ where $[W,H]$ is the size of the input featuremap $\mathbf{F}$, and $[N_\rho,N_\theta]$ is the size of the $\mathcal{HT}$ map. 
Thus, in practice when performing the Hough transform, we multiply the input feature map $\mathbf{F}$ with $\mathbf{B}$, channel-wise: 
\begin{alignat}{1}
    \mathcal{HT} &= \mathbf{F} \mathbf{B}.
\end{alignat}
For gradient stability, we additionally normalize the $\mathcal{HT}$ by the width of the input featuremap.

We transform to the Hough domain for each featuremap channel by looping over all input pixels, $\mathbf{F}$, rather than only the pixels along a certain line, and we consider a range of discrete line parameters, $(\rho, \theta)$ where the pixels can vote. 
The $(\rho, \theta)$ pair is mapped into Hough bins by uniformly sampling $60$ angles in the range $[0,\pi]$ and 183 offsets in the range $[0,d]$, where $d$ is the image diagonal, and the computed offsets from $\theta$ are assigned to the closest sampled offset values. 

\subsection{$\mathcal{IHT}$: From Hough domain to image domain}

The $\mathcal{HT}$ layer has no learnable parameters, and therefore the gradient is simply a mapping from Hough bins to pixel locations in the input featuremap, $\mathbf{F}$. Following \cite{beatty2012radon}, we define the $\mathcal{IHT}$ at pixel location $(x,y)$ as the average of all the bins in $\mathcal{HT}$ where the pixel has voted:
\begin{alignat}{1}
    \mathcal{IHT}(x,y) &= \frac{1}{N_\theta}\sum_\theta \mathcal{HT}(x \cos \theta + y \sin \theta, \theta).
    \label{eq:iht}
\end{alignat} 
In the backward pass, $\frac{\partial \mathcal{HT}}{\partial F(x,y)}$, we use equation~(\ref{eq:iht}) without the normalization over the number of angles, $N_\theta$. 

Similar to the forward Hough transform pass, we store the correspondence between the pixels in the input featuremap $(x,y)$ and the Hough transform bins $(\rho, \theta)$, in the binary matrix, $\mathbf{B}$. 
We implement the inverse Hough transform as a matrix multiplication of $\mathbf{B}$ with the learned $\mathcal{HT}$ map, for each channel:
\begin{alignat}{1}
    \mathcal{IHT} &= \mathbf{B} \left( \frac{1}{N_\theta} \mathcal{HT} \right).
\end{alignat}
Figure~\ref{fig:forward}.(c) shows the $\mathcal{IHT}$ of the Hough transform map 
in figure~\ref{fig:forward}.(b), while figure~\ref{fig:forward}.(d) shows the binary mask $\mathbf{B}$ for the pixel $(x^\prime,y^\prime)$ highlighted in blue in figure~\ref{fig:forward}.(c), mapping it to its corresponding set of bins in the Hough map. 
\subsection{Convolution in Hough Transform space}
\label{section:Convolution in Hough Transform space}

\begin{figure}[t]
    \centering
    \begin{tabular}{cccc}
        \includegraphics[width=0.24\textwidth]{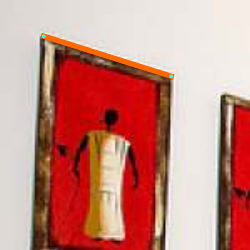} &
        \includegraphics[width=0.10\textwidth]{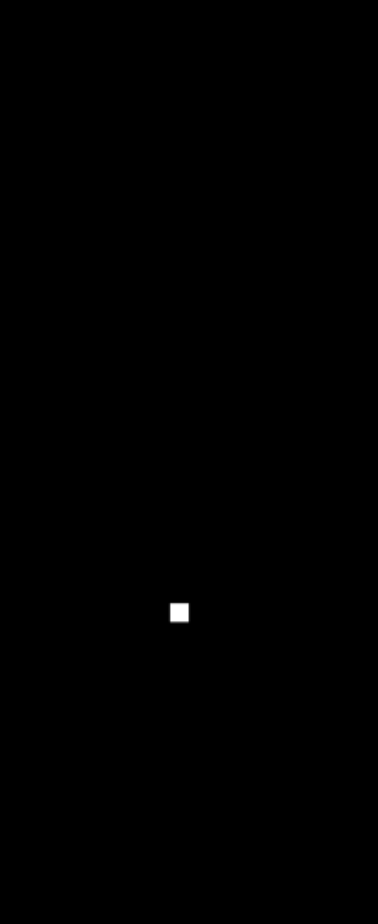} &
        \includegraphics[width=0.10\textwidth]{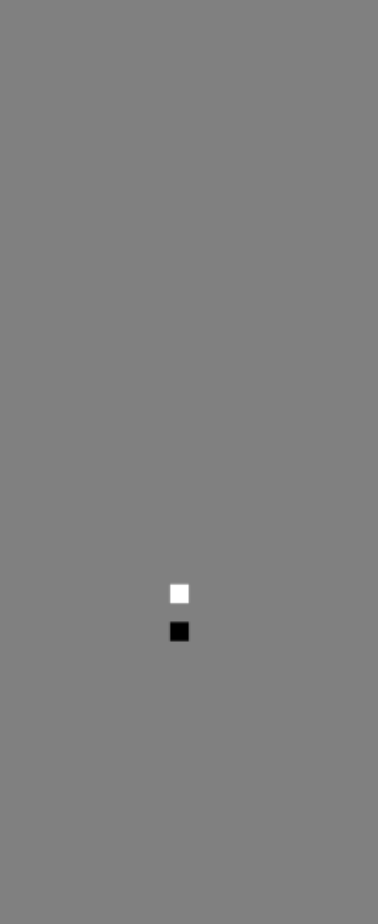} & 
        \includegraphics[width=0.24\textwidth]{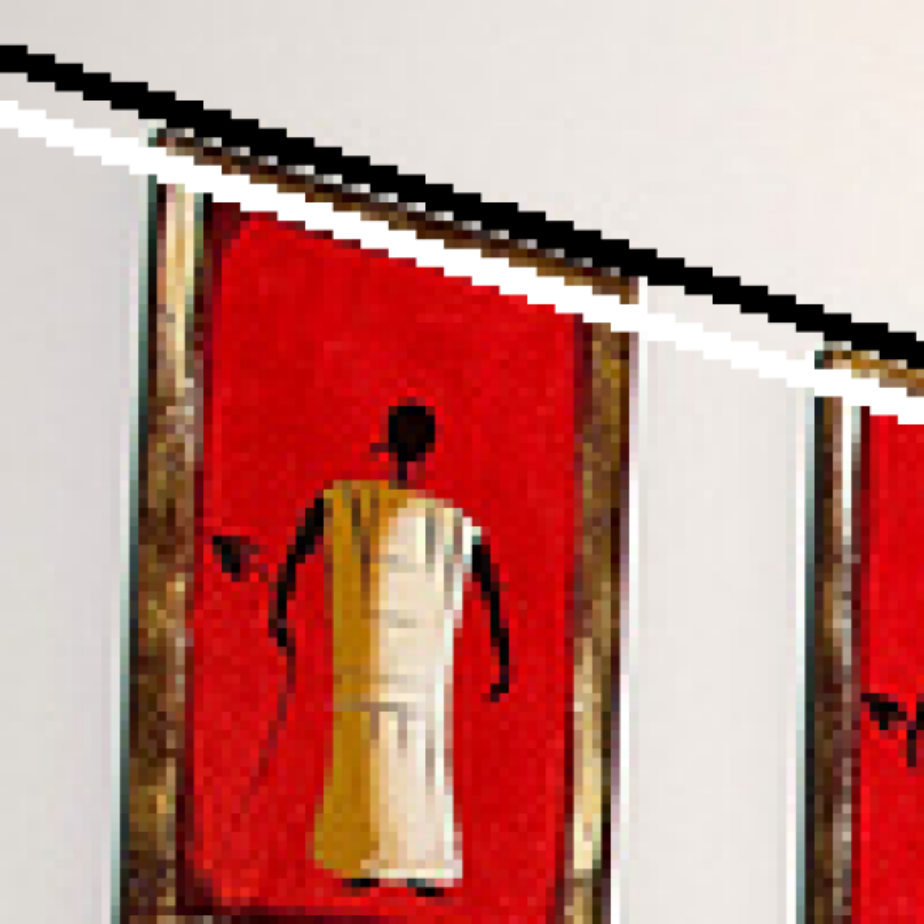} \\
        \footnotesize{(a) Line (orange)} & \footnotesize{(b) Bin in $\mathcal{HT}$ } & \footnotesize{(c) Filter in $\mathcal{HT}$ } & \footnotesize{(d) $\mathcal{IHT}$  } \\
    \end{tabular}
    \caption{Local filters in the Hough domain correspond to global structure in the image domain. (a) An input line in orange. (b) The line becomes a point in Hough domain. (c) A local $[-1, 0, 1]^\intercal$ filter in Hough domain. (d) The inverse of the local Hough filter corresponds to a global line filter in the image domain. }
    \label{fig:edge_in_HT}
\end{figure}

\begin{figure}[t!]
    \centering
    \begin{tabular}{c@{\hskip 0.25 in}c@{\hskip 0.10 in}c@{\hskip 0.10 in}c}
    \includegraphics[width=0.24\textwidth]{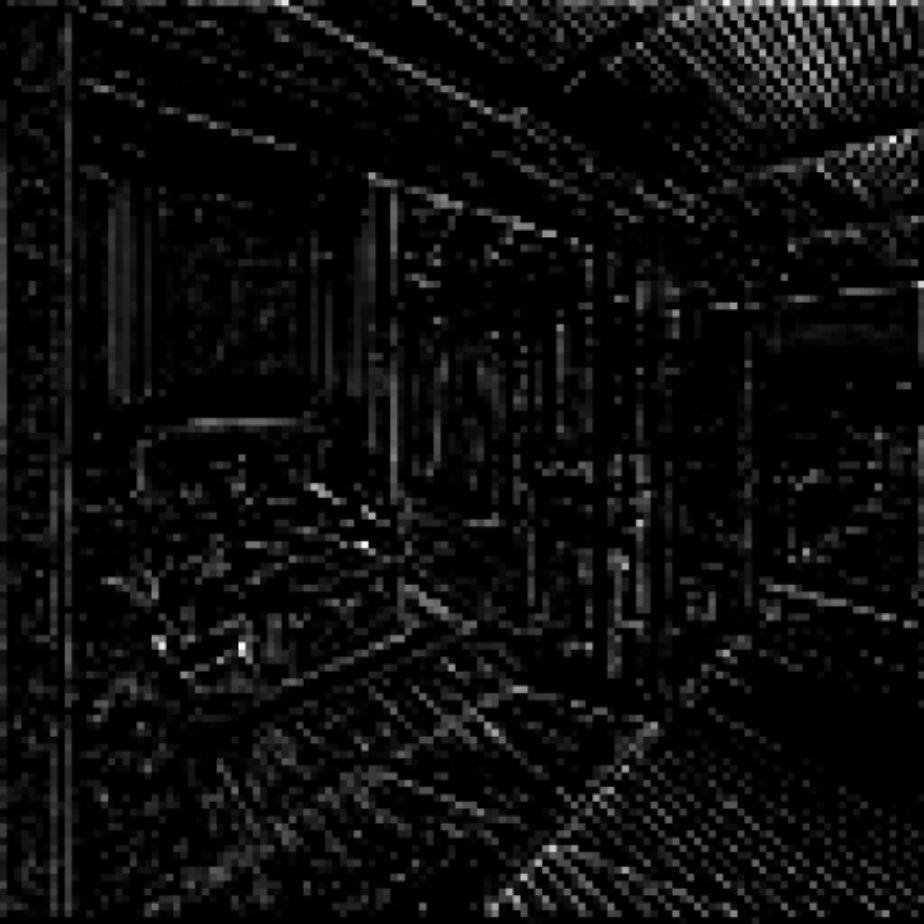} &
    \includegraphics[width=0.08\textwidth]{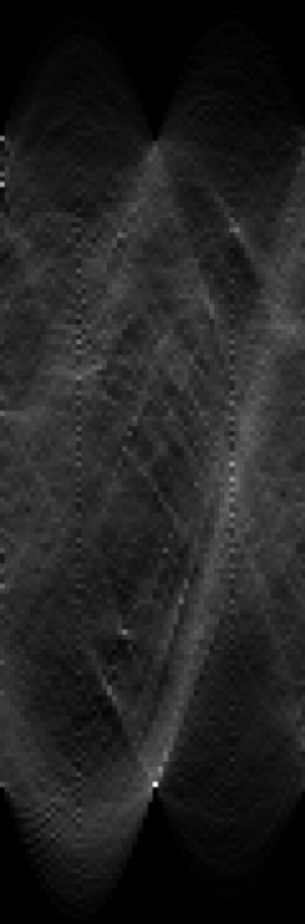} &
    \includegraphics[width=0.08\textwidth]{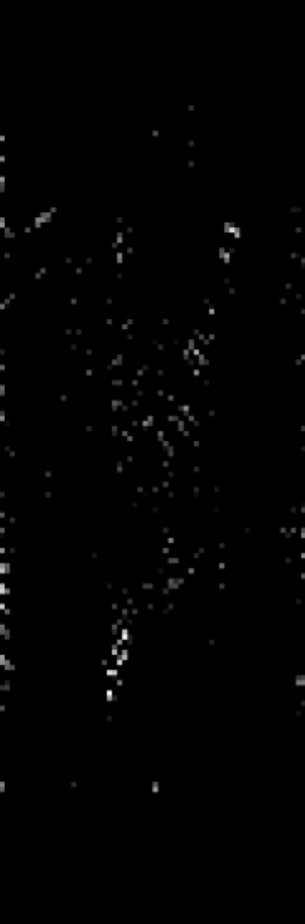} &
    \includegraphics[width=0.24\textwidth]{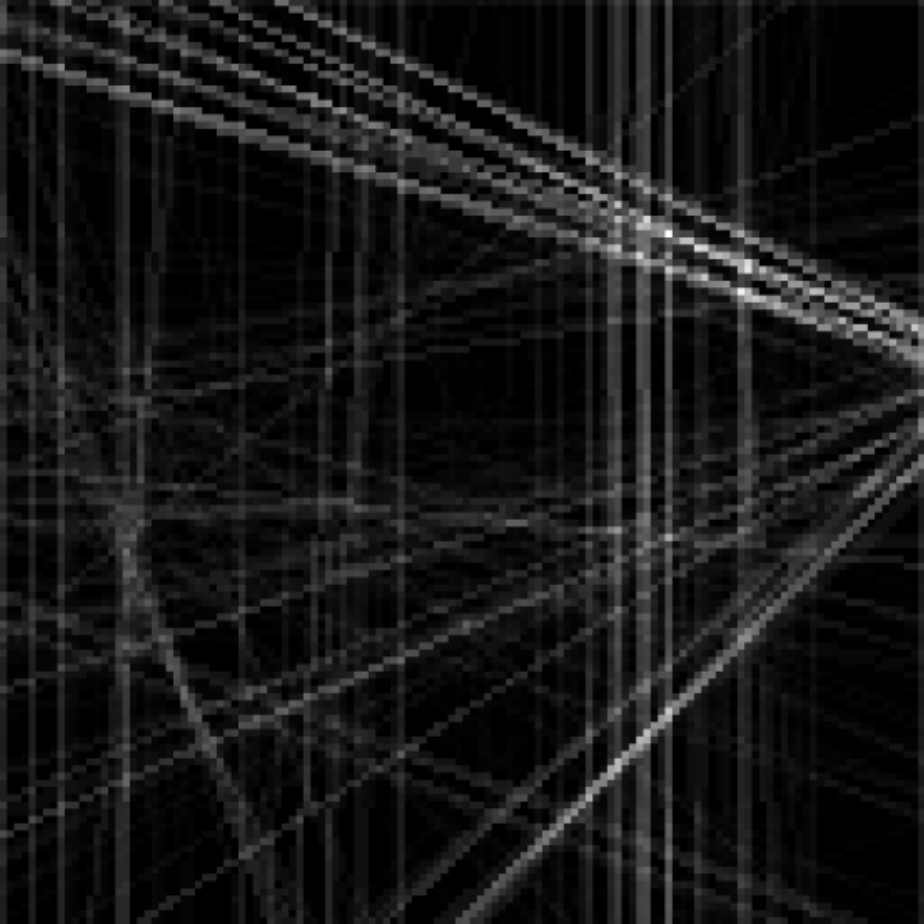} \\
    (a) Input featuremap & (b) $\mathcal{HT}$ & (c) Filtered $\mathcal{HT}$ & (d) $\mathcal{IHT}$\\
    \end{tabular}
    \caption{ Noisy local features aggregated globally by learning filters in the Hough domain. 
    (a) Input featuremap with noisy discontinuous lines. 
    (b) The output of the $\mathcal{HT}$ layer using $183$ offsets and $60$ angles. 
    (c) The result after filtering in the Hough domain. The Hough map contains only the bins corresponding to lines.
    (d) The output of $\mathcal{IHT}$ layer which receives as input the filtered Hough map. The lines are now clearly visible.}
    \label{fig:conv}
\end{figure}

Local operations in Hough space correspond to global operations in the image space, see figure~\ref{fig:edge_in_HT}. Therefore, local convolutions over Hough bins are global convolutions over lines in the image.
We learn filters in the Hough domain to take advantage of the global structure, as done in the Radon transform literature~\cite{magnusson1993linogram}. The filtering in the Hough domain is done locally over the offsets, for each angle direction \cite{nikolaev2008hough,wei2019x}.
We perform channel-wise 1$D$ convolutions in the Hough space over the offsets, $\rho$, as the Hough transform is also computed channel-wise over the input featuremaps.  In Figure~\ref{fig:conv} we show an example; note that the input featuremap lines are noisy and discontinuous and after applying 1$D$ convolutions in Hough space the informative bins are kept and when transformed back to the image domain by the $\mathcal{IHT}$ contains clean lines.  

Inspired by the Radon literature~\cite{magnusson1993linogram,nikolaev2008hough,wei2019x} we initialize the channel-wise filters, $f$, with sign-inverted Laplacians by using the second order derivative of a 1$D$ Gaussian with randomly sampled scale, $\sigma$:  
\begin{alignat}{1}
    f(\rho) \overset{init}{=} - \frac{\partial^2 g(\rho,\sigma)}{\partial \rho^2}, 
\end{alignat}
where $g(\rho, \sigma)$ is a 1$D$ Gaussian kernel. 
We normalize each filter to have unit $L_1$ norm and clip it to match the predefined spatial support. 
We, subsequently, add two more 1$D$ convolutional layers for reducing and merging the channels of the Hough transform map.
This lowers the computations needed in the inverse Hough transform. Our block is visualized in Figure~\ref{fig:block}.

%% file: experiments.tex
\section{Experiments}
We conduct experiments on three datasets: a controlled Line-Circle dataset, the Wireframe (ShanghaiTech) \cite{huang2018learning} dataset and the York Urban \cite{denis2008efficient} dataset. We evaluate the added value of global Hough priors, convolutions in the Hough domain, and data efficiency.
We provide our source code online\footnote{\url{https://github.com/yanconglin/Deep-Hough-Transform-Line-Priors}}.

\begin{figure}[t!]
    \centering
        \begin{tabular}{ccccc}        
            \includegraphics[width=0.18\textwidth]{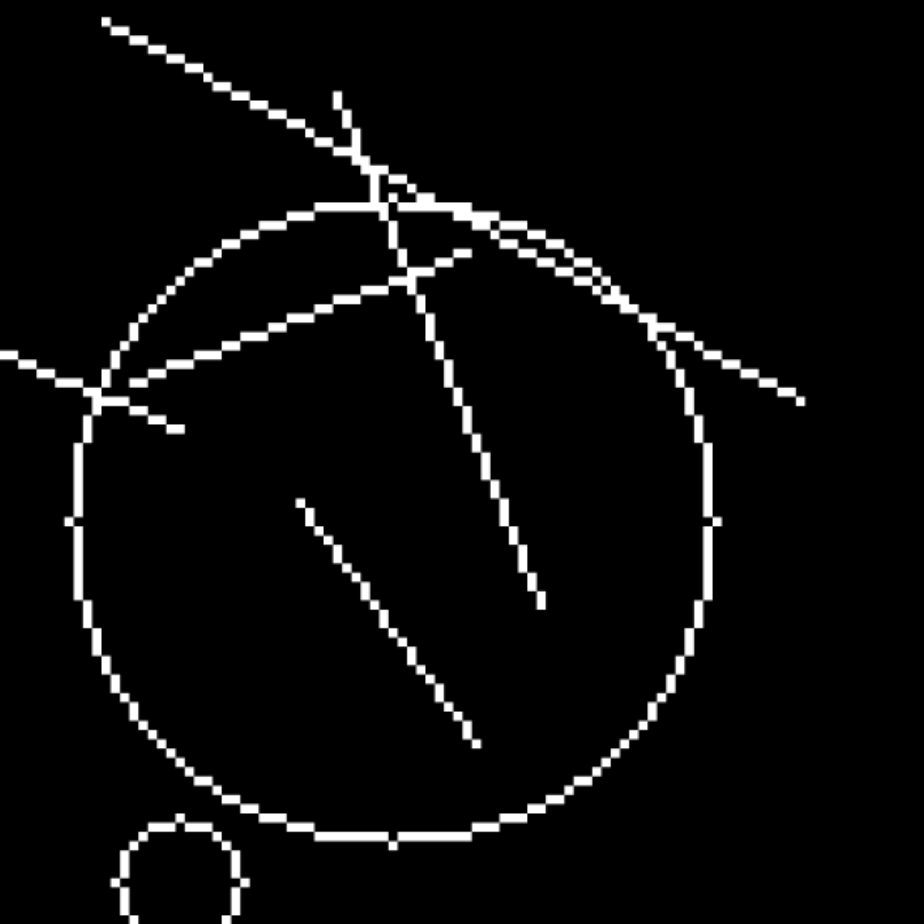} &
            \includegraphics[width=0.18\textwidth]{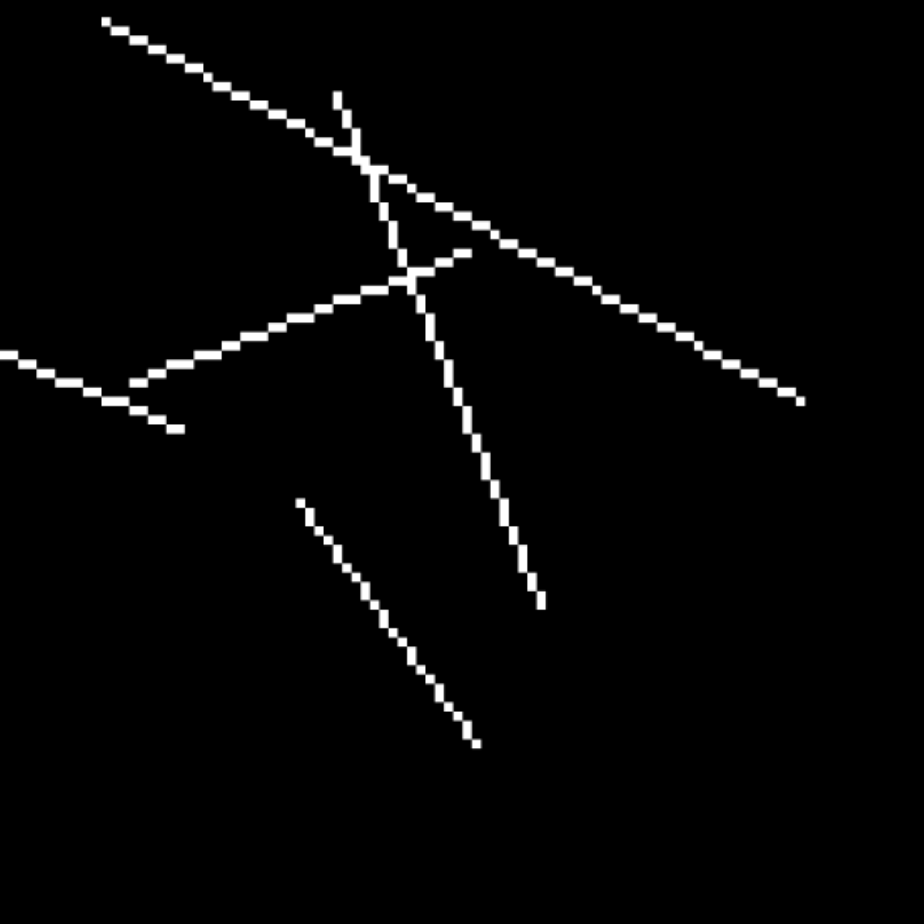} &
            \includegraphics[width=0.18\textwidth]{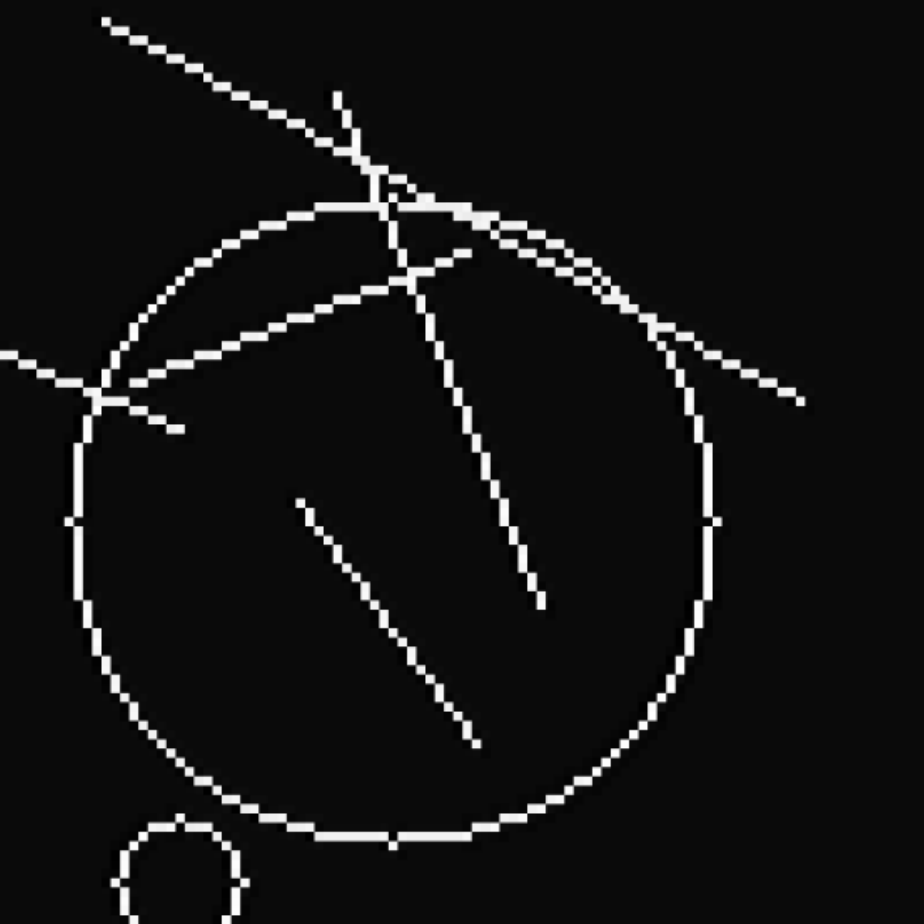} &
            \includegraphics[width=0.18\textwidth]{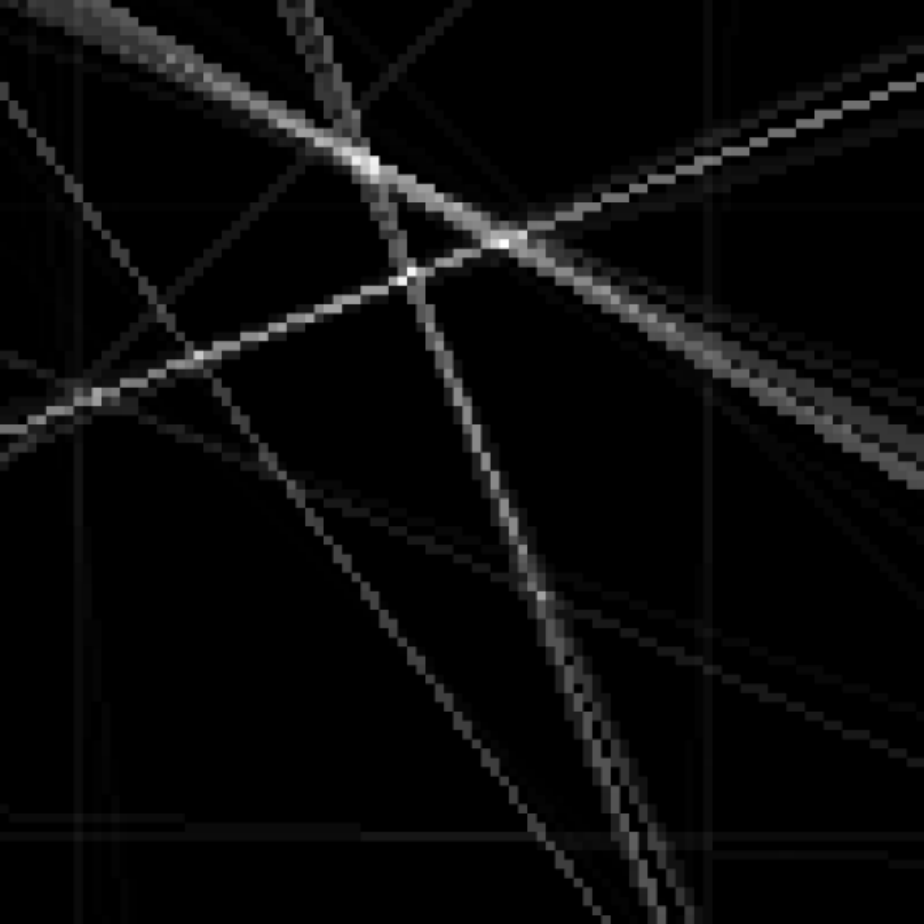} &
            \includegraphics[width=0.18\textwidth]{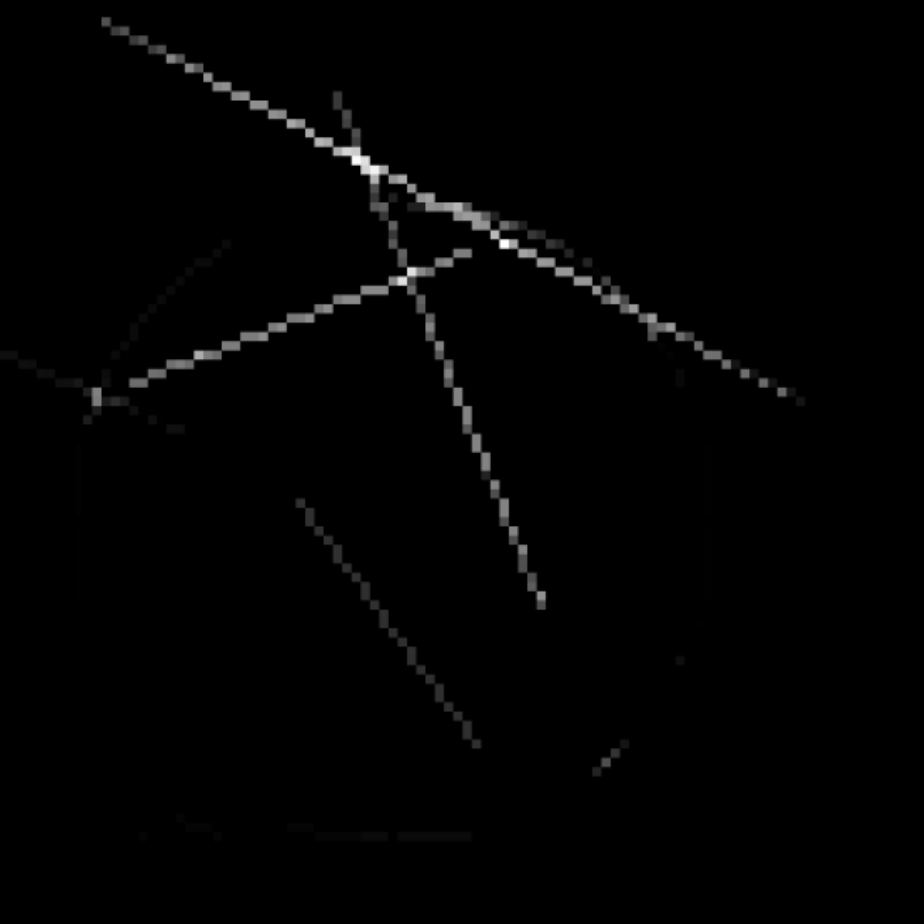}\\
            \includegraphics[width=0.18\textwidth]{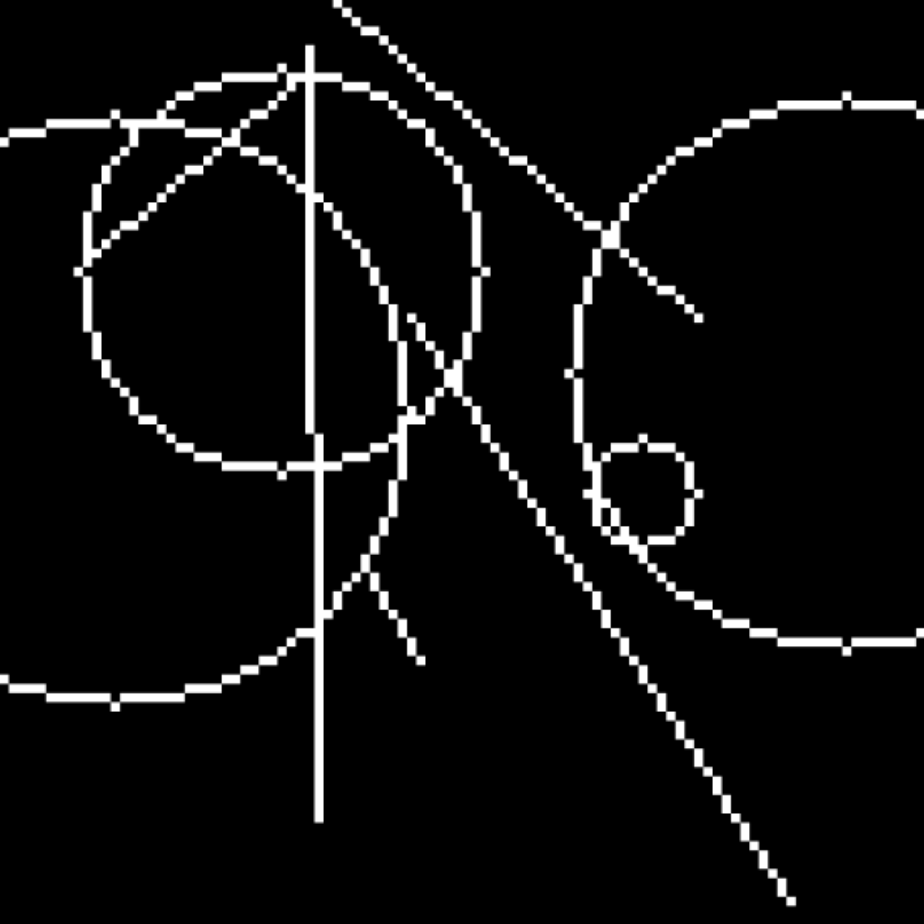} &
            \includegraphics[width=0.18\textwidth]{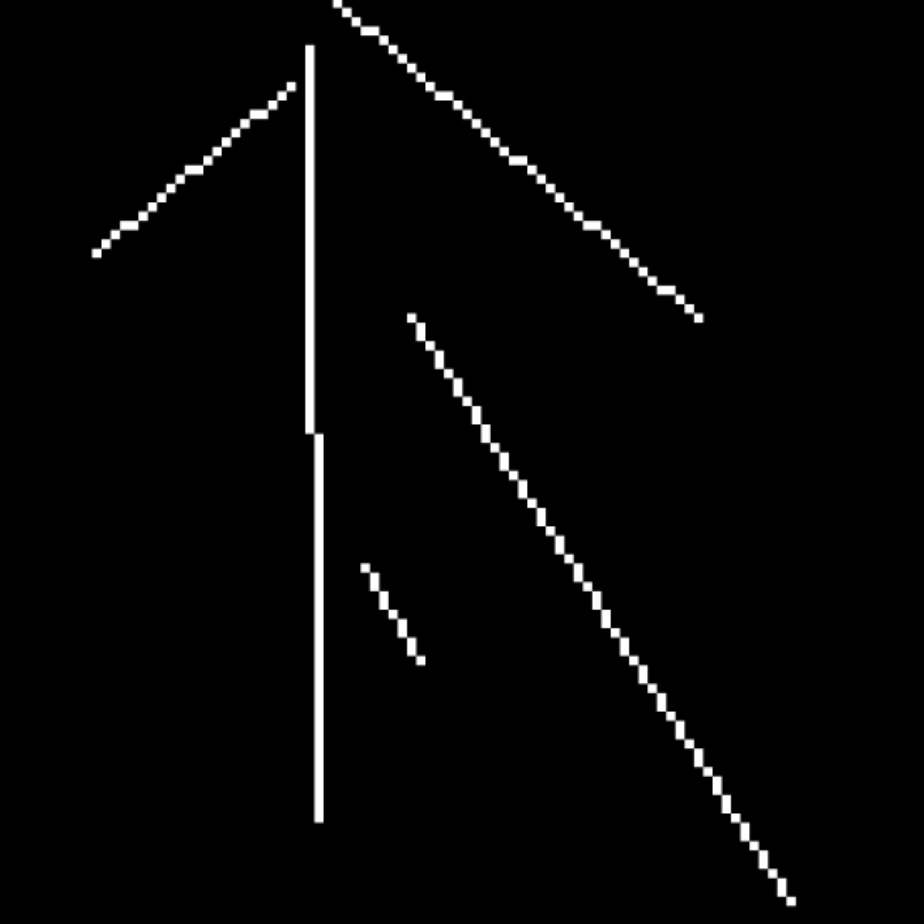} &
            \includegraphics[width=0.18\textwidth]{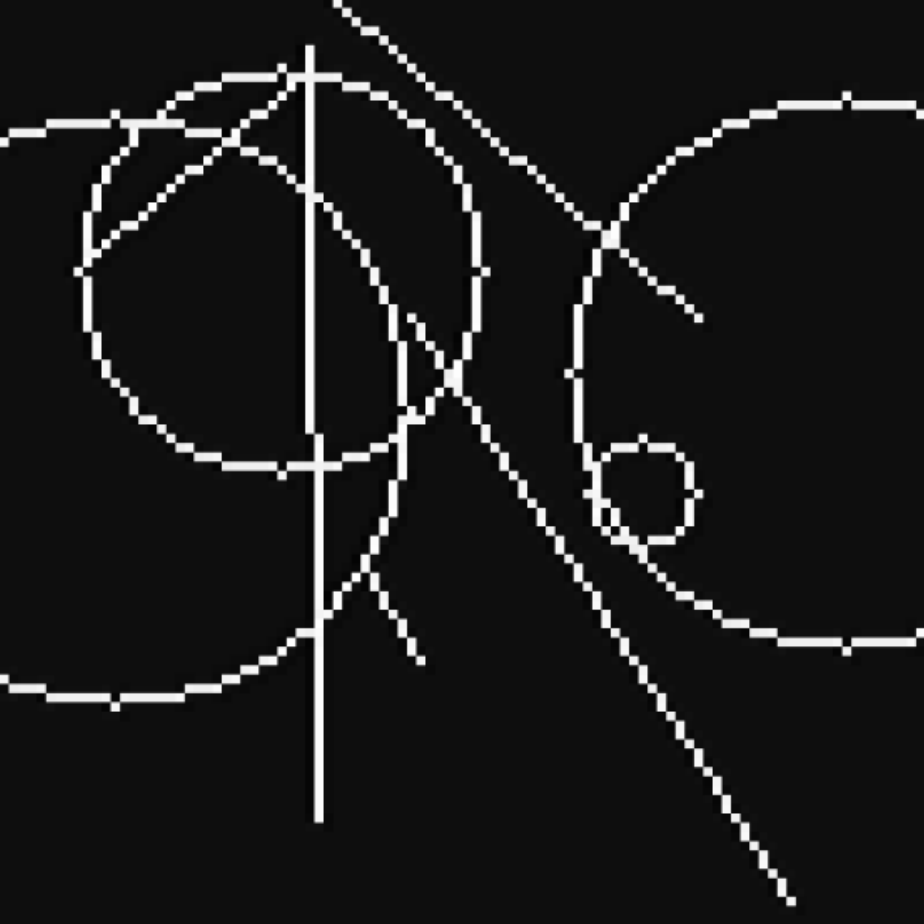} &
            \includegraphics[width=0.18\textwidth]{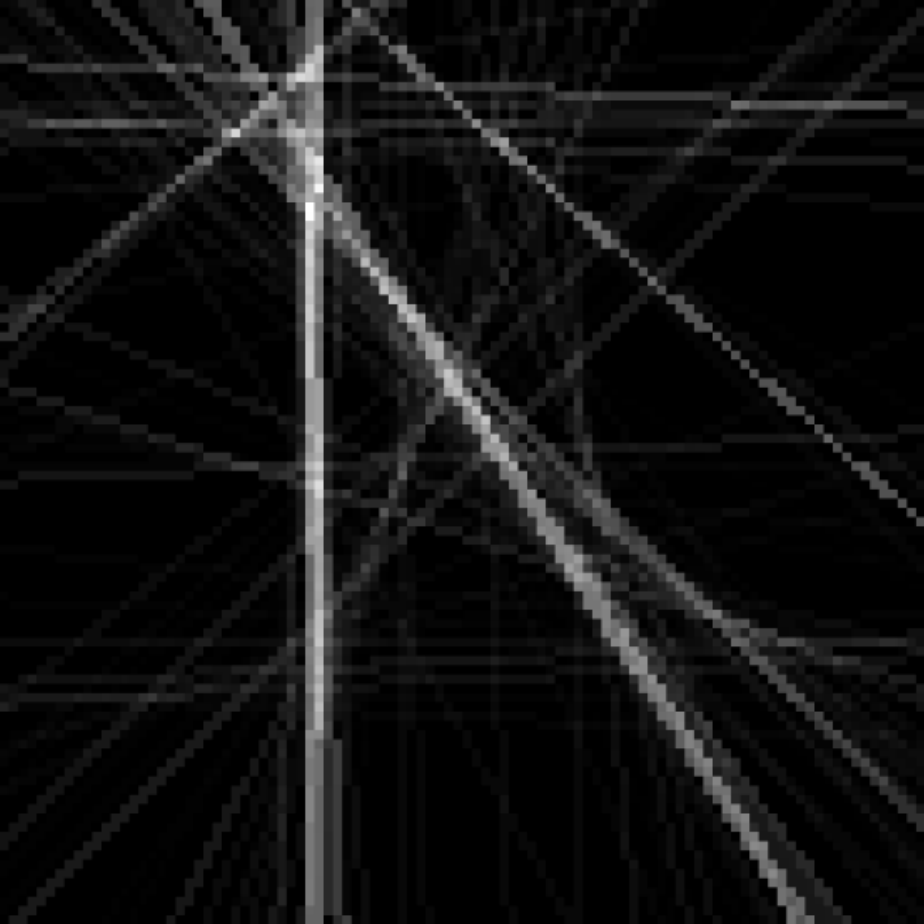} &
            \includegraphics[width=0.18\textwidth]{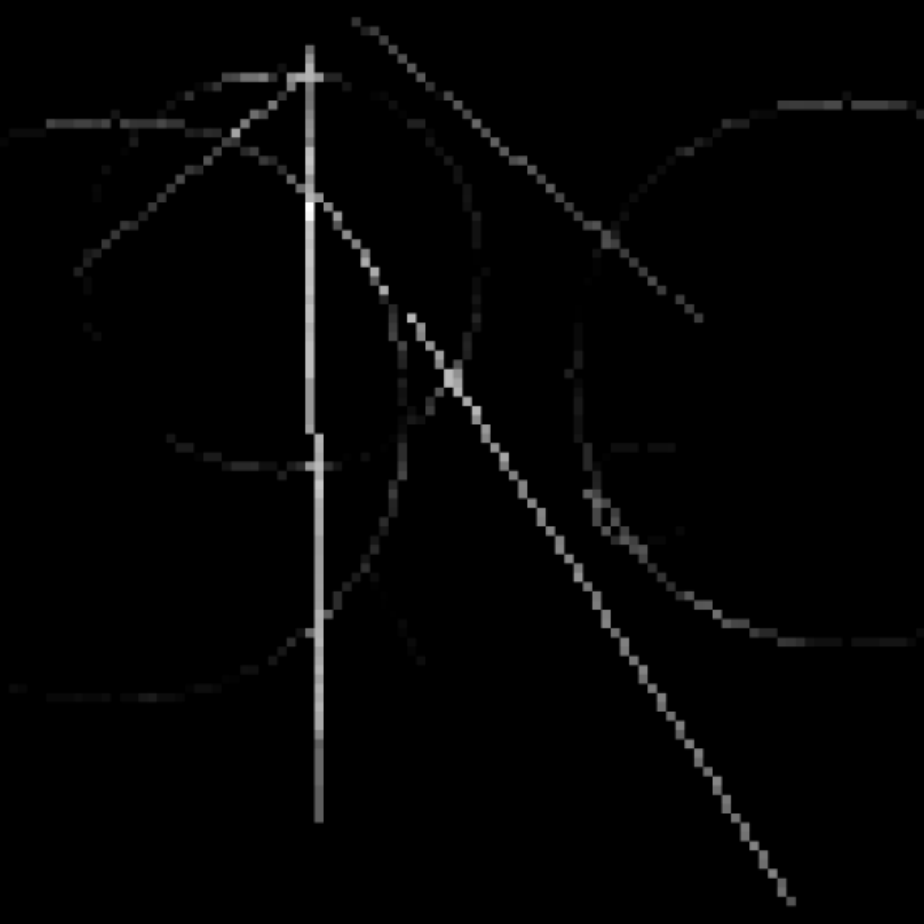} \\ \cmidrule(r){3-5}
             &  & AP: 24.97\% & AP: 38.57\% & \textbf{AP: 56.33\%} \\ \cmidrule(r){3-5}
             \scriptsize{(a) Input} & \scriptsize{(b) GT} & \scriptsize{(c) Local-only} & \scriptsize{(d) Global-only} & \scriptsize{(e) Local+global} \\             
        \end{tabular} 
       \caption{\textbf{Exp 1:} 
    Results in AP (average precision) and image examples of the Line-Circle dataset. 
    Using local+global information detects not only the direction of the lines, as the global-only does, but also their extent.
    (See the appendix for more results).}
    \label{fig:exp1}
\end{figure}

\subsection{\textbf{Exp 1:}  Local and global information for line detection.}
\label{exp1}
\noindent\emph{Experimental setup.} 
We do a controlled experiment to evaluate the combination of global Hough line priors with learned local features. We target a setting where local-only is difficult and create a Line-Circle dataset of 1,500 binary images of size 100x100 px, split into 744 training, 256 validation, and 500 test images, see figure~\ref{fig:exp1}.
Each image contains 1 to 5 randomly positioned lines and circles of varying sizes. 
The ground truth has only line segments and we optimize the $L_2$ pixel difference.
We follow the evaluation protocol described in \cite{huang2018learning,martin2004learning,maire2008using} and report AP (average precision) over a number of binarization thresholds varying from $0.1$ to $0.9$, with a matching tolerance of $0.0075$ of the diagonal length~\cite{martin2004learning}.

We evaluate three settings: local-only, global-only, and local+global. 
The aim is not fully solving the toy problem, but rather testing the added value of the $\mathcal{HT}$ and $\mathcal{IHT}$ layers. Therefore, all networks have only 1  layer with 1 filter, where the observed gain in AP cannot be attributed to the network complexity. 
For local-only we use a a single $3\times 3$ convolutional layer followed by ReLU. For global-only we use an $\mathcal{HT}$ layer followed by a $3\times 1$ convolutional layer, ReLU, and an $\mathcal{IHT}$ layer. For local+global we use the same setting as for global-only, but multiply the output of the $\mathcal{IHT}$ layer with the input image, thus combining global and local image information.
All networks have only 1 filter and they are trained from scratch with the same configuration.

\noindent\emph{Experimental analysis.} 
In the caption of figure~\ref{fig:exp1} we show the AP on the Line-Circle dataset.
The global-only model can correctly detect the line directions thus it outperforms the local-only model.
The global+local model can predict both the line directions and their extent, by combining local and global image information. Local information only is not enough, and indeed the $\mathcal{HT}$ and $\mathcal{IHT}$ layers are effective.

\subsection{\textbf{Exp 2:} The effect of convolution in the Hough domain}
\label{exp2}
\noindent\emph{Experimental setup.}
We evaluate our \model design, specifically, the effect of convolutions in the Hough domain on a subset of the Wireframe (ShanghaiTech) dataset \cite{huang2018learning}.
The Wireframe dataset contains 5,462 images. We sample from the training set 1,000 images for training, and 256 images for validation, and use the official test split.
As in \cite{zhou2019learning}, we resize all images to 512 $\times$ 512 px.
The goal is predicting pixels along line segments, where we report AP using the same evaluation setup as in \textbf{Exp 1}, and we optimize a binary cross entropy loss.

We use a ResNet \cite{he2016deep} backbone architecture, containing 2 convolutional layers with ReLU, followed by 2 residual blocks, and another convolutional layer with a sigmoid activation. The evaluation is done on predictions of $128\times 128$ px, and the ground truth are binary images with line segments.
We insert our \model after every residual block. 
All layers are initialized with the He initialization \cite{he2015delving}. 

We test the effect of convolutions in the Hough domain by considering in our \model:
(0) not using any convolutions, 
(1) using a 1$D$ convolution over the offsets,
(2) a channel-wise 1$D$ convolution initialized with sign-inverted Laplacian filters,
(3) our complete \model containing Laplacian-initialized 1$D$ convolution and two additional 1$D$ convolutions for merging and reducing the channels, and
(4) using three standard $3\times3$ convolutions.\\

\begin{table}[t!]
    \centering
    \begin{tabular}{l@{\hskip 0.2in}l@{\hskip 0.4in}l}
             \toprule
           Networks & \model     & AP \\ \midrule
           (0) & $w/o$ convolution                                    & 61.77 \% \\
           (1) & $[9\times1]$                                         & 63.02 \% \\
           (2) & $[9\times1]$-Laplacian                               & 66.19 \% \\
           (3) & $[9\times1]$-Laplacian + $[9\times1]$ + $[9\times1]$ & \textbf{66.46} \% \\
           (4) &$[3\times3]$ + $[3\times3]$ + $[3\times3]$            & 63.90 \%  \\
           \bottomrule
        \end{tabular} 
    \caption{\textbf{Exp 2:} The effect of convolution in the Hough domain, in terms of AP on a subset of the Wireframe (ShanghaiTech) dataset \cite{huang2018learning}.
    No convolutions perform worst (0). 
    The channel-wise Laplacian-initialized filters (2) perform better than the standard 1$D$ convolutions (1). 
    Our proposed \model (3) versus using $[3\times3]$ convolutions (4), shows the added value of following the Radon transform practices.}
    \label{tab:exp2}
\end{table}

\noindent\emph{Experimental analysis.} 
Table~\ref{tab:exp2} shows that using convolutions in the Hough domain is beneficial. 
The channel-wise Laplacian-initialized convolution is more effective than the standard 1$D$ convolution using the He initialization \cite{he2015delving}. 
Adding extra convolutions for merging and reducing the channels gives a small improvement in AP, however we use these for practical reasons rather than improved performance. 
When comparing option (3) with (4), we see clearly the added value of performing 1$D$ convolutions over the offsets instead of using standard $3\times 3$ convolutions. 
This experiment confirms that our choices, inspired from the Radon transform practices, are indeed effective for line detection.


\begin{figure}[t!]
    \centering
    \begin{tabular}{ccc}
        \includegraphics[width=0.33\textwidth]{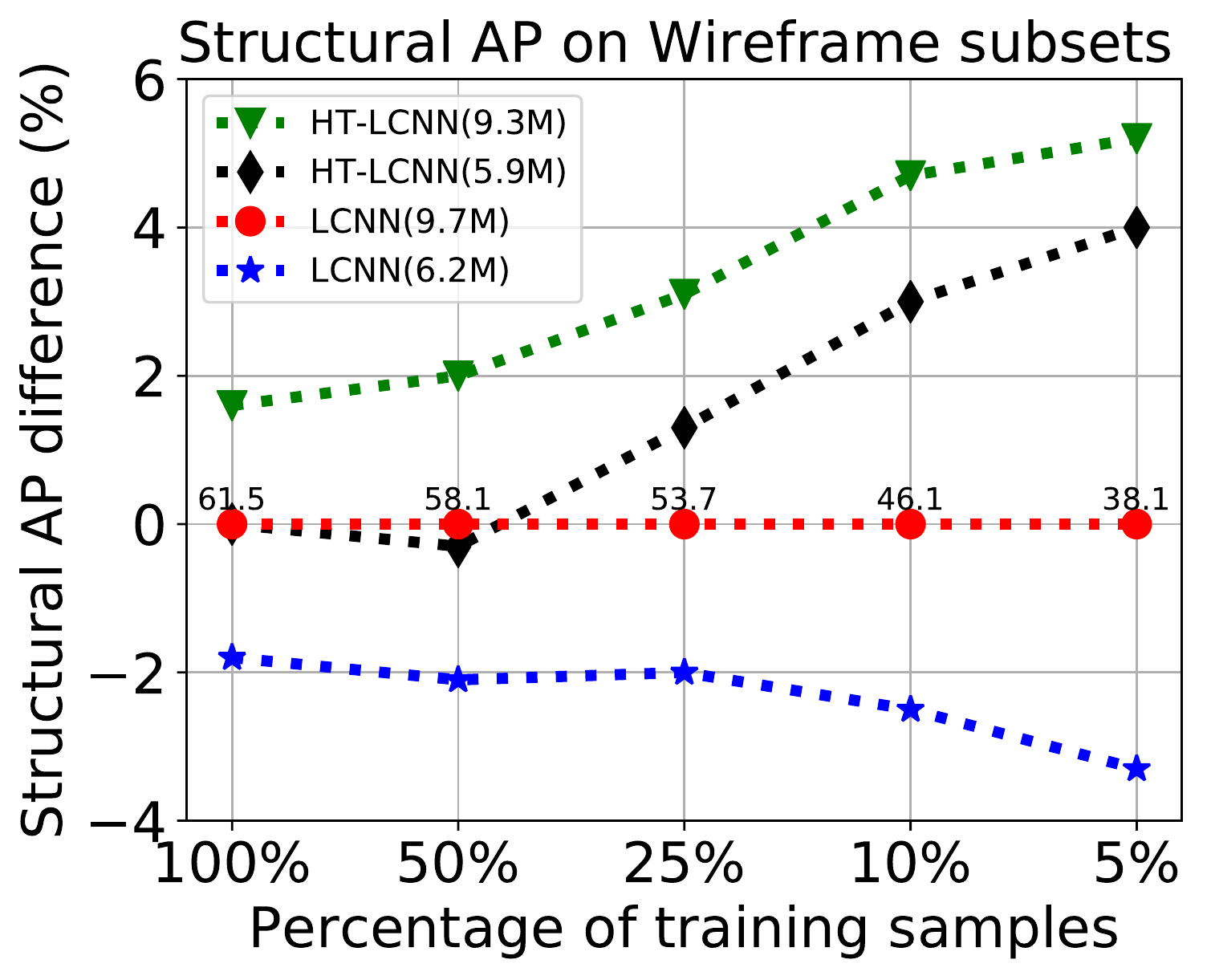} &
        \includegraphics[width=0.33\textwidth]{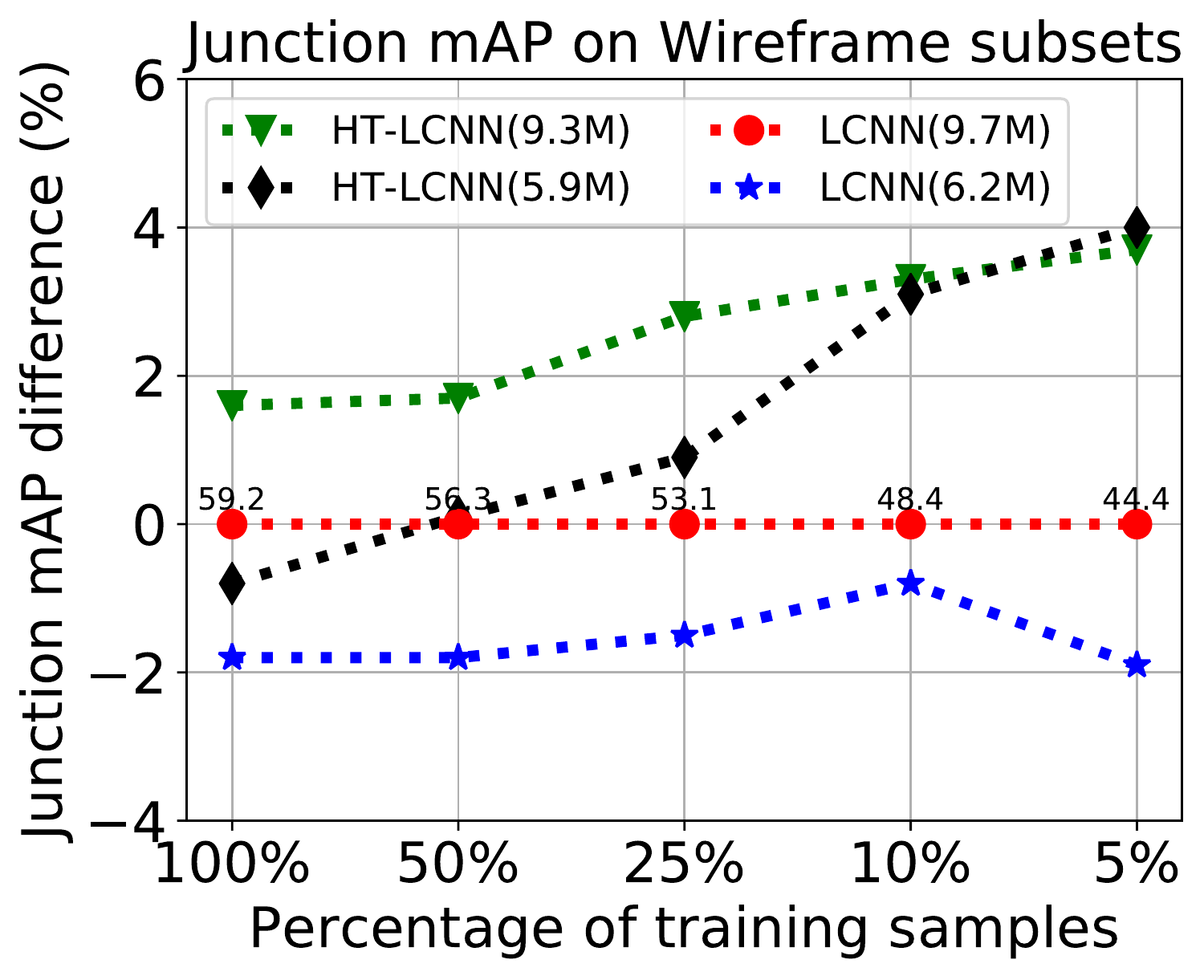} &
        \includegraphics[width=0.33\textwidth]{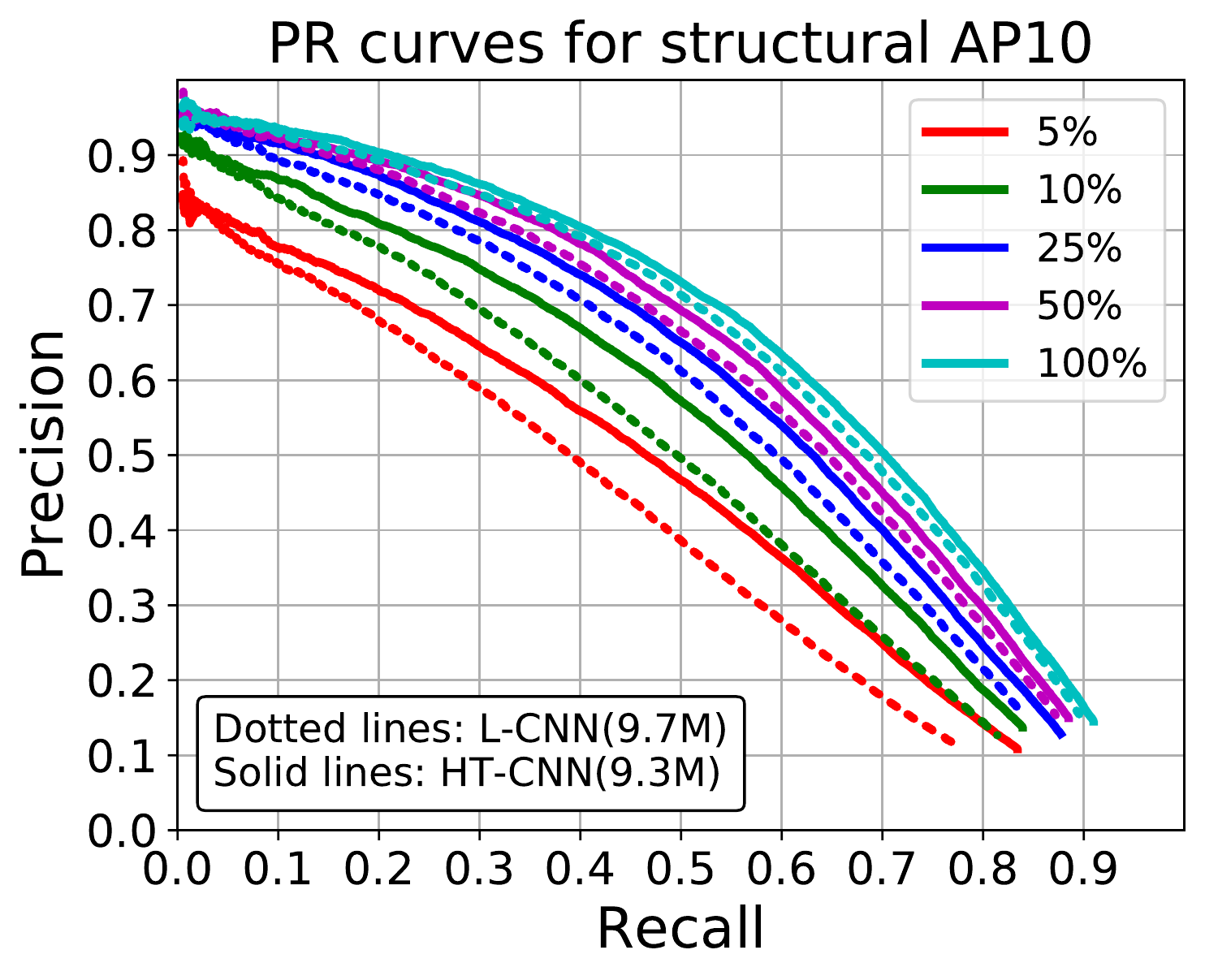} \\
        
        \includegraphics[width=0.33\textwidth]{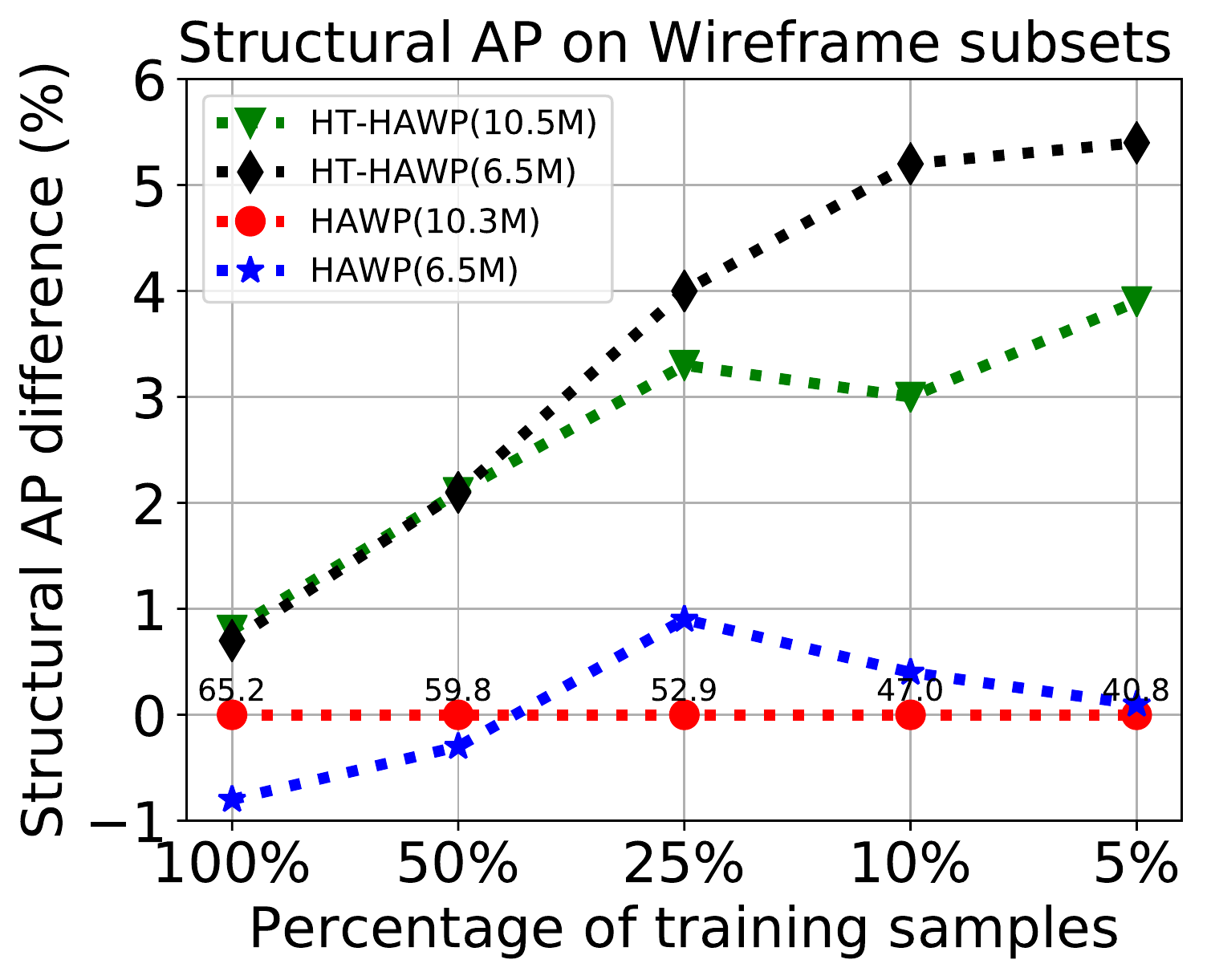} &
        \includegraphics[width=0.33\textwidth]{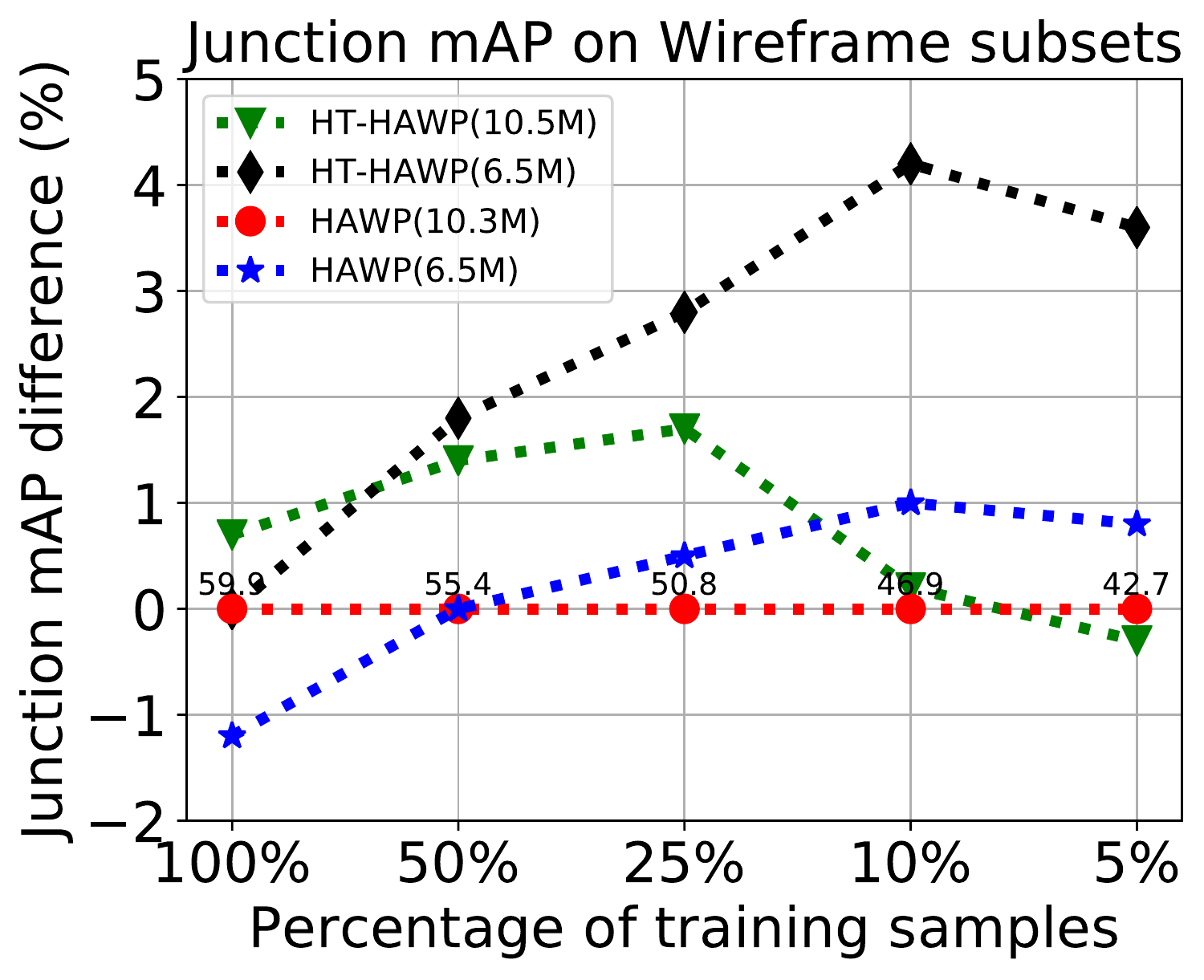} &
        \includegraphics[width=0.33\textwidth]{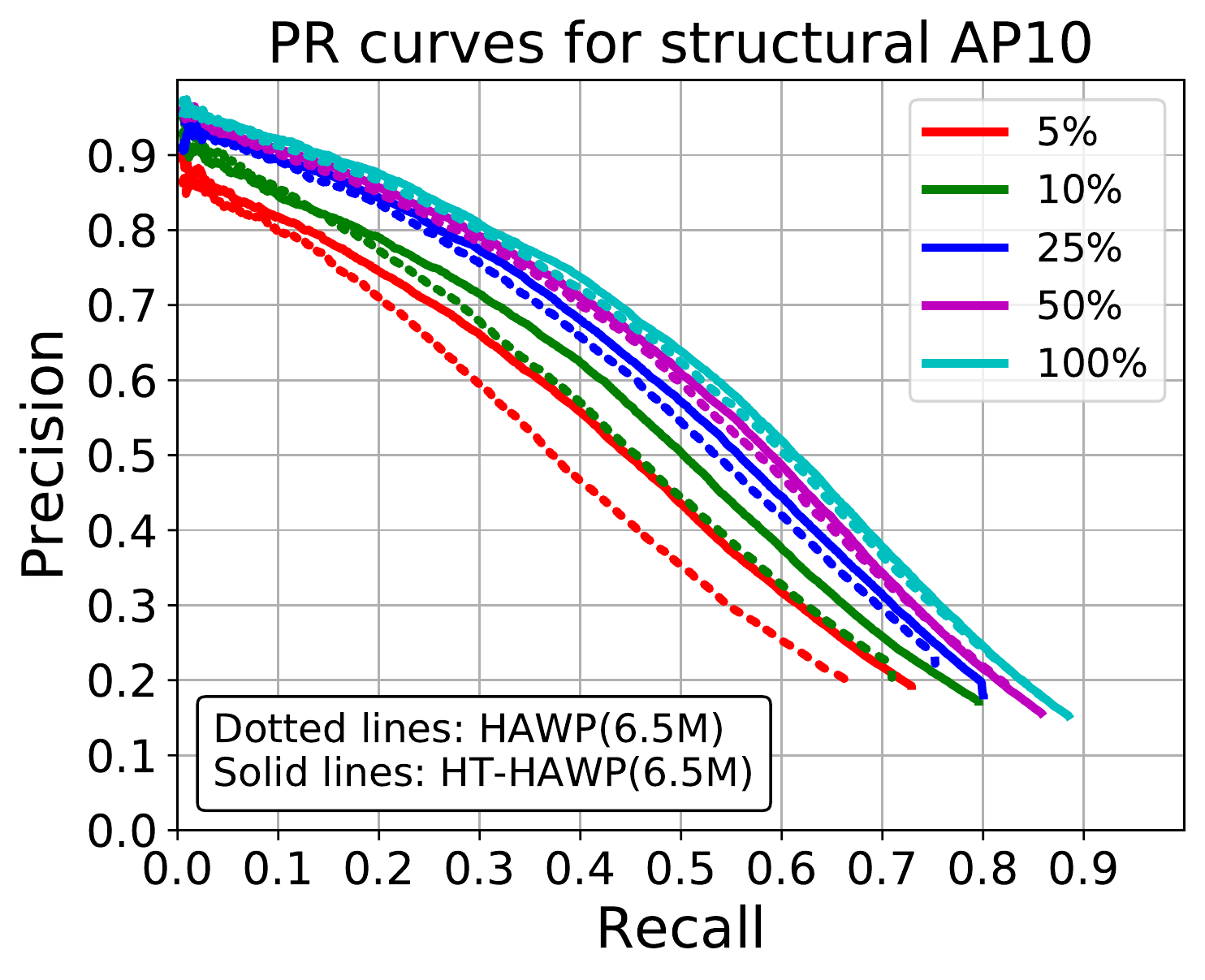} \\
        (a) Structural-AP$^{10}$ & (b)Junction-mAP & (c) PR for structural-AP$^{10}$\\
    \end{tabular}
    \caption{\textbf{Exp 3.(a):} Data efficiency on subsets of the Wireframe (ShanghaiTech) dataset.
    We compare different sized variants of our HT-LCNNs and HT-HAWPs with LCNNs \cite{zhou2019end} and HAWPs \cite{xue2020holistically}. 
    In (a) and (b) we show the absolute difference for structural-AP and junction-mAP compared to the best baseline. 
    In (c) we show PR curves for structural-$AP^{10}$.
    Our HT-LCNN and HT-HAWP models are consistently better than their counterparts. 
    The benefit of our \model is accentuated for fewer training samples, where with half the number of parameters our models outperform the LCNN and HAWP baselines.
    Adding geometric priors improves data efficiency.}
    \label{fig:exp3_a}
\end{figure}

\subsection{\textbf{Exp 3:} \model for line segment detection}
\noindent\emph{Experimental setup.} 
We evaluate our \model on the official splits of the Wireframe (ShanghaiTech) \cite{huang2018learning} and York Urban \cite{denis2008efficient} datasets. 
We report structural-AP and junction-mAP. 
Structural-AP is evaluated at AP$^{5}$, AP$^{10}$ thresholds, and the junction-mAP is averaged over the thresholds 0.5, 1.0, and 2.0, as in \cite{zhou2019learning}.
We also report precision-recall, following \cite{almazan2017mcmlsd}, which penalizes both under-segmentation and over-segmentation. 
We use the same distance threshold of $2\sqrt{2}$ px on full-resolution images, as in \cite{almazan2017mcmlsd}. 
For precision-recall, all line segments are ranked by confidence, and the number of top ranking line segments is varied from 10 to 500. 

We build on the successful LCNN \cite{zhou2019end} and HAWP \cite{xue2020holistically} models, where we replace all the hourglass blocks with our \model to create HT-LCNN and HT-HAWP, respectively. 
The hourglass block has twice as many parameters as our \model, thus we vary the number of HT-IHT blocks to match the number of parameters of LCNN, HAWP respectively. 
The networks are trained by the procedure in \cite{xue2020holistically,zhou2019learning}: optimizing binary cross-entropy loss for junction and line prediction, and $L_1$ loss for junction offsets.
The training uses the ADAM optimizer, with scheduled learning rate starting at $4e-4$, and $1e-4$ weight decay, for a maximum of 30 epoch.

\subsubsection{\textbf{Exp 3.(a):} Evaluating data efficiency.}
\label{Exp 3.(a)}

We evaluate data efficiency by reducing the percentage of training samples to $\{ 50\%, 25\%, 10\%, 5\%\}$  and training from scratch on each subset.  
We set aside 256 images for validation, and train all the networks on the same training split and evaluate on the official test split. 
We compare: LCNN(9.7M), LCNN(6.2M) with HT-LCNN(9.3M), HT-LCNN(5.9M), and HAWP(10.3M), HAWP(6.5M) with HT-HAWP(10.5M) and HT-HAWP(6.5M), where we show in brackets the number of parameters.

Figure~\ref{fig:exp3_a} shows structural-$AP^{10}$, junction-mAP and the PR (precision recall) curve of structural-$AP^{10}$ on the subsets of the Wireframe dataset.
Results are plotted relative to our strongest baselines: the LCNN(9.7M) and HAWP(10.3M) models.   
The HT-LCNN and HT-HAWP models consistently outperform their counterparts.
Noteworthy, the HT-LCNN(5.9M) outperforms the LCNN(9.7M) when training on fewer samples, while having 40\% fewer parameters.
This trend becomes more pronounced with the decrease in training data. 
We also observe similar improvement for HT-HAWP over HAWP.
Figure~\ref{fig:exp3_a}(c) shows the PR curve for the structural-$AP^{10}$. 
The continuous lines corresponding to HT-LCNN and HT-HAWP are consistently above the dotted lines corresponding to their counterparts, validating that the geometric priors of our \model are effective when the amount of training samples is reduced.

\begin{figure}[t!]
    \centering
    \begin{tabular}{ccccc}

        \includegraphics[width=0.19\textwidth]{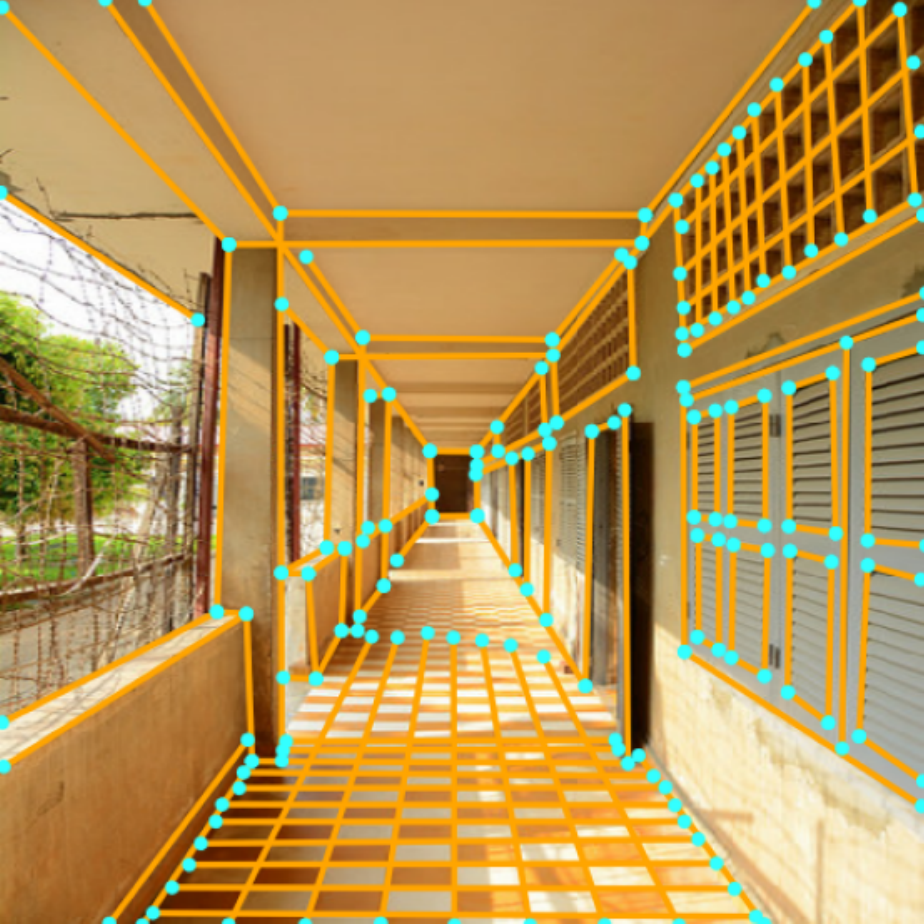} &
        \includegraphics[width=0.19\textwidth]{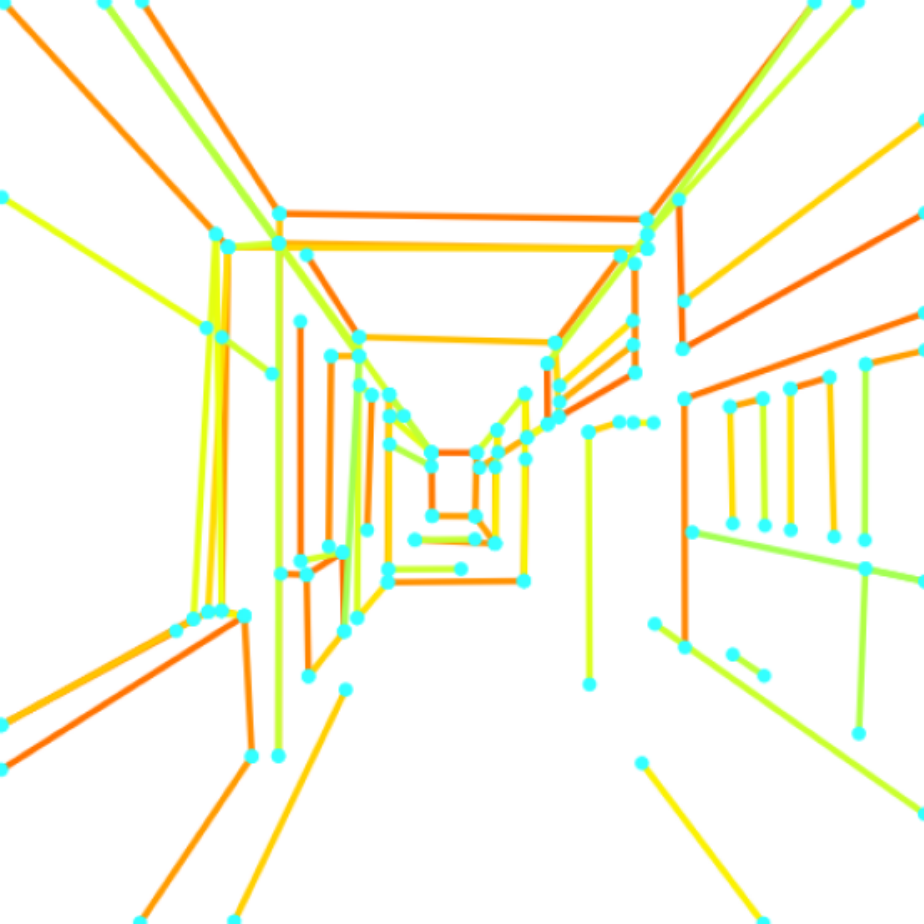} &
        \includegraphics[width=0.19\textwidth]{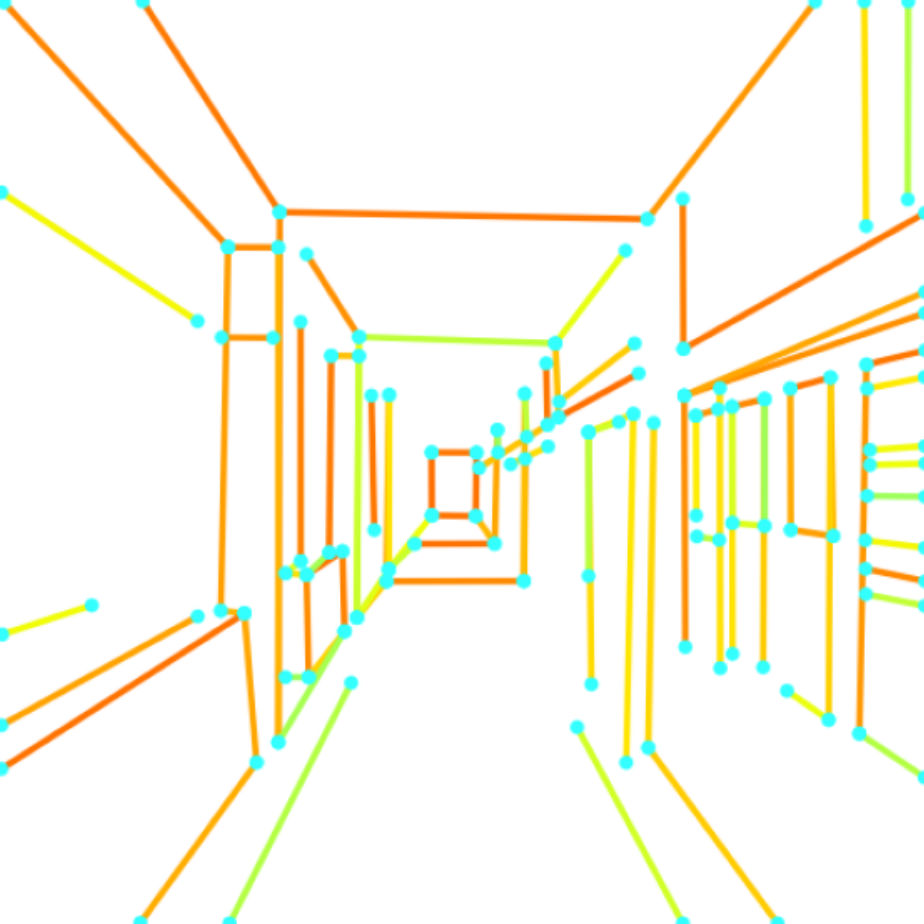} &
        \includegraphics[width=0.19\textwidth]{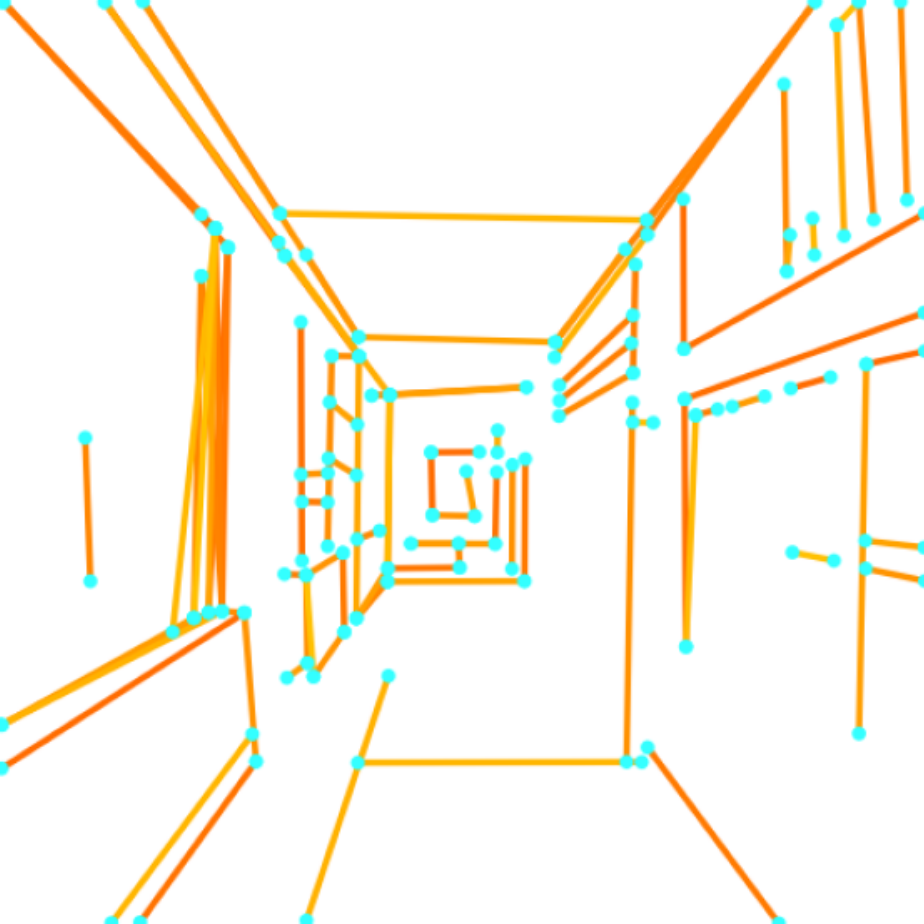} & 
        \includegraphics[width=0.19\textwidth]{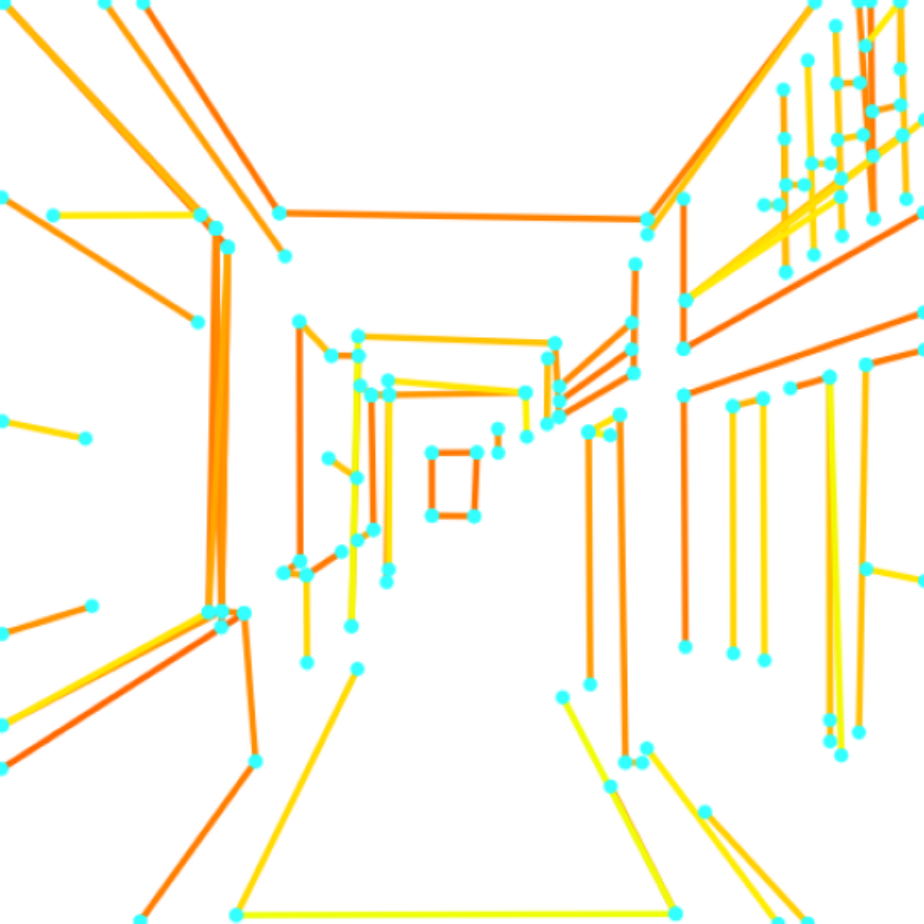}  \\
        \includegraphics[width=0.19\textwidth]{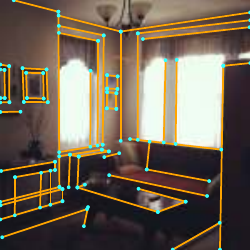} &
        \includegraphics[width=0.19\textwidth]{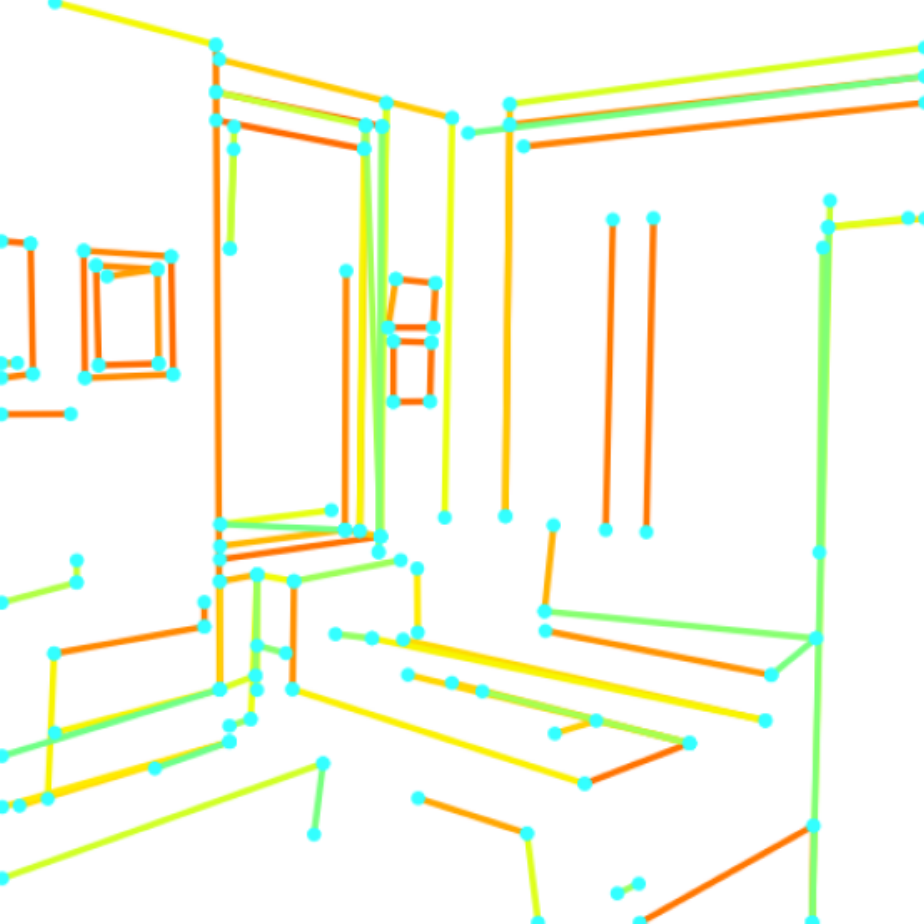} &
        \includegraphics[width=0.19\textwidth]{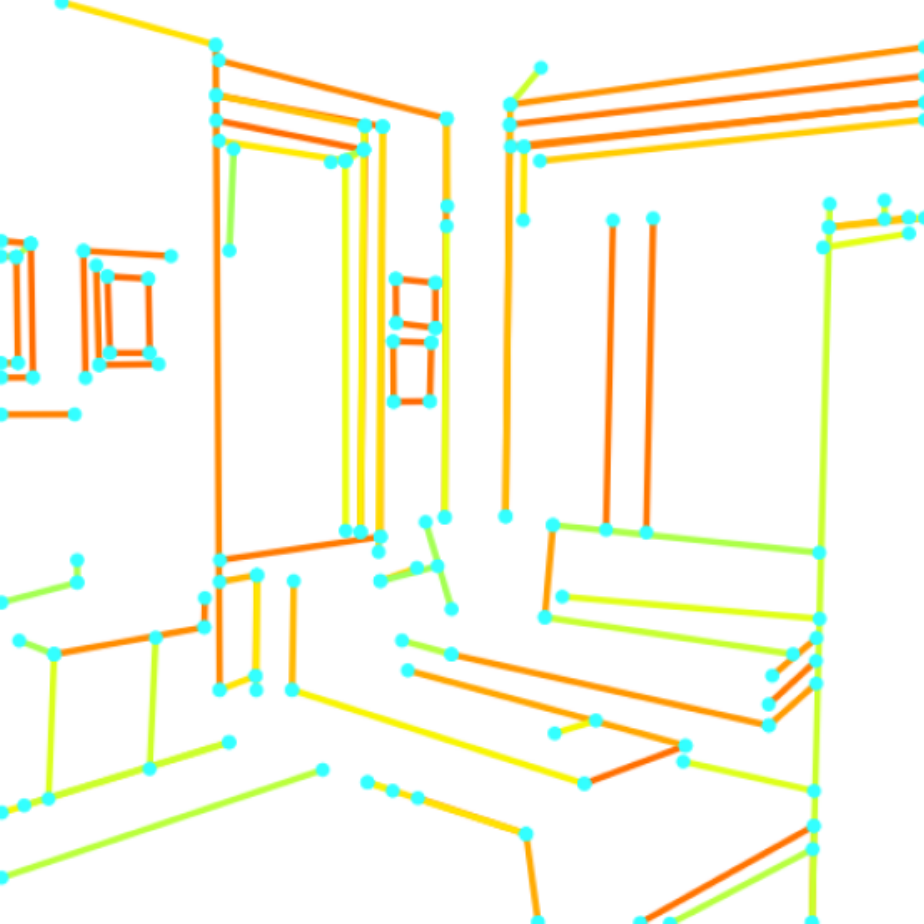} &
        \includegraphics[width=0.19\textwidth]{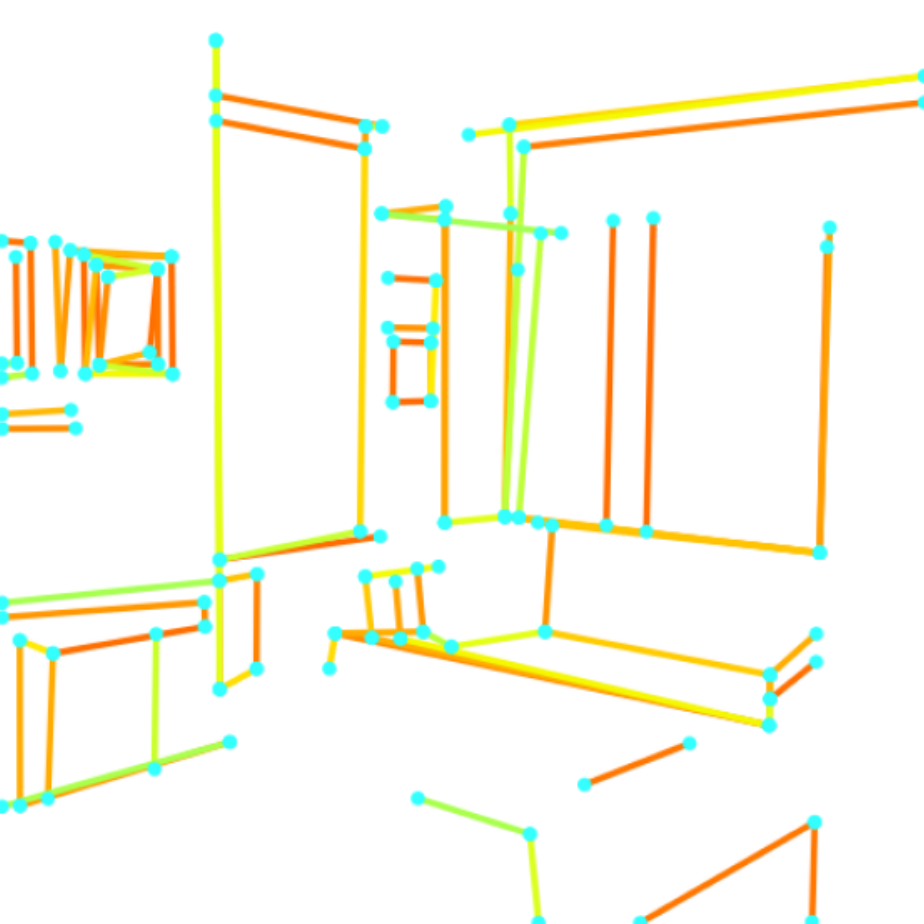} & 
        \includegraphics[width=0.19\textwidth]{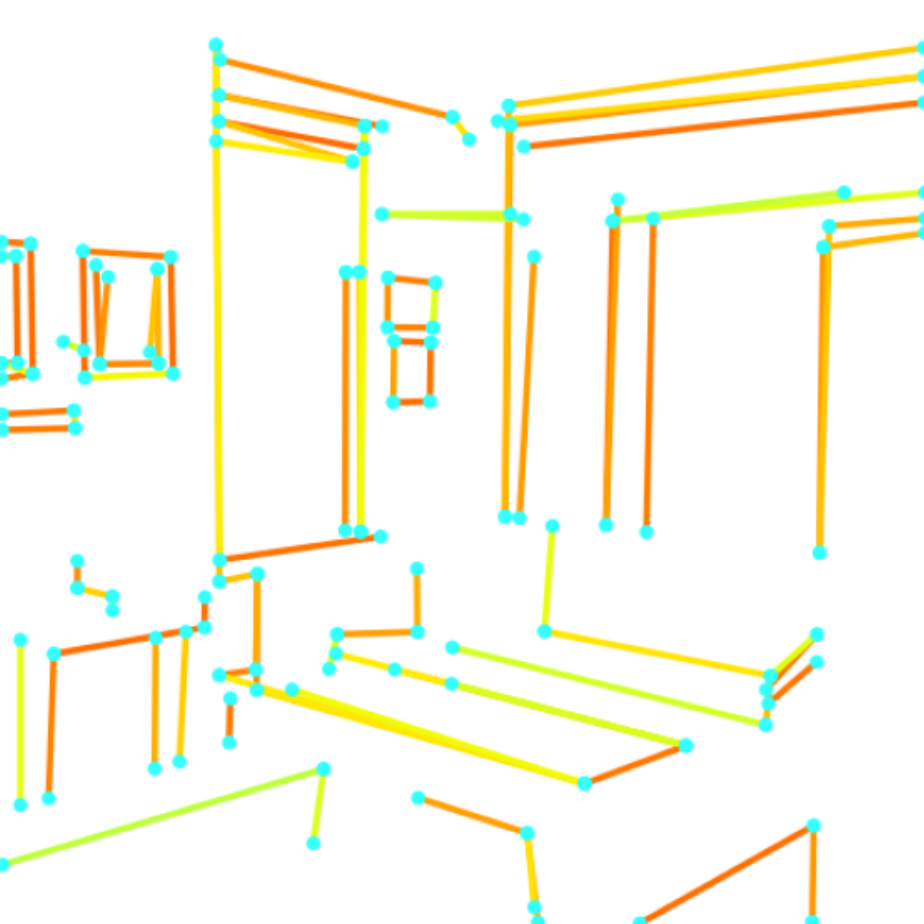}  \\
        \includegraphics[width=0.19\textwidth]{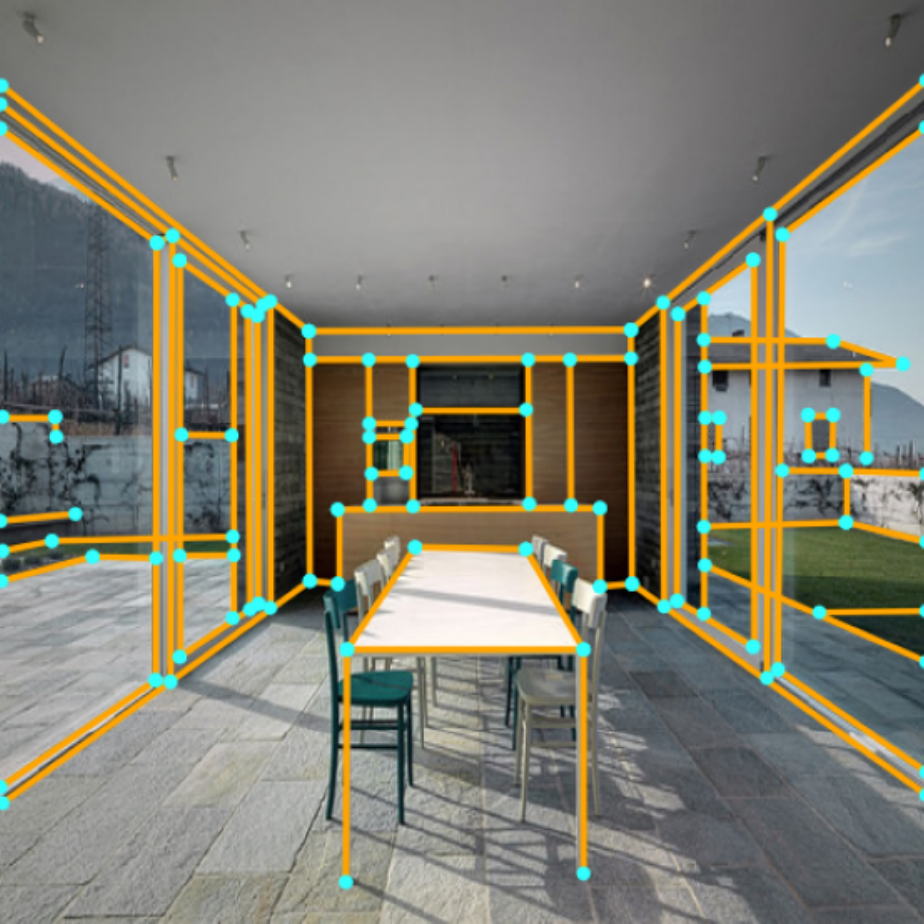} &
        \includegraphics[width=0.19\textwidth]{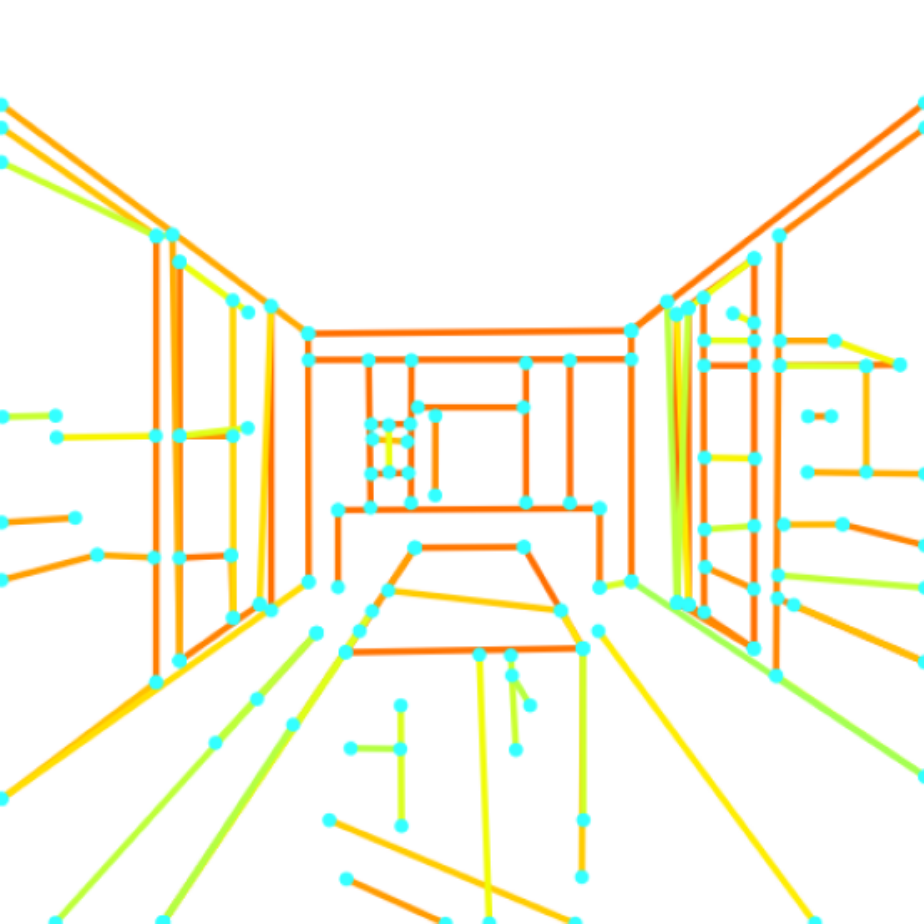} &
        \includegraphics[width=0.19\textwidth]{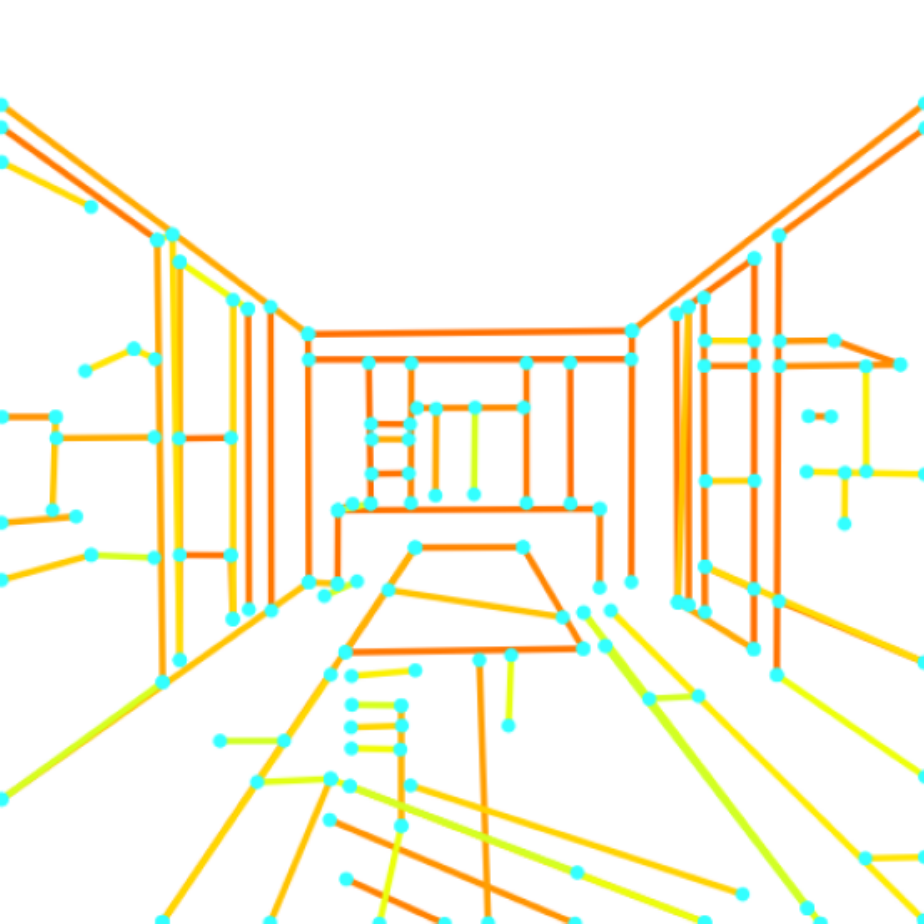} &
        \includegraphics[width=0.19\textwidth]{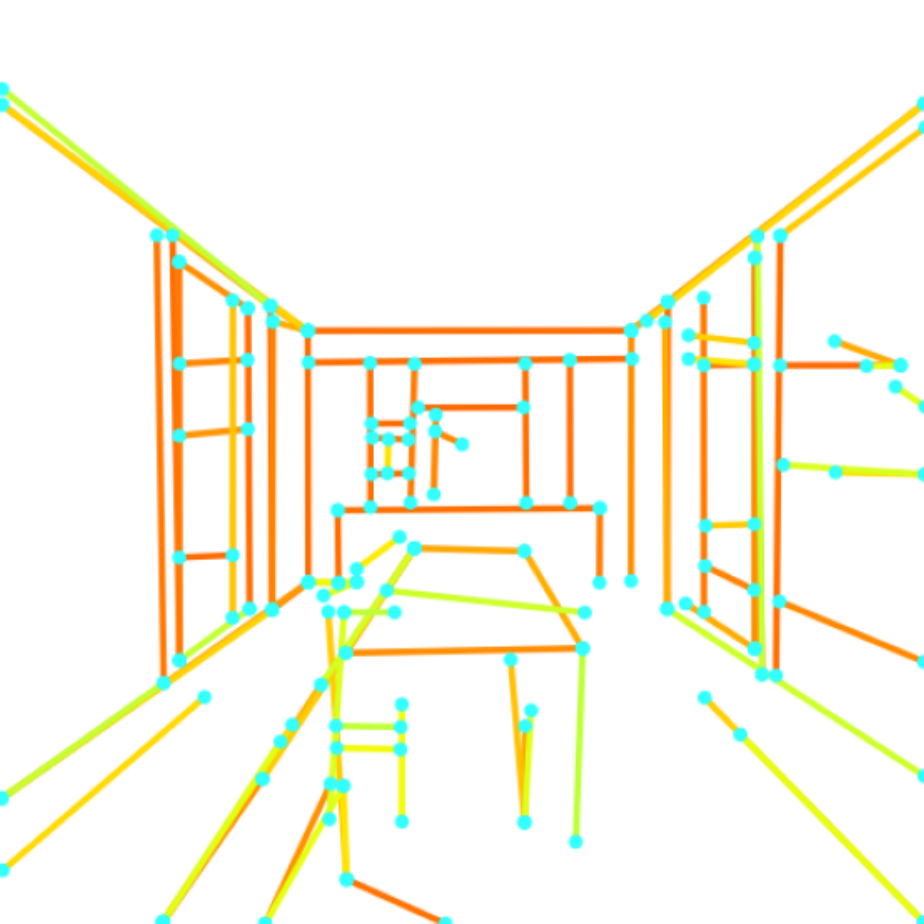} & 
        \includegraphics[width=0.19\textwidth]{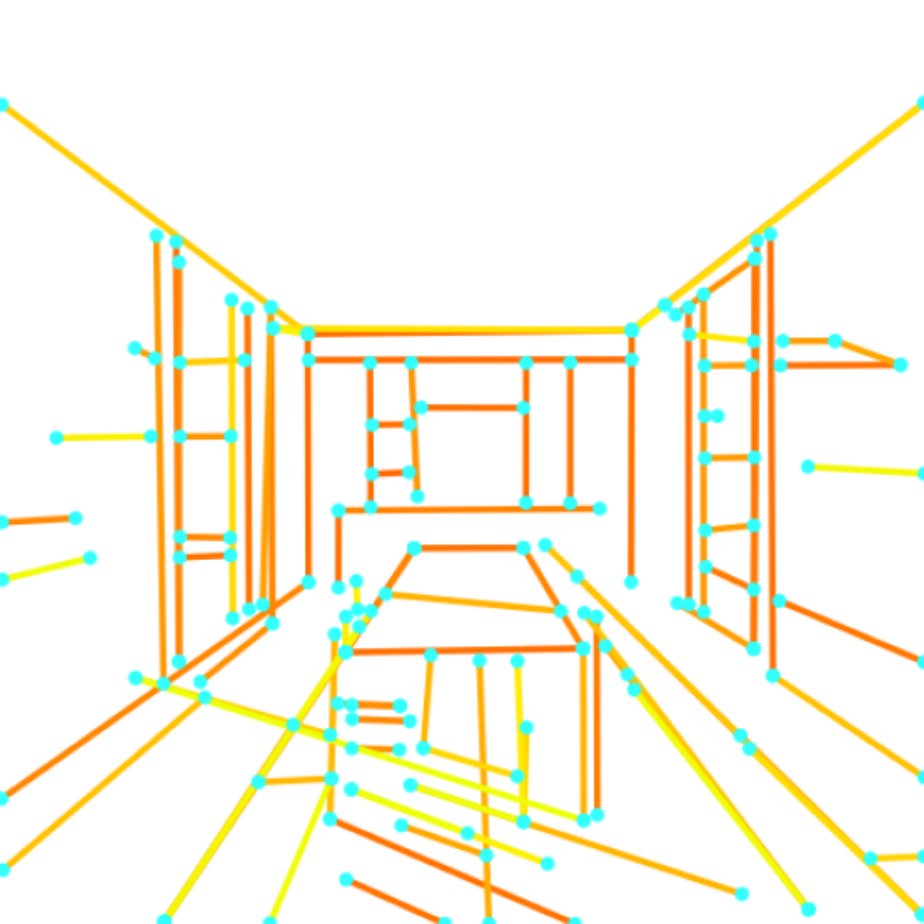} \\
        \scriptsize{Input image} & \scriptsize{LCNN (100\%)} & \scriptsize{HT-LCNN (100\%)} & \scriptsize{LCNN (10\%)} & \scriptsize{HT-LCNN (10\%)}\\
    \end{tabular}
    \caption{\textbf{Exp 3.(a):}  Visualization of detected wireframes on the Wireframe (ShanghaiTech) dataset, from LCNN(9.7M) and HT-LCNN(9.3M) trained on 100\% and 10\% data subsets.
    HT-LCNN can more consistently detects the wireframes, when trained on 10\% subset, compared to LCNN. (See the appendix for more results). 
    }
    \label{fig:exp3_a_vis}
\end{figure}

Figure~\ref{fig:exp3_a_vis} visualizes top 100 line-segment predictions of LCNN(9.7M) and HT-LCNN(9.3M) trained on 100\% and 10\% subsets of the Wireframe dataset. 
When comparing the LCNN and HT-LCNN in the top row, we notice that HT-LCNN is more precise, especially when training on only 10\% of the data. 
HT-LCNN detects more lines and junctions than LCNN because it identifies lines as local maxima in the Hough space. HT-LCNN relies less on contextual information, and thus it predicts all possible lines as wireframes (e.g. shadows of objects in the third row). In comparison, L-CNN correctly ignores those line segments. Junctions benefit from more lines, as they are intersections of lines.
These results shows the added value of HT-LCNN when training on limited data. 

\subsubsection{\textbf{Exp 3.(b):} Comparison with state-of-the-art.}
We compare our HT-LCNN and HT-HAWP, starting from  LCNN \cite{zhou2019end} and HAWP \cite{xue2020holistically} and using HT-IHT blocks instead of the hourglass blocks, with five state-of-the-art models on the Wireframe (ShanghaiTech) \cite{huang2018learning} and York Urban \cite{denis2008efficient} datasets.  
The official training split of the Wireframe dataset is used for training, and we evaluate on the respective test splits of the Wireframe\slash York Urban datasets. 
We consider three methods employing knowledge-based features: LSD \cite{von2008lsd}, Linelet \cite{cho2017novel} and MCMLSD \cite{almazan2017mcmlsd}, and four learning-based methods: 
 AFM \cite{xue2019learning}, WF-Parser (Wireframe Parser) \cite{huang2018learning}, LCNN \cite{zhou2019end}, HAWP \cite{xue2020holistically}.
We use the pre-trained models provided by the authors for AFM, LCNN and HAWP, while the WF-Parser, HT-LCNN, and HT-HAWP are trained from scratch by us.

\begin{table}[t!]
    \centering
        \begin{tabular}{p{8.2em} c@{\hskip 0.1in}ccc cccc }
        \toprule
        Train\slash test & & & \multicolumn{3}{c}{Wireframe / Wireframe} & 
        \multicolumn{3}{c}{Wireframe / York Urban}\\ \cmidrule(l){4-6}\cmidrule(l){7-9}
        & & & \multicolumn{2}{c}{Structural} & \multicolumn{1}{c}{Junction} &  \multicolumn{2}{c}{Structural} & \multicolumn{1}{c}{Junction} \\ \cmidrule(l){4-5}\cmidrule(l){6-6} \cmidrule(l){7-8} \cmidrule(l){9-9}
        Metrics & \#Params & FPS & AP$^5$ & AP$^{10}$ & mAP & AP$^5$ & AP$^{10}$ & mAP \\  \midrule
        LSD \cite{von2008lsd}       & --- & 15.3  & 7.1  & 9.3  & 16.5  & 7.5  & 9.2  & 14.9  \\
        Linelet \cite{cho2017novel}  & --- & 0.04 & 8.3  & 10.9 & 17.4 & 9.0  & 10.8 & 18.2 \\
        MCMLSD \cite{almazan2017mcmlsd} & ---& 0.2 & 7.6  & 10.4 & 13.8 & 7.2  & 9.2  & 14.8  \\
        WF-Parser \cite{huang2018learning} & 31 M & 1.7 & 6.9 & 9.0 & 36.1 & 2.8 & 3.9  & 22.5\\
        AFM \cite{xue2019learning} & 43 M & 6.5 & 18.3 & 23.9 & 23.3 & 7.1 & 9.1 & 12.3\\
        LCNN \cite{zhou2019end} & 9.7 M &10.8  & 58.9 & 62.9 & 59.3 & 24.3 & 26.4 & 30.4 \\
        \emph{HT-LCNN (Our)} & 9.3 M & 7.5 & 60.3 & 64.2 & 60.6 & 25.7 & 28.0 & \textbf{32.5}\\
        HAWP \cite{xue2020holistically}  & 10.3 M & 13.6 & 62.5 & 66.5 & 60.2 & \textbf{26.1} & \textbf{28.5} & 31.6\\
        \emph{HT-HAWP (Our)} & 10.5 M &12.2 & \textbf{62.9} & \textbf{66.6} & \textbf{61.1} & 25.0 & 27.4 & 31.5\\
        \bottomrule
        \end{tabular} 
    \caption{\textbf{Exp 3.(b):} 
    Comparing state-of-the-art line detection methods on the Wireframe (ShanghaiTech) and York Urban datasets.
    We report the number of parameters and FPS timing for every method. 
    Our HT-LCNN and HT-HAWP using HT-IHT blocks, show competitive performance. 
    HT-HAWP is similar to HAWP on the Wireframe dataset, while being less precise on the York Urban dataset. When compared to LCNN, our HT-LCNN consistently outperforms the baseline, illustrating the added value of the Hough priors. 
    }
    \label{tab:exp3_b}
\end{table}

Table~\ref{tab:exp3_b} compares structural-$AP^5$, -$AP^{10}$ and junction-mAP for seven state-of-the-art methods.
We report the number of parameters for the learning-based models as well as the frames per second (FPS) measured by using a single CPU thread or a single GPU (GTX 1080 Ti) over the test set.
Our models using the \model outperform existing methods on the Wireframe dataset, and show rivaling performance on the York Urban dataset.
HT-HAWP performs similar to HAWP on the Wireframe dataset while being less competitive on the York Urban dataset. 
HAWP uses a proposal refinement module, which further removes unmatched line proposals. 
This dampens the advantage of our \model.
Given that the York Urban dataset is not fully annotated, this may negatively affect the performance of our \model.
However, adding \model improves the performance of HT-LCNN over LCNN on both datasets, which shows the added value of the geometric line priors. 
Moreover, HAWP and LCNN perform well when ample training data is available. 
When limiting the training data, their performances decrease by a large margin compared with our models, as exposed in \textbf{Exp 3.(a)}.

\begin{figure}[t!]
    \centering
    \begin{tabular}{ccc}
        \includegraphics[width=0.5\textwidth]{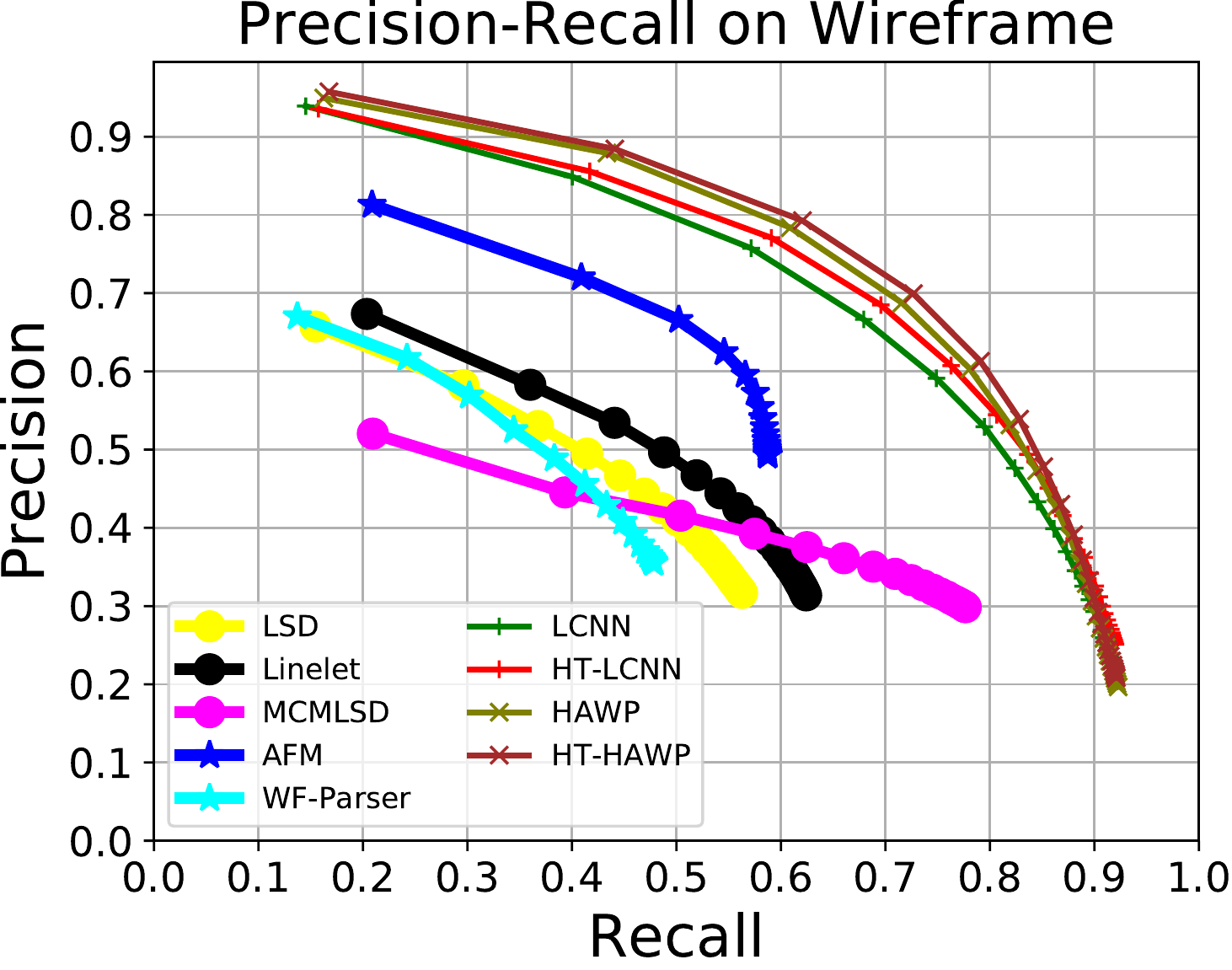} &
        \includegraphics[width=0.5\textwidth]{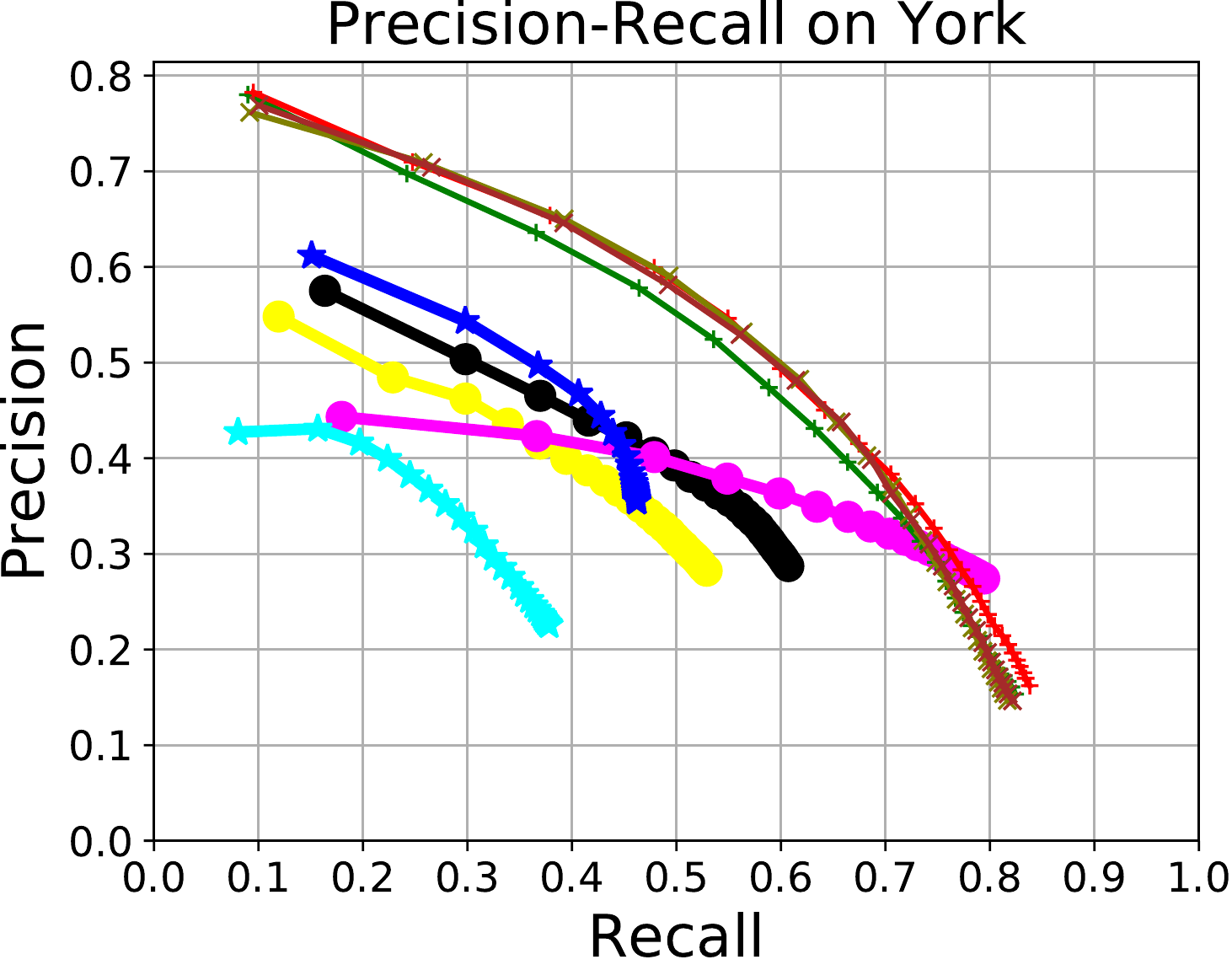} \\
        \scriptsize{(a) Precision-recall on Wireframe (ShanghaiTech)} & 
        \scriptsize{(b) Precision-recall on York Urban}\\
    \end{tabular}
    \caption{\textbf{Exp 3.(b):} 
    Comparing our HT-LCNN and HT-HAWP with seven existing methods using precision-recall scores on the Wireframe (ShanghaiTech) and York Urban datasets.
    Traditional knowledge-based methods are outperformed by deep learning methods.
    Among the learning-based methods, our proposed HT-LCNN and HT-HAWP achieve state-of-the-art performance, even in the full-data regime.}
    \label{fig:exp3_b}
\end{figure}

Figure~\ref{fig:exp3_b} shows precision-recall scores \cite{almazan2017mcmlsd} on the Wireframe (ShanghaiTech) and York Urban datasets.
MCMLSD \cite{almazan2017mcmlsd} shows good performance in the high-recall zone on the York Urban dataset, but its performance is lacking in the low-recall zone. 
AFM \cite{xue2019learning} predicts a limited number of line segments, and thus it lacks in the high-recall zone. 
One advantage of (HT-)LCNN and (HT-)HAWP over other models such as AFM, is their performance in the high-recall zone, indicating that they can detect more ground truth line segments. 
However, they predict more overlapping line segments due to co-linear junctions, which results in a rapid decrease in precision. 
Our proposed HT-LCNN and HT-HAWP show competitive performance when compared to state-of-the-art models, thus validating the usefulness of the \model. 

\begin{figure}[t!]
    \centering
    \begin{tabular}{ccc}
        \includegraphics[width=0.5\textwidth]{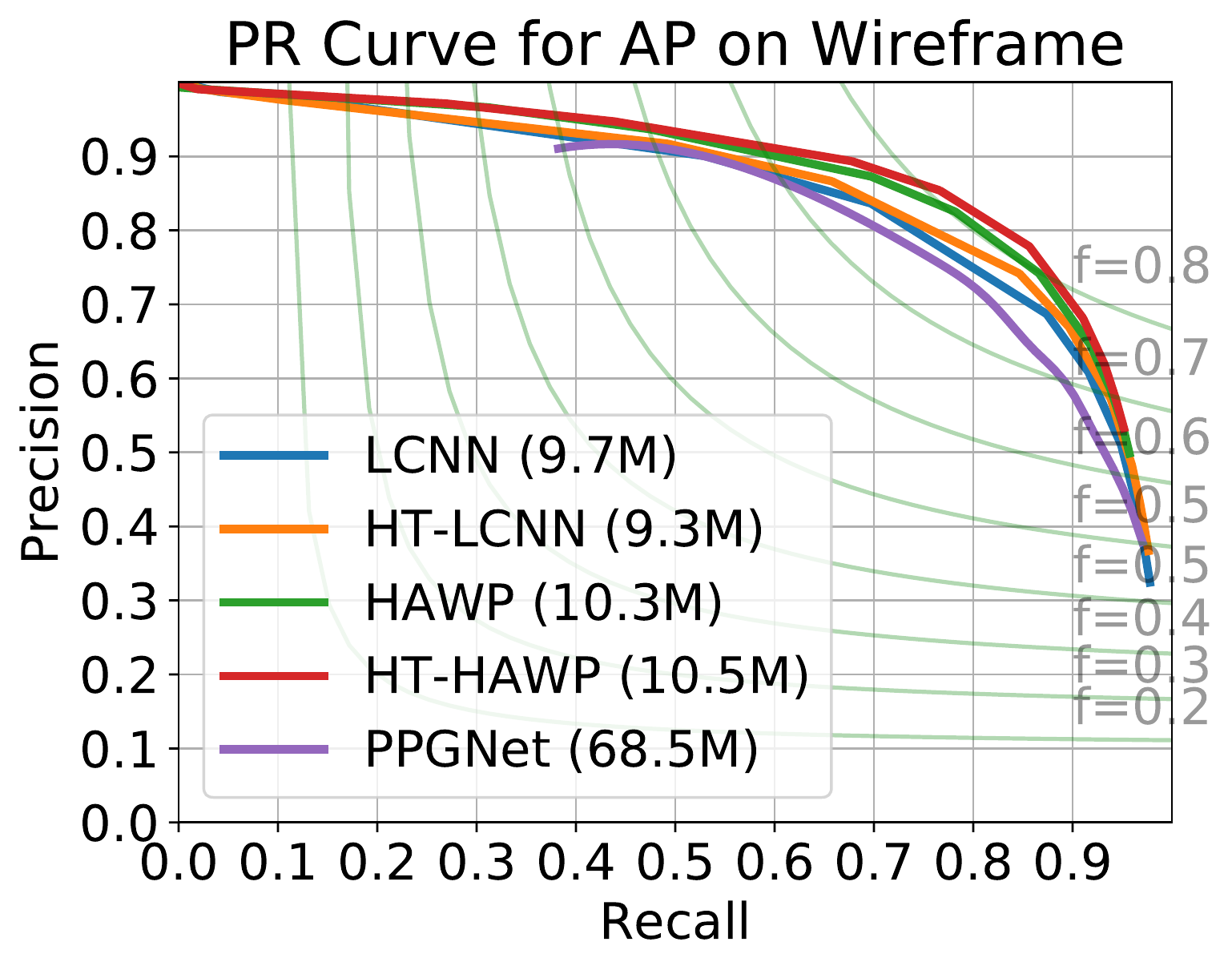} &
        \includegraphics[width=0.5\textwidth]{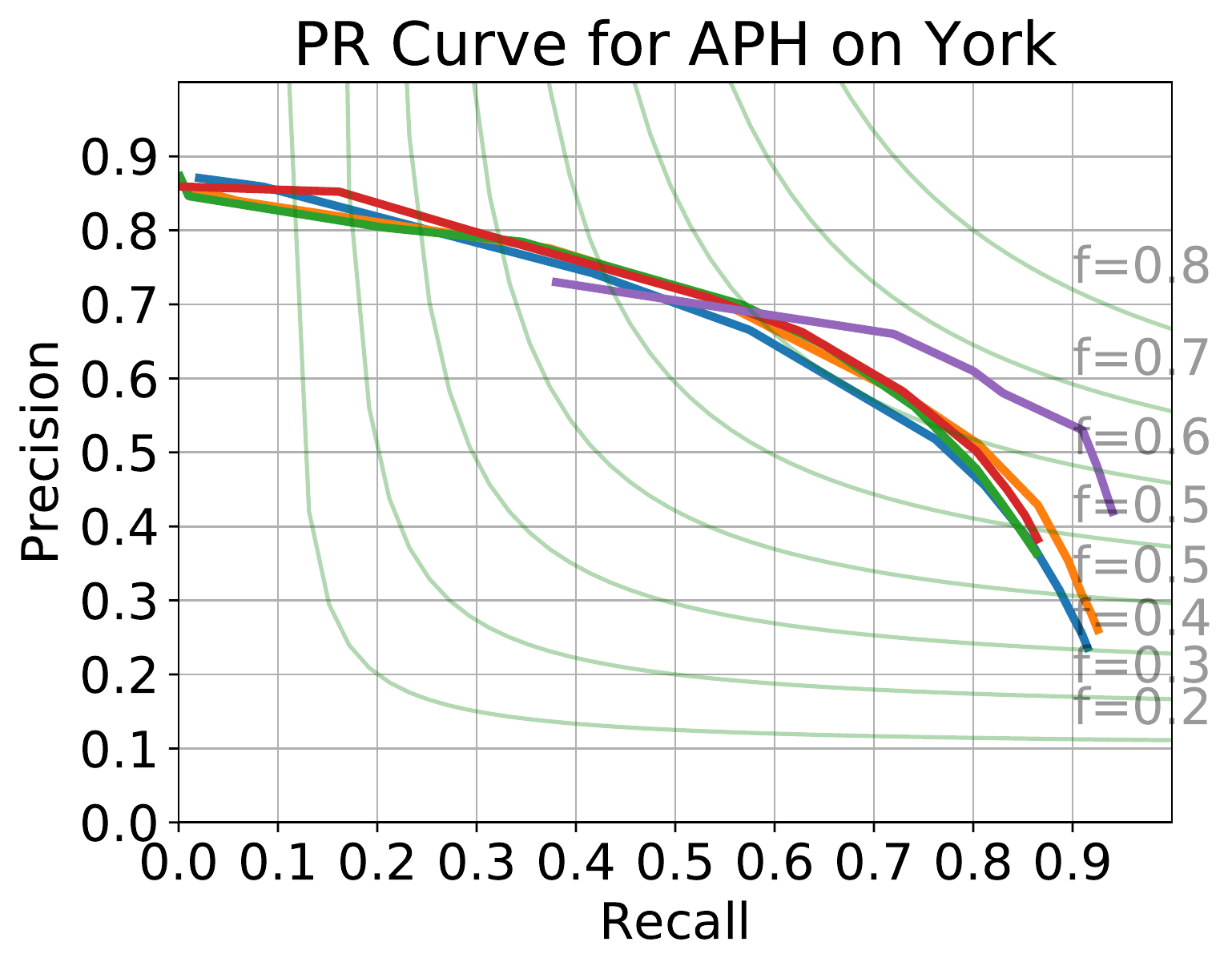} \\
        \scriptsize{(a) AP on Wireframe (ShanghaiTech)} & 
        \scriptsize{(b) AP on York Urban}\\
    \end{tabular}
    \caption{
    \textbf{Exp 3.(b):} Comparing PPGNet\cite{zhang2019ppgnet} with (HT-)LCNN and (HT-)HAWP on the Wireframe (ShanghaiTech) and York Urban datasets. 
    PPGNet shows better performance on the York Urban dataset, especially in high-recall region, while
    being slightly less precise on the Wireframe dataset when compared to our HT-LCNN and HT-HAWP methods. 
    We show between brackets the number of parameters.}
    \label{fig:exp3_ppg}
\end{figure}

In figure \ref{fig:exp3_ppg}, we compare our HT-LCNN and HT-HAWP with PPGNet \cite{zhang2019ppgnet}. 
The PPGNet result is estimated from the original paper, since we are not able to replicate the results using the author\rq s code \footnote{\url{https://github.com/svip-lab/PPGNet}}. 
We follow the same protocol as PPGNet to evaluate (HT-)LCNN and (HT-)HAWP. 
In general, PPGNet shows superior performance on the York Urban dataset, especially in the high-recall region, while using a lot more parameters.
However, our HT-LCNN and HT-HAWP methods are slightly more precise on the Wireframe dataset.

%% file: conclusion.tex
\section{Conclusion}
We propose adding geometric priors based on Hough transform, for improved data efficiency.
The Hough transform priors are added end-to-end in a deep network, where we 
detail the forward and backward passes of our proposed \model.
We additionally introduce the use of convolutions in the Hough domain, which are effective at retaining only the line information. 
We demonstrate experimentally on a toy Line-Circle dataset that our $\mathcal{HT}$ (Hough transform) and $\mathcal{IHT}$ (inverse Hough transform) layers, inside the \model, help detect lines by combining local and global image information.
Furthermore, we validate on the Wireframe (ShanghaiTech) and York Urban datasets that the Hough line priors, included in our \model, are effective when reducing the training data size. 
Finally, we show that our proposed approach achieves competitive performance when compared to state-of-the-art methods.

%% file: appendix.tex
\appendix
\section*{\Large\textbf{Appendix}}

\section{\textbf{Exp 1:} Qualitative results on the Line-Circle dataset}

\begin{figure}
    \centering
    \begin{tabular}{ccccc}
        \includegraphics[width=0.19\textwidth]{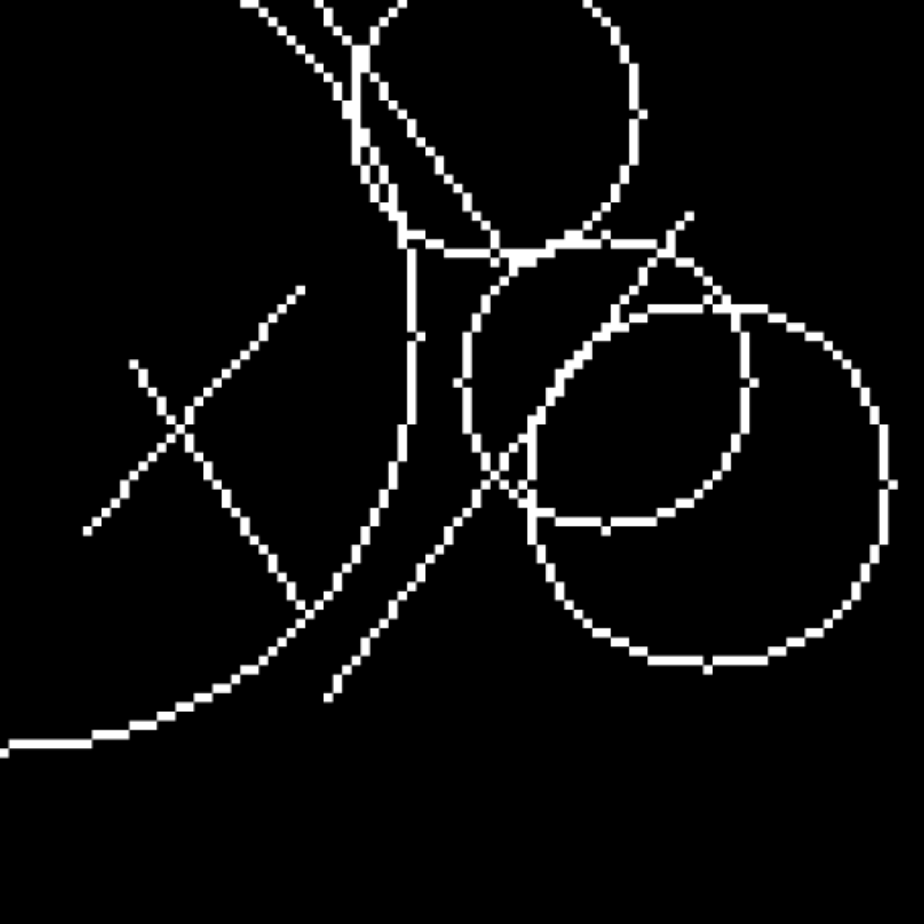} &
        \includegraphics[width=0.19\textwidth]{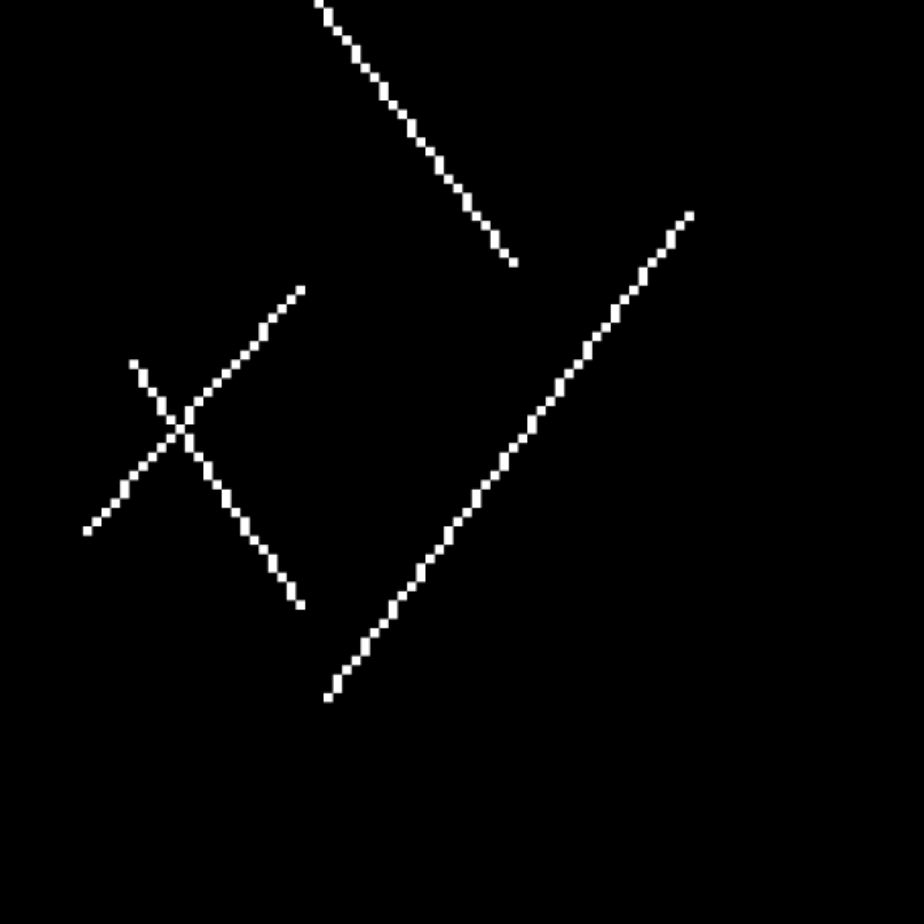} &
        \includegraphics[width=0.19\textwidth]{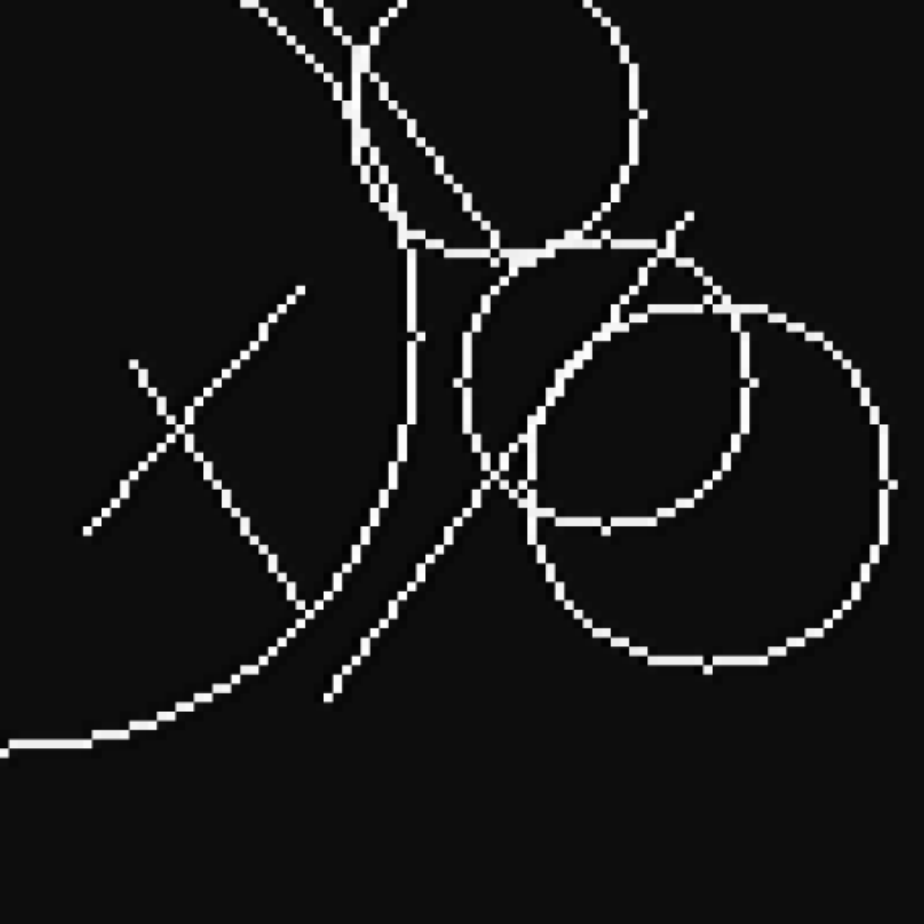} &
        \includegraphics[width=0.19\textwidth]{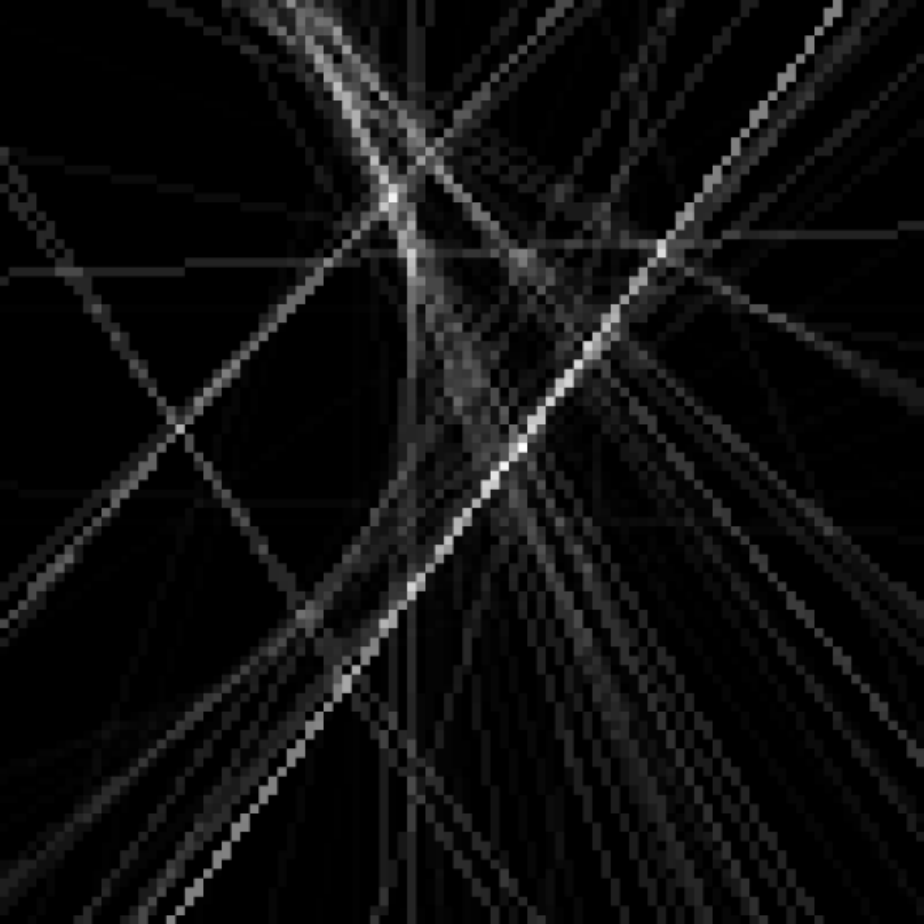} &
        \includegraphics[width=0.19\textwidth]{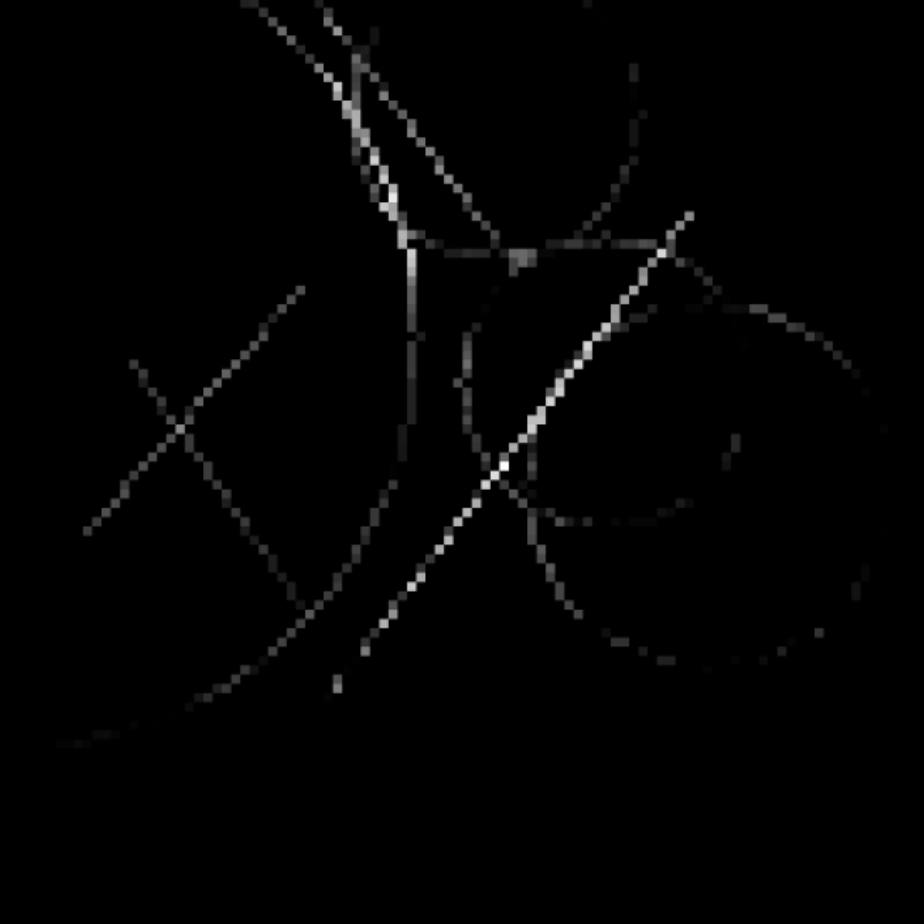}  \\ 
        \scriptsize{Input} & \scriptsize{Ground truth} & \scriptsize{Local-only} & \scriptsize{Global-only} &\scriptsize{Local+global}\\
        \includegraphics[width=0.19\textwidth]{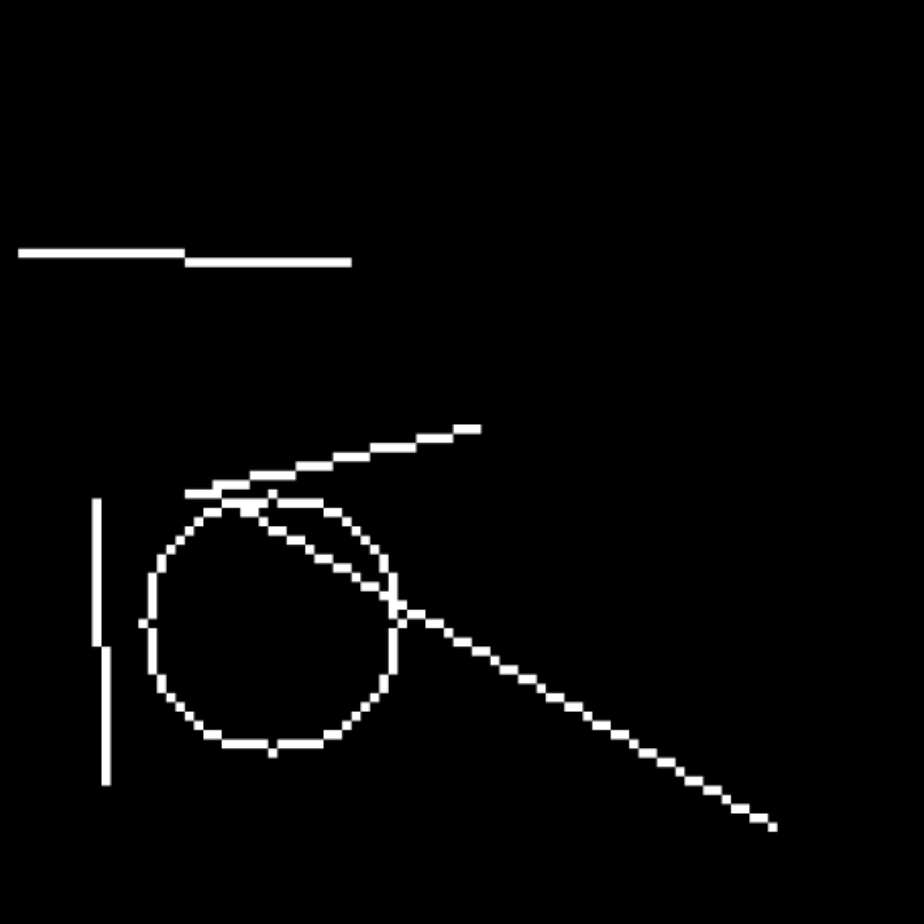} &
        \includegraphics[width=0.19\textwidth]{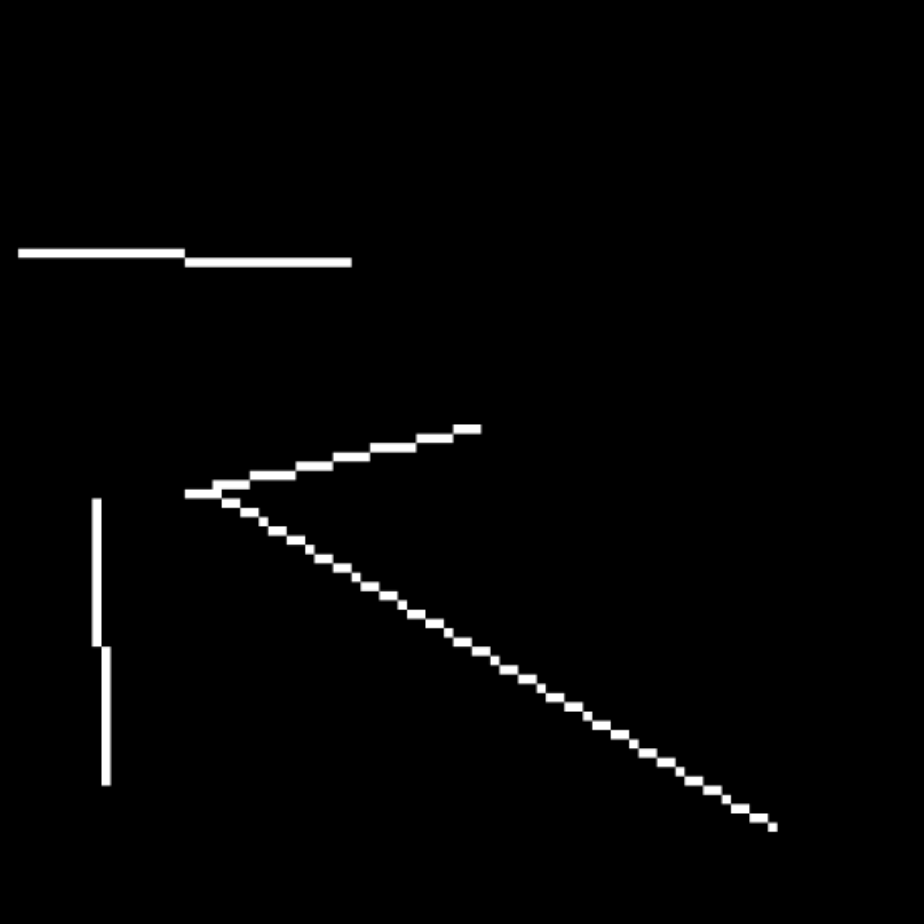} &
        \includegraphics[width=0.19\textwidth]{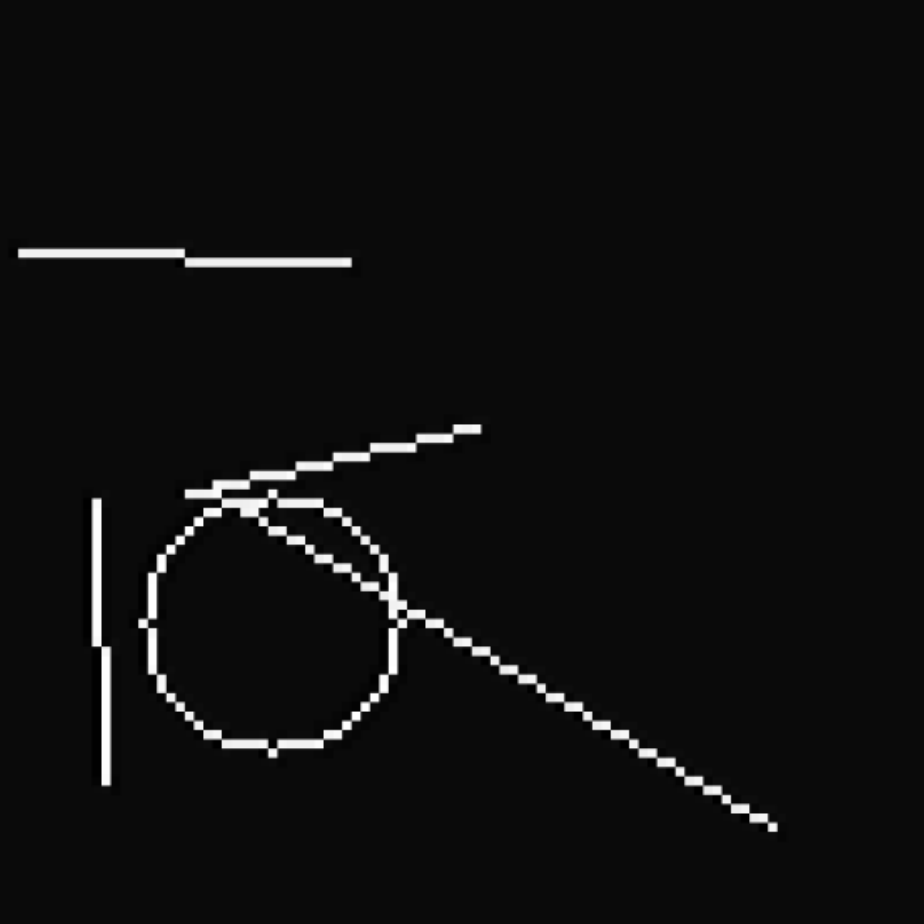} &
        \includegraphics[width=0.19\textwidth]{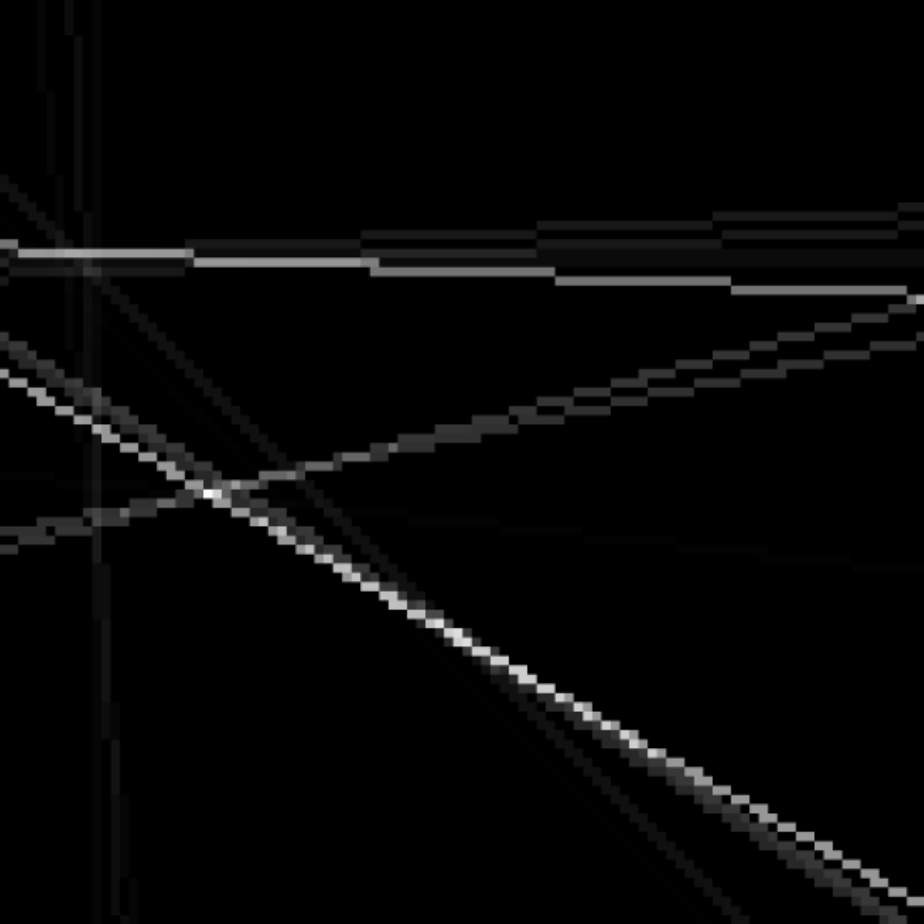} &
        \includegraphics[width=0.19\textwidth]{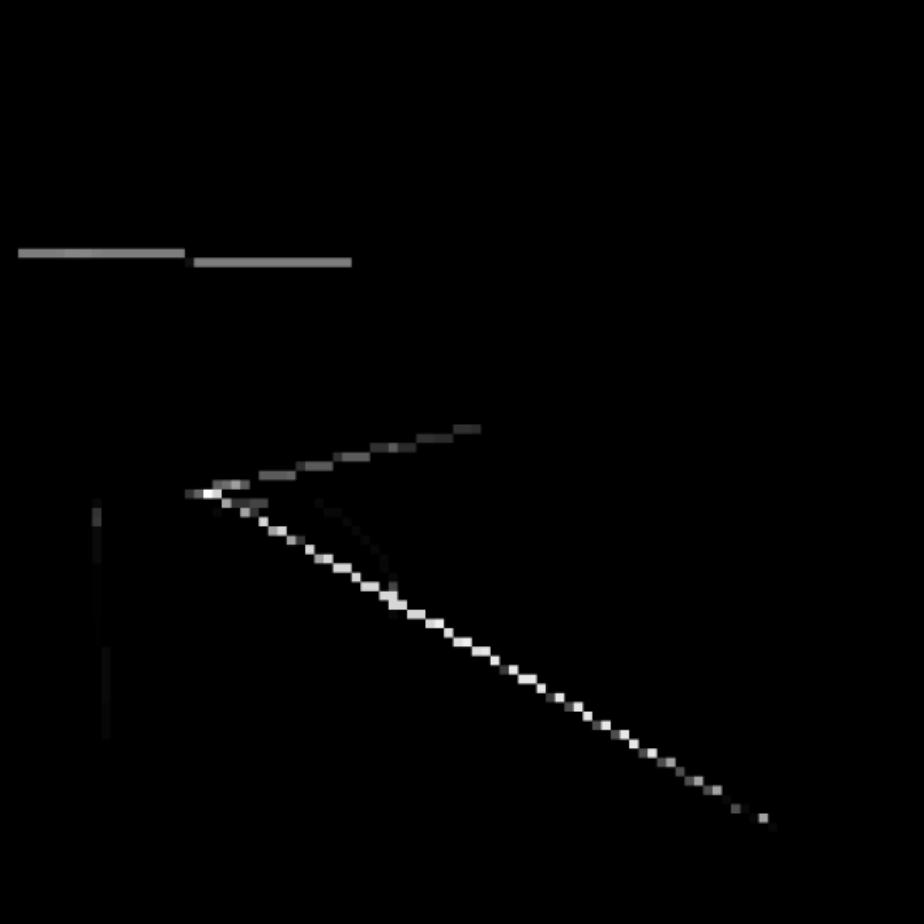}  \\ 
        \scriptsize{Input} & \scriptsize{Ground truth} & \scriptsize{Local-only} & \scriptsize{Global-only} &\scriptsize{Local+global}\\
        \includegraphics[width=0.19\textwidth]{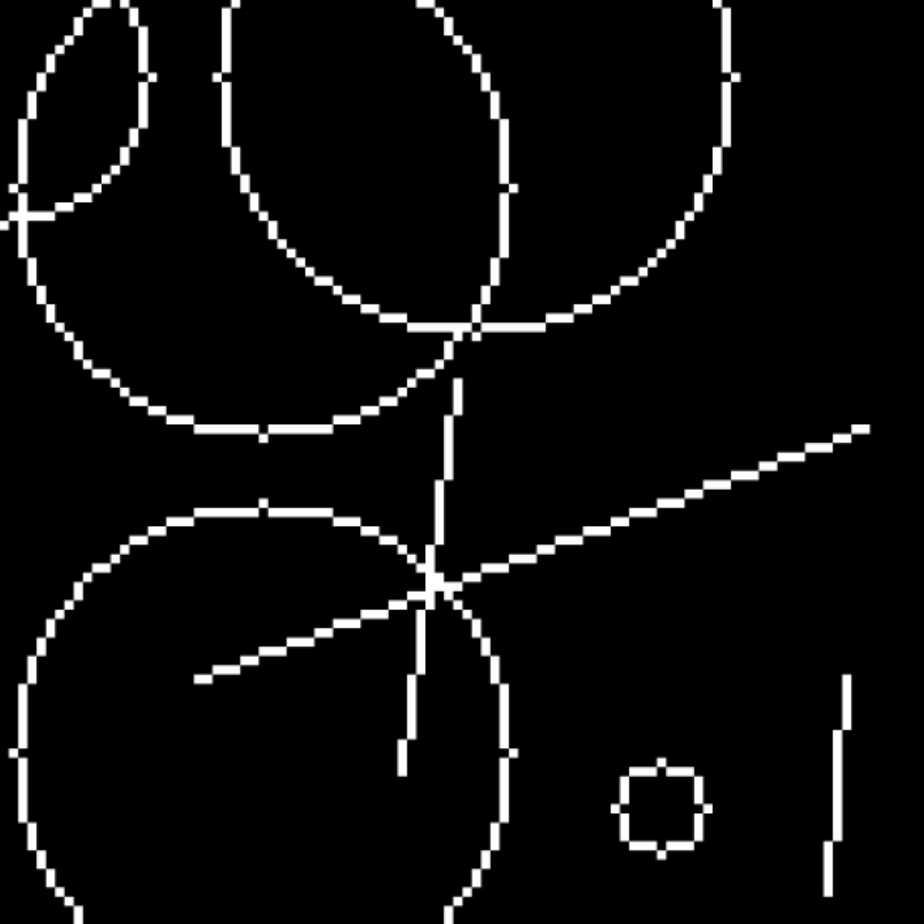} &
        \includegraphics[width=0.19\textwidth]{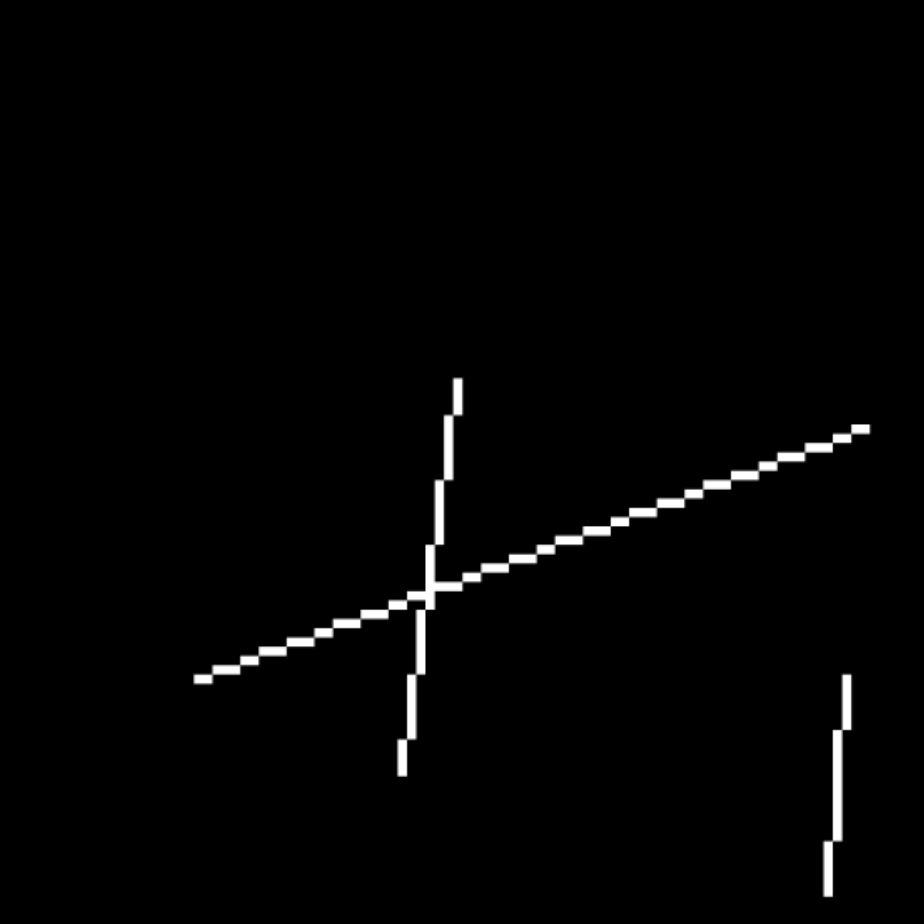} &
        \includegraphics[width=0.19\textwidth]{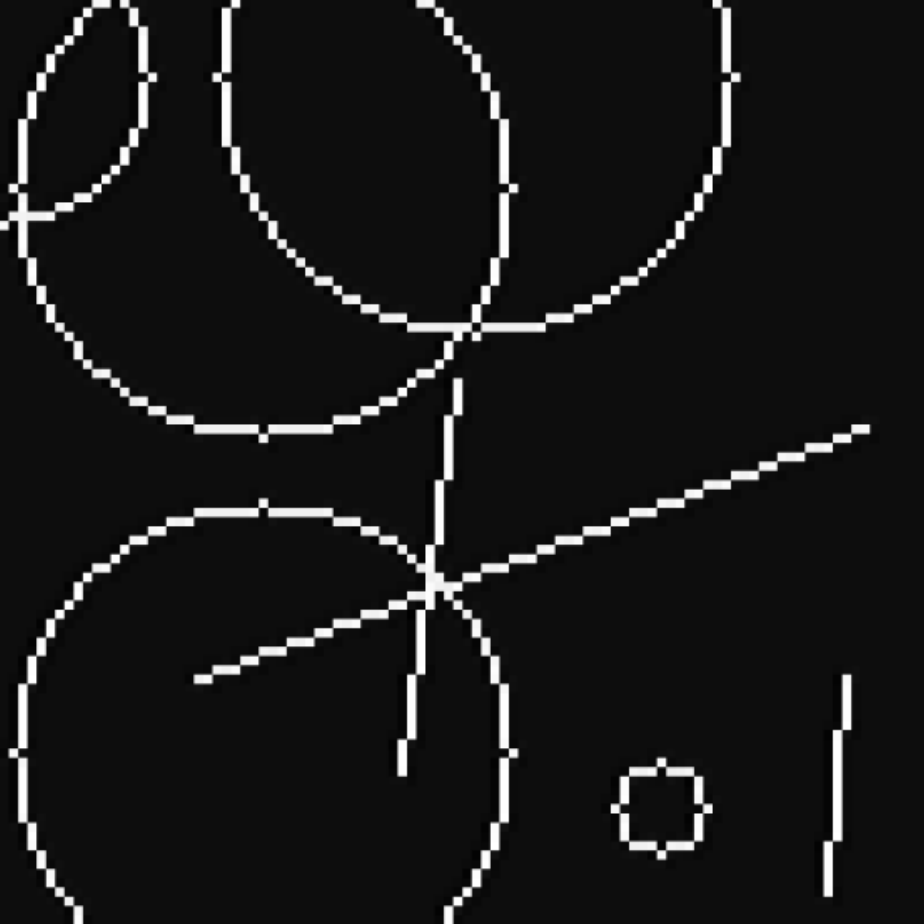} &
        \includegraphics[width=0.19\textwidth]{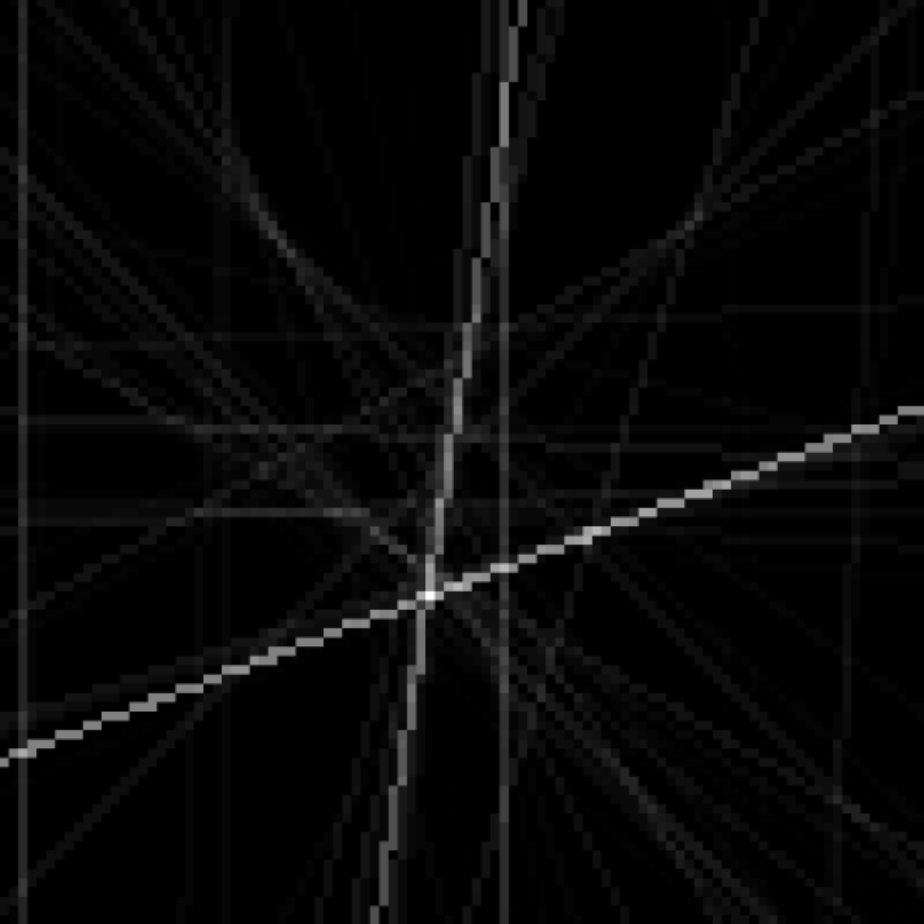} &
        \includegraphics[width=0.19\textwidth]{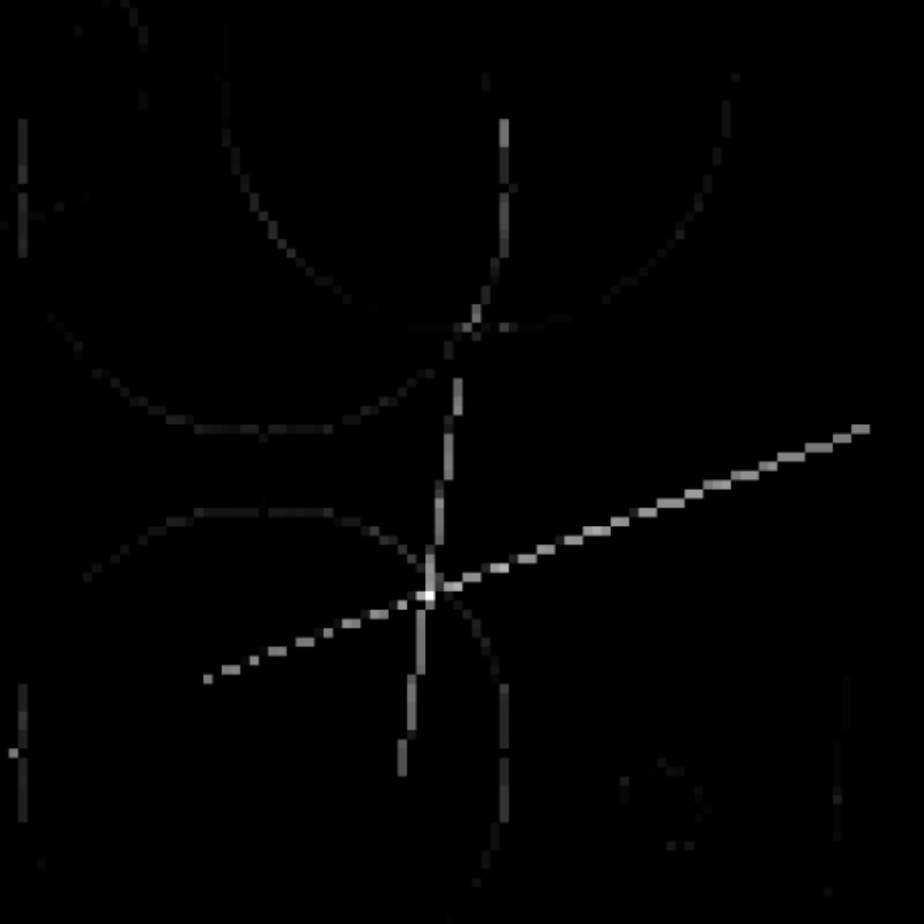}  \\ 
        \scriptsize{Input} & \scriptsize{Ground truth} & \scriptsize{Local-only} & \scriptsize{Global-only} &\scriptsize{Local+global}\\
        \includegraphics[width=0.19\textwidth]{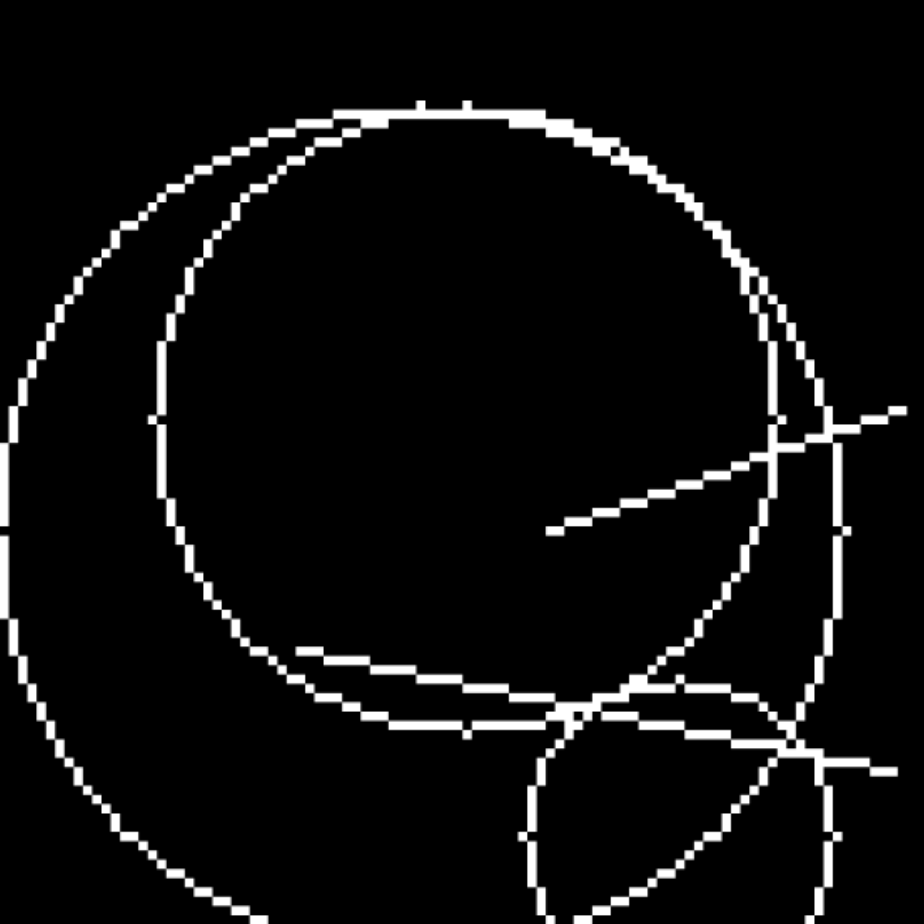} &
        \includegraphics[width=0.19\textwidth]{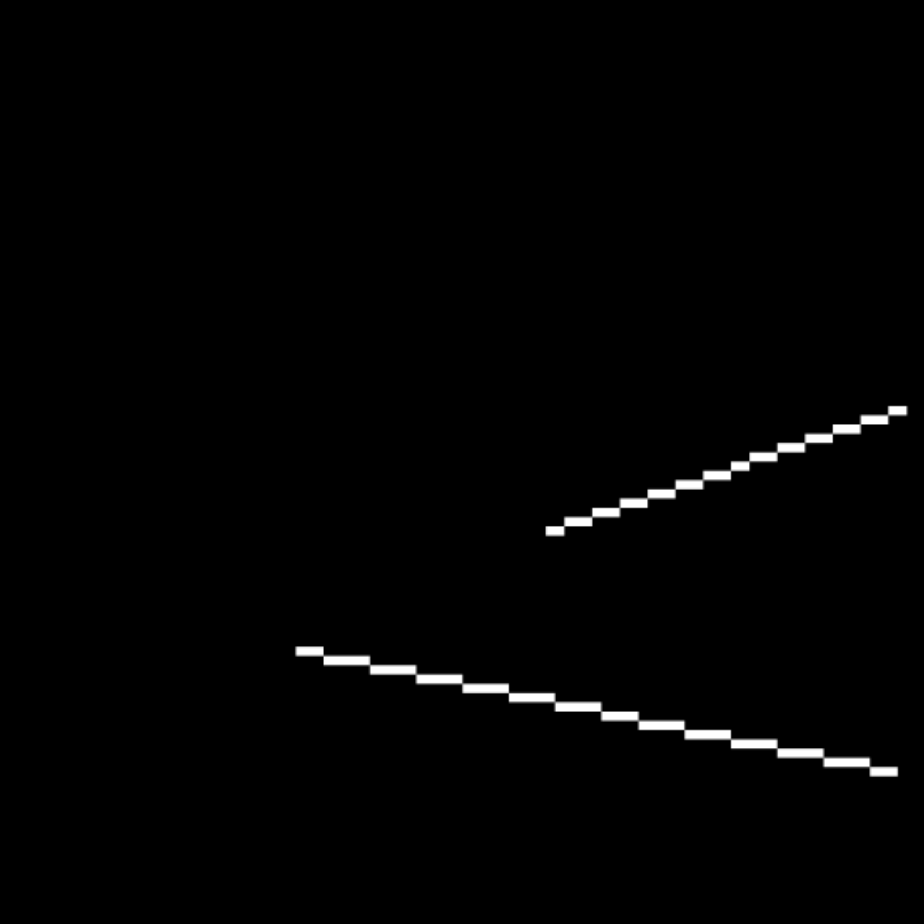} &
        \includegraphics[width=0.19\textwidth]{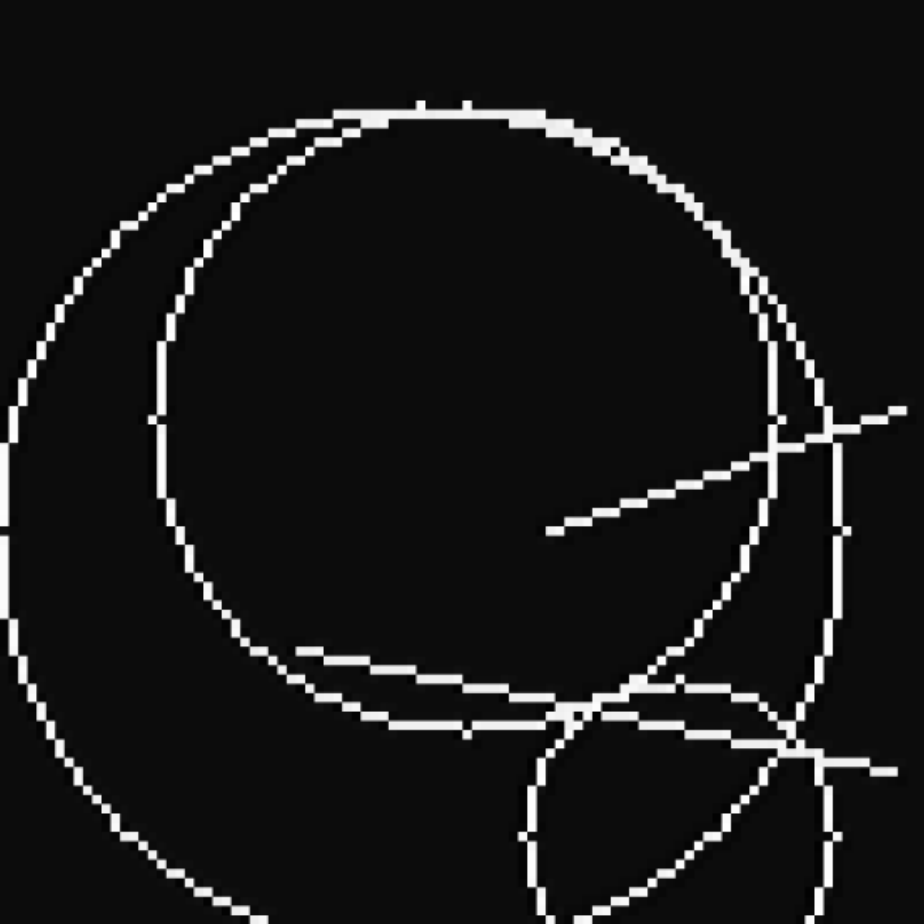} &
        \includegraphics[width=0.19\textwidth]{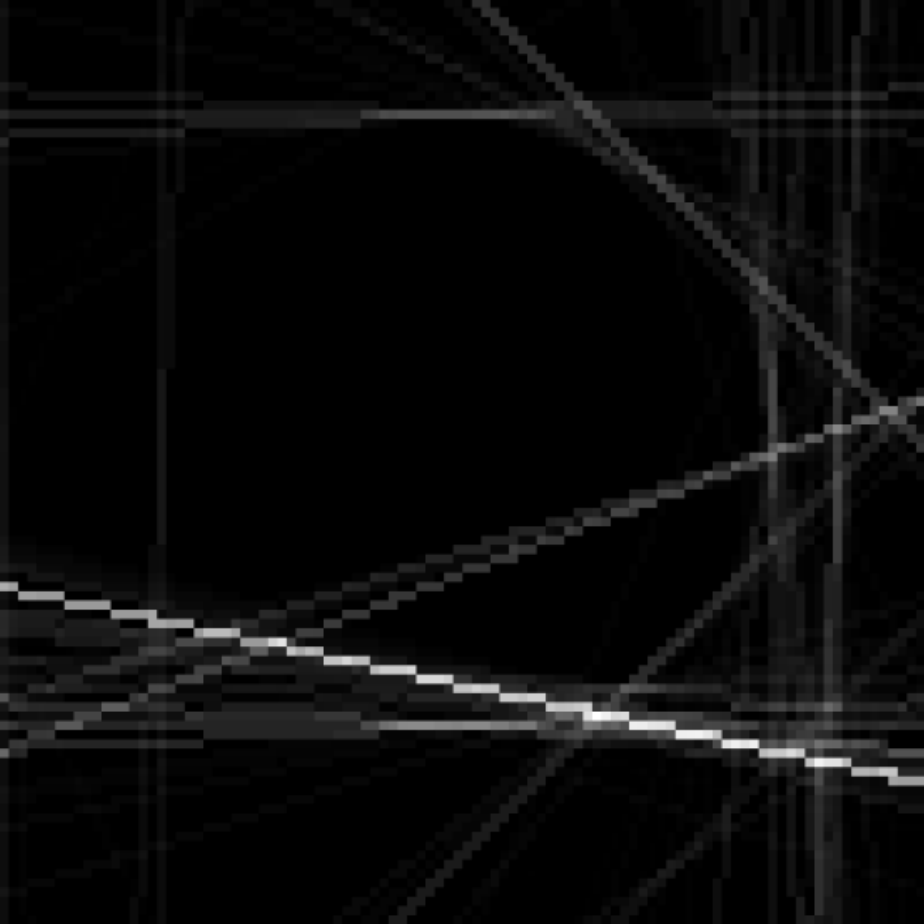} &
        \includegraphics[width=0.19\textwidth]{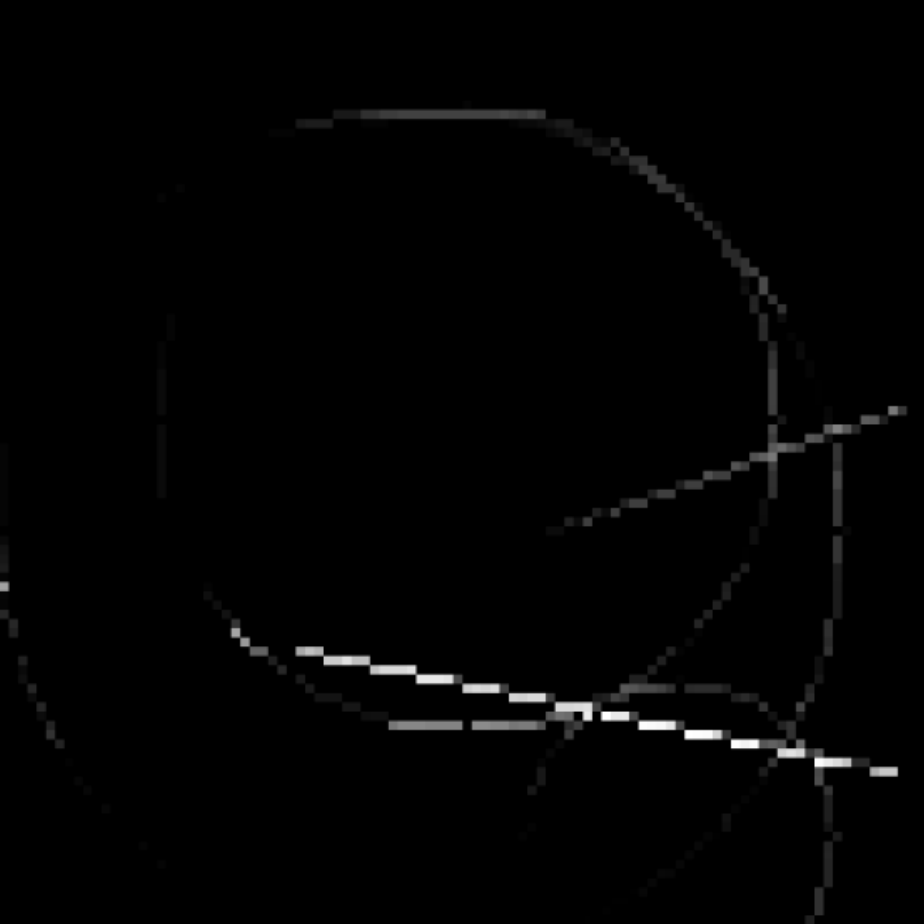}  \\ 
        \scriptsize{Input} & \scriptsize{Ground truth} & \scriptsize{Local-only} & \scriptsize{Global-only} &\scriptsize{Local+global}\\

        \includegraphics[width=0.19\textwidth]{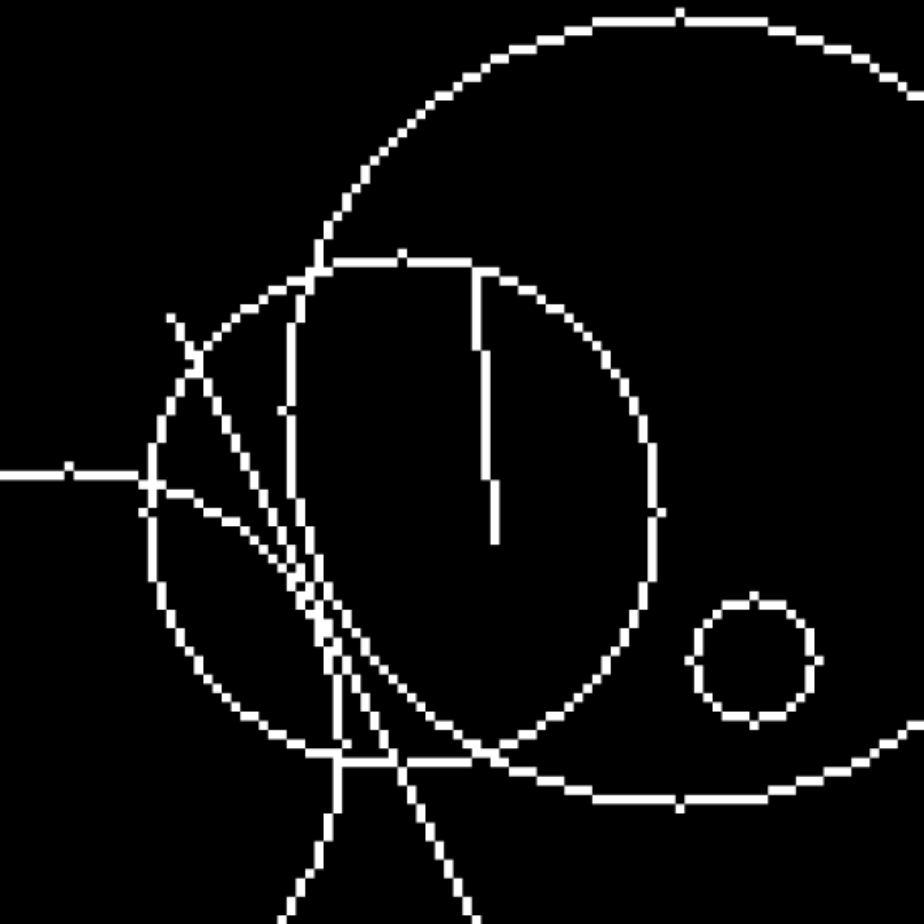} &
        \includegraphics[width=0.19\textwidth]{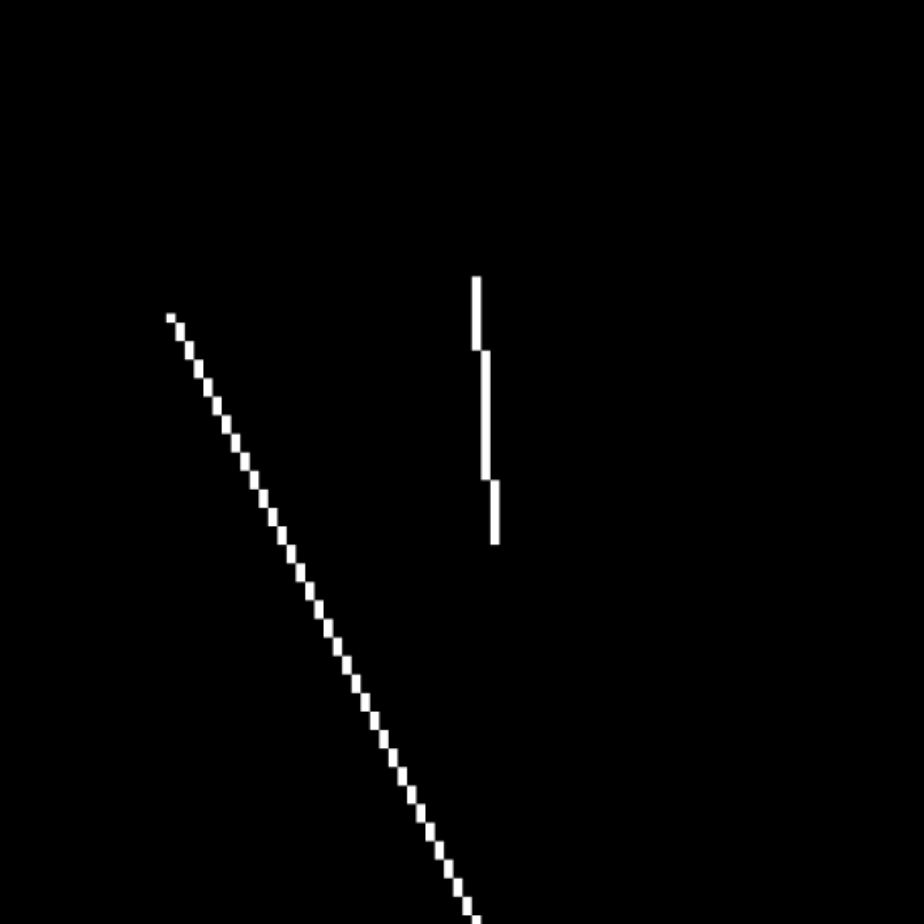} &
        \includegraphics[width=0.19\textwidth]{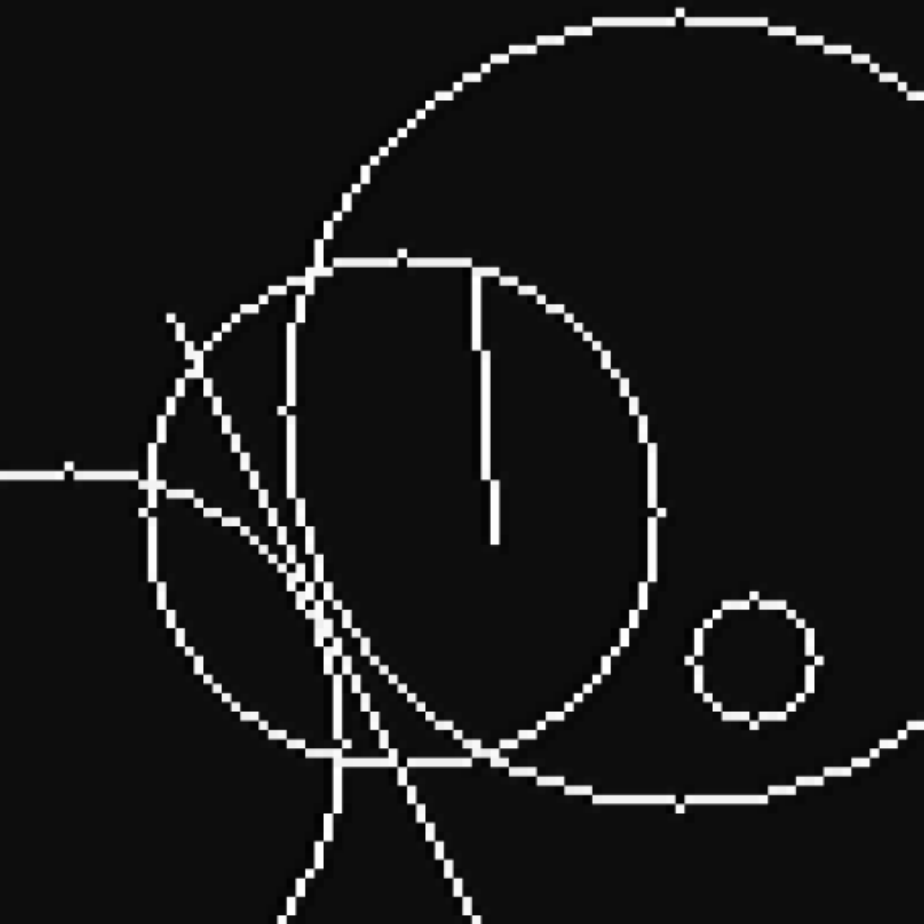} &
        \includegraphics[width=0.19\textwidth]{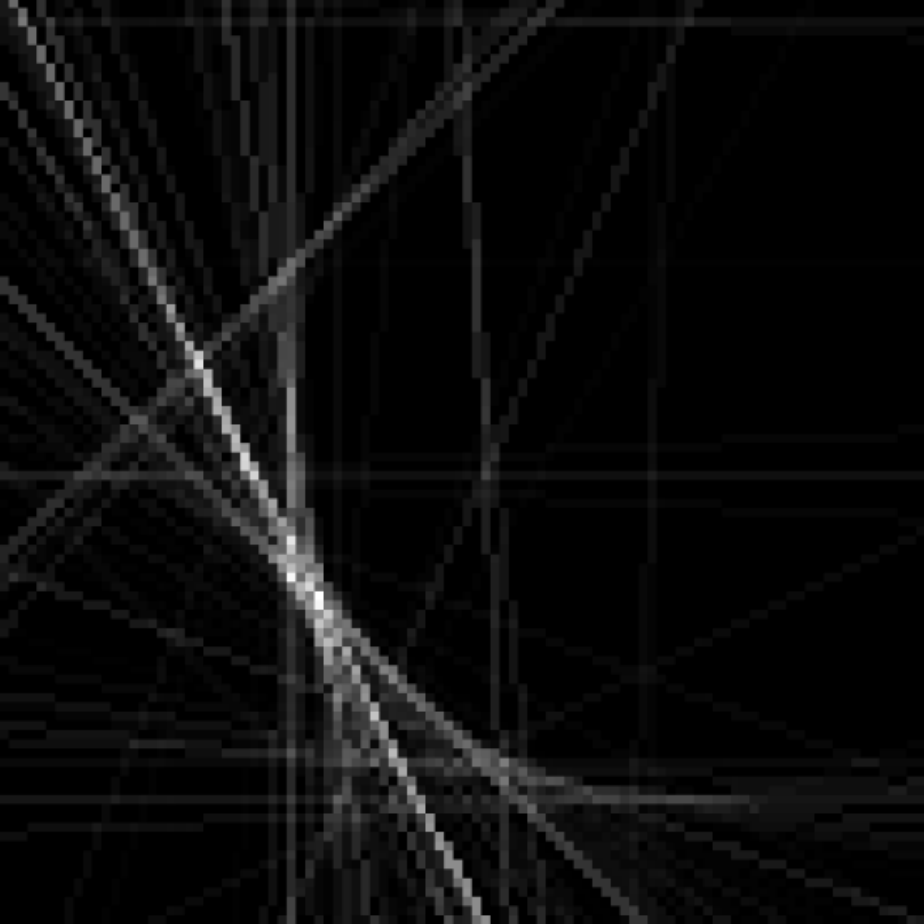} &
        \includegraphics[width=0.19\textwidth]{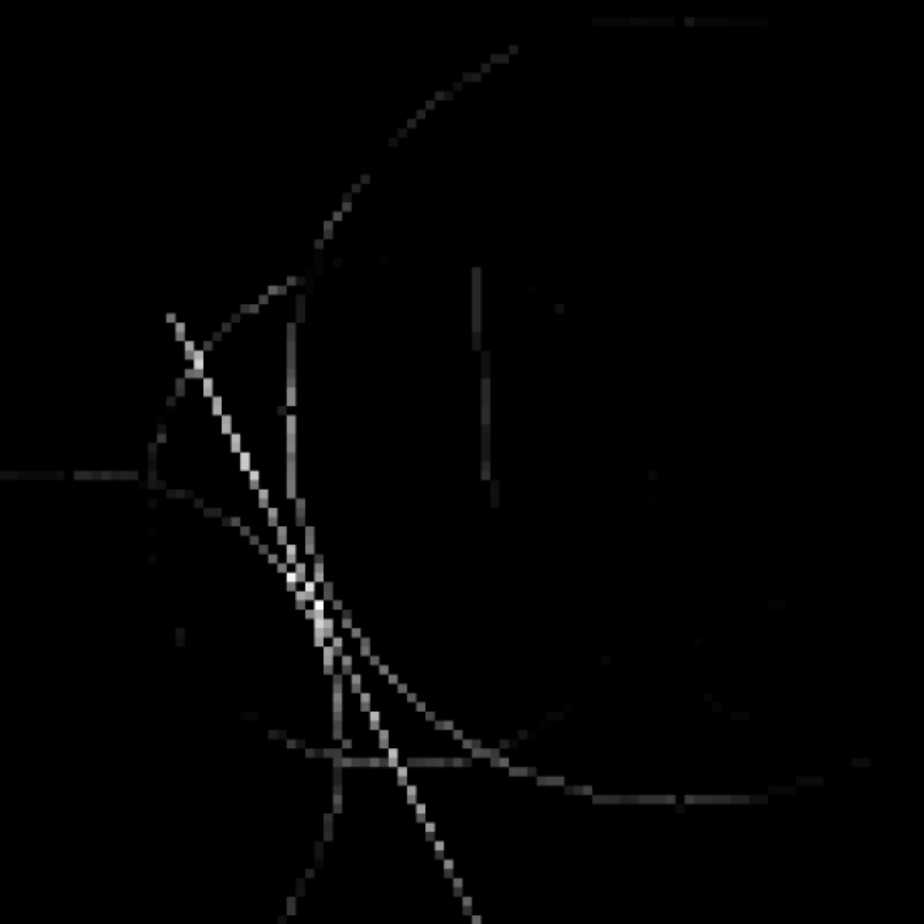}  \\ 
        \scriptsize{Input} & \scriptsize{Ground truth} & \scriptsize{Local-only} & \scriptsize{Global-only} &\scriptsize{Local+global}\\
    \end{tabular}
    \caption{\textbf{Exp 1:} Visualization of detected lines on the toy Line-Circle dataset.
    The local+global model successfully removed the circle pixels and retains the pixels along the line. 
    Combing local and global information detects not only the direction of the lines but also their extent.
    }
   \label{fig:sup_line_circle}
\end{figure}

Figure~\ref{fig:sup_line_circle} visualizes detected lines on the Line-Circle dataset from the local-only, global-only and local+global models. 
Using the global information learned by our \model combined with the local information provided by the convolutional layers, we propose a local+global approach that can predict both the direction of the lines and their extent.

\section{\textbf{Exp 3.(a):} Qualitative results using Wireframe subsets}
\begin{figure}
    \centering
    \begin{tabular}{ccccc}
        \includegraphics[width=0.19\textwidth]{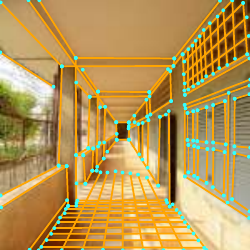} &
        \includegraphics[width=0.19\textwidth]{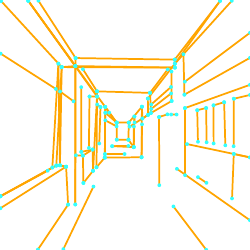} &
        \includegraphics[width=0.19\textwidth]{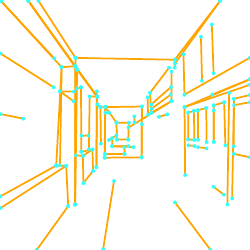} &
        \includegraphics[width=0.19\textwidth]{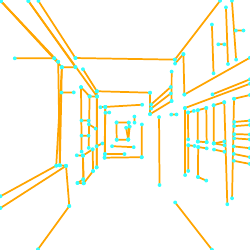} &
        \includegraphics[width=0.19\textwidth]{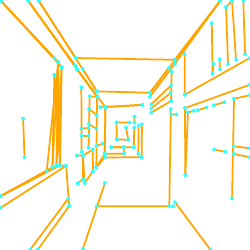} \\ 
        \scriptsize{Ground truth} & \scriptsize{LCNN(100\%)} & \scriptsize{LCNN(50\%)}  & \scriptsize{LCNN(25\%)} &\scriptsize{LCNN(10\%)}\\
        \includegraphics[width=0.19\textwidth]{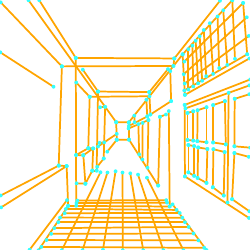} &
        \includegraphics[width=0.19\textwidth]{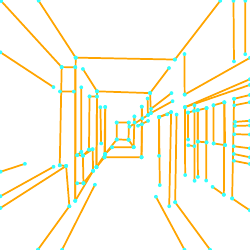} &
        \includegraphics[width=0.19\textwidth]{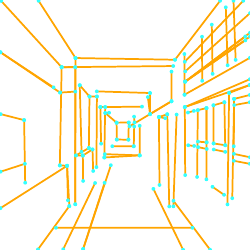} &
        \includegraphics[width=0.19\textwidth]{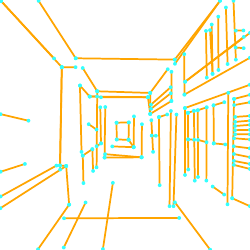} &
        \includegraphics[width=0.19\textwidth]{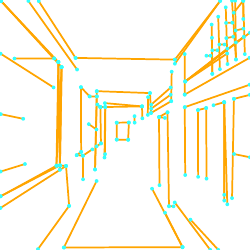} \\ 
        \scriptsize{Ground truth} & \scriptsize{HT-LCNN(100\%)} & \scriptsize{HT-LCNN(50\%)}  & \scriptsize{HT-LCNN(25\%)} &\scriptsize{HT-LCNN(10\%)}\\
        \includegraphics[width=0.19\textwidth]{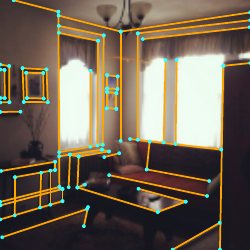} &
        \includegraphics[width=0.19\textwidth]{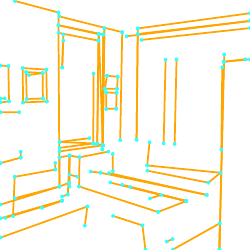} &
        \includegraphics[width=0.19\textwidth]{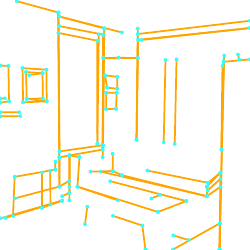} &
        \includegraphics[width=0.19\textwidth]{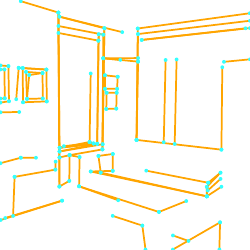} &
        \includegraphics[width=0.19\textwidth]{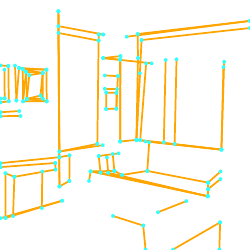}  \\ 
       \scriptsize{Ground truth} & \scriptsize{LCNN(100\%)} & \scriptsize{LCNN(50\%)}  & \scriptsize{LCNN(25\%)} &\scriptsize{LCNN(10\%)}\\
        \includegraphics[width=0.19\textwidth]{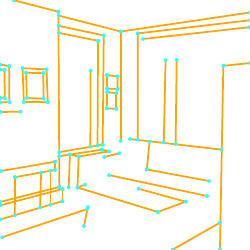} &
        \includegraphics[width=0.19\textwidth]{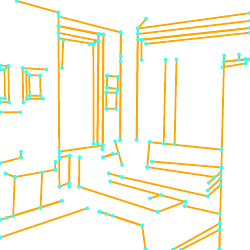} &
        \includegraphics[width=0.19\textwidth]{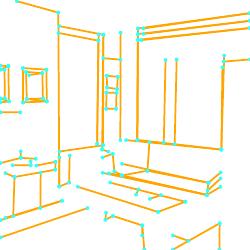} &
        \includegraphics[width=0.19\textwidth]{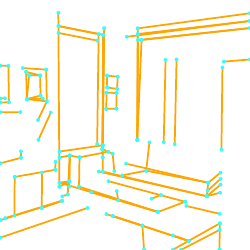} &
        \includegraphics[width=0.19\textwidth]{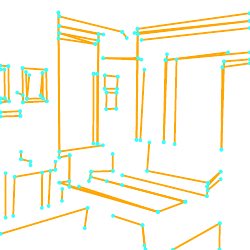} \\ \scriptsize{Ground truth} & \scriptsize{HT-CNN(100\%)} & \scriptsize{HT-LCNN(50\%)}  & \scriptsize{HT-LCNN(25\%)} &\scriptsize{HT-LCNN(10\%)}\\
        \includegraphics[width=0.19\textwidth]{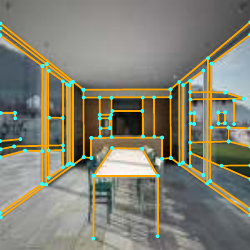} &
        \includegraphics[width=0.19\textwidth]{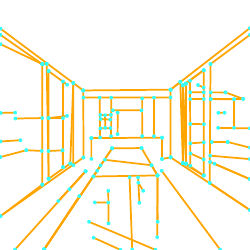} &
        \includegraphics[width=0.19\textwidth]{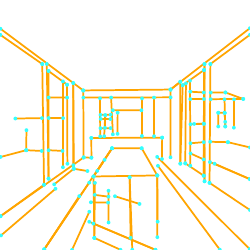} &
        \includegraphics[width=0.19\textwidth]{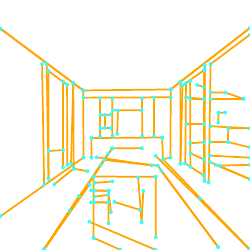} &
        \includegraphics[width=0.19\textwidth]{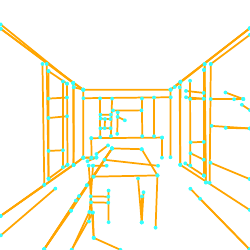} \\ 
        \scriptsize{Ground truth} & \scriptsize{LCNN(100\%)} & \scriptsize{LCNN(50\%)}  & \scriptsize{LCNN(25\%)} &\scriptsize{LCNN(10\%)}\\
        \includegraphics[width=0.19\textwidth]{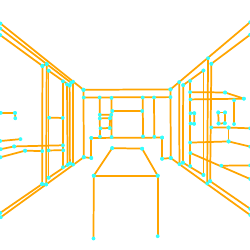} &
        \includegraphics[width=0.19\textwidth]{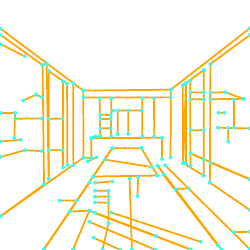} &
        \includegraphics[width=0.19\textwidth]{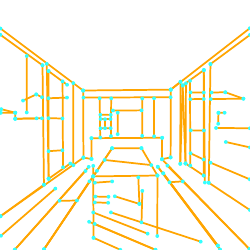} &
        \includegraphics[width=0.19\textwidth]{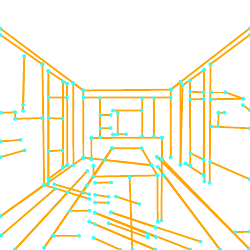} &
        \includegraphics[width=0.19\textwidth]{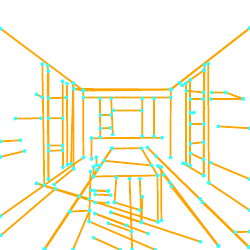}  \\ 
        \scriptsize{Ground truth} & \scriptsize{HT-LCNN(100\%)} & \scriptsize{HT-LCNN(50\%)}  & \scriptsize{HT-LCNN(25\%)} &\scriptsize{HT-LCNN(10\%)}\\          
    \end{tabular}
    \caption{\textbf{Exp 3.(a):} Visualization of detected wireframes from HT-LCNN (9.3M) and LCNN (9.7M) \cite{zhou2019end} trained on various Wireframe subsets \cite{huang2018learning}. Our HT-LCNN can more precisely detect the wireframes of the windows than LCNN, as shown in the first example. However, our HT-LCNN generates more false-positive predictions from the shadow of objects, when compared to LCNN, as shown in the last example.}
   \label{fig:sup_subsets_noimg}
\end{figure}

Figure~\ref{fig:sup_subsets_noimg} visualizes detected wireframes from our HT-LCNN (9.3M) and LCNN (9.7M) \cite{zhou2019end} trained on various Wireframe subsets \cite{huang2018learning}. We display the top 100 line segments. In the first example, our HT-LCNN is better than LCNN in detecting wireframes of windows on various subsets. However, our HT-LCNN is not able to ignore the shadow of objects, compared to LCNN, as shown in the last example. In general, HT-LCNN outperforms LCNN when training data is limited.

\section{\textbf{Exp 3.(b):} Qualitative comparison with the state-of-the-art on the Wireframe dataset}
\begin{figure}
    \centering
    \begin{tabular}{ccccc}
        \includegraphics[width=0.2\textwidth]{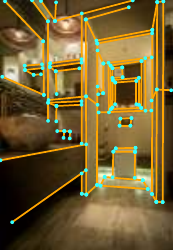} &
        \includegraphics[width=0.2\textwidth]{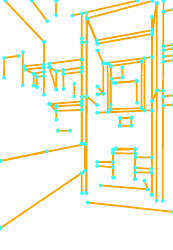} &
        \includegraphics[width=0.2\textwidth]{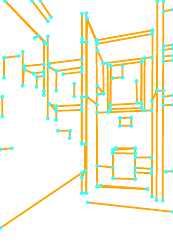} &
        \includegraphics[width=0.2\textwidth]{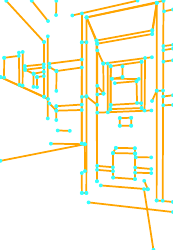} &
        \includegraphics[width=0.2\textwidth]{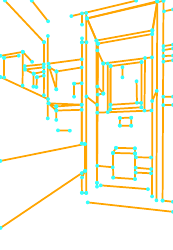} \\ 
        \scriptsize{Ground truth} & \scriptsize{HT-LCNN} & 
        \scriptsize{LCNN} & \scriptsize{HT-HAWP}& \scriptsize{HAWP}\\
        \includegraphics[width=0.2\textwidth]{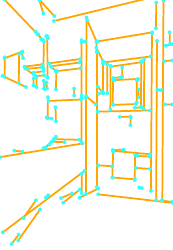}  &
        \includegraphics[width=0.2\textwidth]{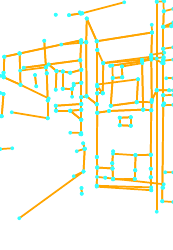}  &
        \includegraphics[width=0.2\textwidth]{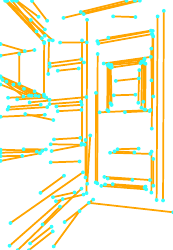} &
        \includegraphics[width=0.2\textwidth]{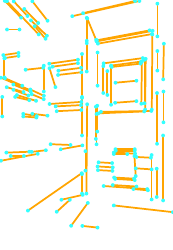} &
        \includegraphics[width=0.2\textwidth]{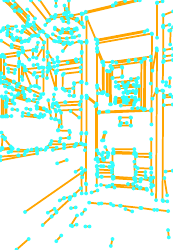} \\
        \scriptsize{AFM} &\scriptsize{WF-Parser} & \scriptsize{MCMLSD} & 
        \scriptsize{Linelet} & \scriptsize{LSD}\\
        \includegraphics[width=0.2\textwidth]{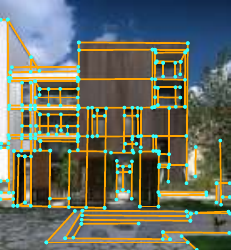} &
        \includegraphics[width=0.2\textwidth]{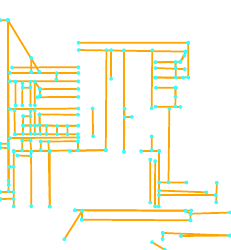} &
        \includegraphics[width=0.2\textwidth]{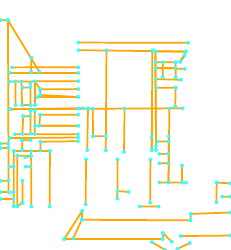} &
        \includegraphics[width=0.2\textwidth]{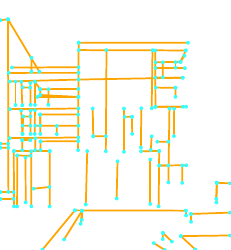} &
        \includegraphics[width=0.2\textwidth]{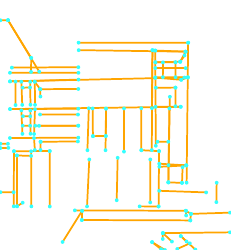} \\ 
        \scriptsize{Ground truth} & \scriptsize{HT-LCNN} & 
        \scriptsize{LCNN} & \scriptsize{HT-HAWP}& \scriptsize{HAWP}\\
        \includegraphics[width=0.2\textwidth]{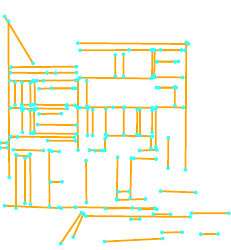} &
        \includegraphics[width=0.2\textwidth]{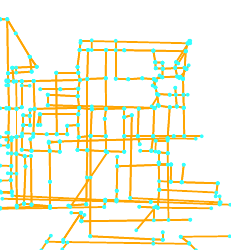}  &
        \includegraphics[width=0.2\textwidth]{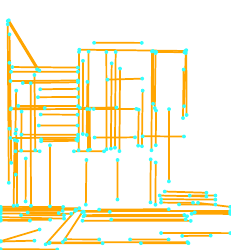} &
        \includegraphics[width=0.2\textwidth]{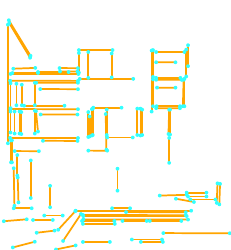} &
        \includegraphics[width=0.2\textwidth]{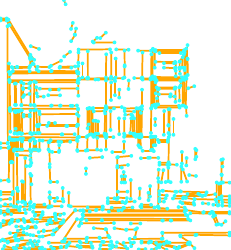} \\
        \scriptsize{AFM} &\scriptsize{WF-Parser} & \scriptsize{MCMLSD} & 
        \scriptsize{Linelet} & \scriptsize{LSD}\\
    \end{tabular}
    \caption{\textbf{Exp 3.(b):} Visualization of detected wireframes (line segments) on the Wireframe dataset \cite{huang2018learning}. We show predictions from our HT-HAWP, HT-LCNN  and seven other leading methods (HAWP\cite{xue2020holistically}, LCNN \cite{zhou2019end}, AFM \cite{xue2019learning}, WF-Parser \cite{huang2018learning}, MCMLSD \cite{almazan2017mcmlsd}, Linelet \cite{cho2017novel} and LSD \cite{von2008lsd}). 
    In general, learning-based methods are able to detect line segments more precisely, while MCMLSD, Linelet and LSD generate more false-positive predictions. The HT-LCNN and HT-HAWP predictions preserve both global structures and local details, and show competitive performance with the leading methods.
    }
    \label{fig:sup_all_noimg}
\end{figure}

Figure~\ref{fig:sup_all_noimg} visualizes detected line segments from different approaches on the Wireframe dataset  \cite{huang2018learning}. We follow \cite{xue2019learning} to set up thresholds for LSD \cite{von2008lsd} and WF-Parser \cite{huang2018learning}, and select the top 100 line segments for other methods (HT-HAWP, HT-LCNN, HAWP\cite{xue2020holistically}, LCNN \cite{zhou2019end}, AFM \cite{xue2019learning}, MCMLSD \cite{almazan2017mcmlsd} and Linelet \cite{cho2017novel}.)
Learning-based models predict line segments more precisely than the non-learning methods. In general, our models with \model perform competitively with the state-of-the-art.

\begin{figure}[t!]
    \centering
    \begin{tabular}{ccccc}
        \includegraphics[width=0.2\textwidth]{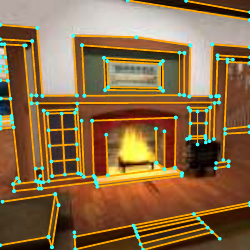} &
        \includegraphics[width=0.2\textwidth]{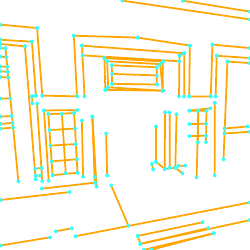} &
        \includegraphics[width=0.2\textwidth]{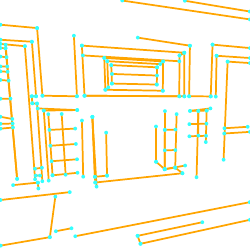} &
        \includegraphics[width=0.2\textwidth]{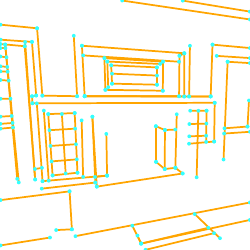} &
        \includegraphics[width=0.2\textwidth]{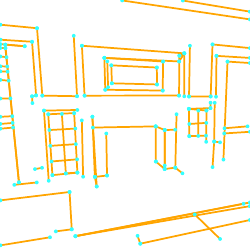}\\ 
        \scriptsize{Ground truth} & \scriptsize{HT-LCNN} & \scriptsize{LCNN} & \scriptsize{HT-HAWP} & \scriptsize{HAWP}\\
        \includegraphics[width=0.2\textwidth]{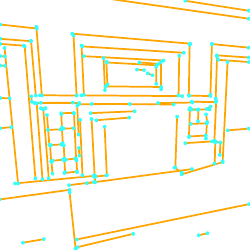}  &
        \includegraphics[width=0.2\textwidth]{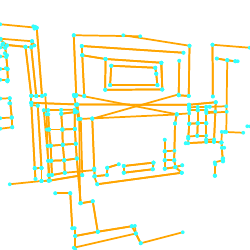}  &
        \includegraphics[width=0.2\textwidth]{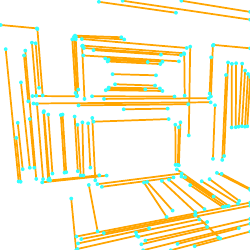} &
        \includegraphics[width=0.2\textwidth]{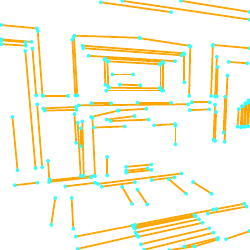} &
        \includegraphics[width=0.2\textwidth]{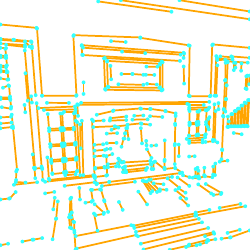} \\
        \scriptsize{AFM} &\scriptsize{WF-Parser} & \scriptsize{MCMLSD} & 
        \scriptsize{Linelet} & \scriptsize{LSD}\\
        \includegraphics[width=0.2\textwidth]{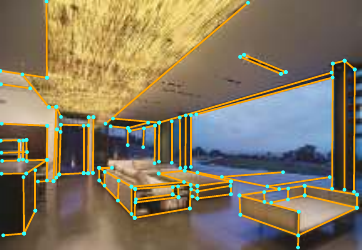} &
        \includegraphics[width=0.2\textwidth]{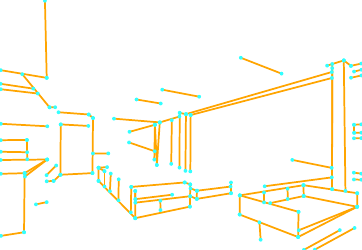} &
        \includegraphics[width=0.2\textwidth]{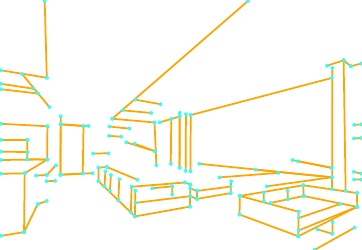} &
        \includegraphics[width=0.2\textwidth]{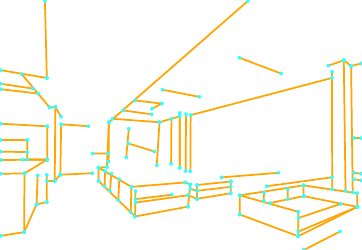} &
        \includegraphics[width=0.2\textwidth]{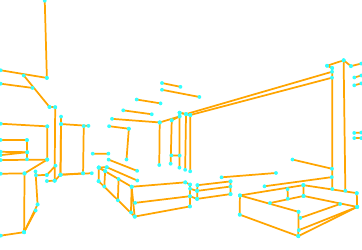} \\ 
        \scriptsize{Ground truth} & \scriptsize{HT-LCNN} & \scriptsize{LCNN} & \scriptsize{HT-HAWP} & \scriptsize{HAWP}\\
        \includegraphics[width=0.2\textwidth]{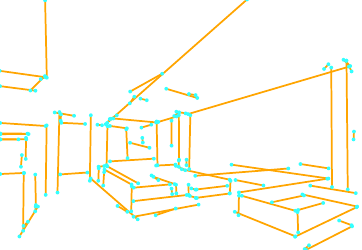} &
        \includegraphics[width=0.2\textwidth]{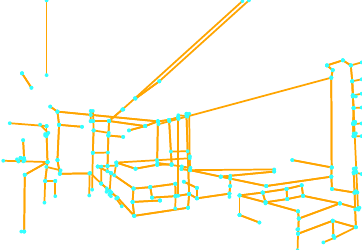}  &
        \includegraphics[width=0.2\textwidth]{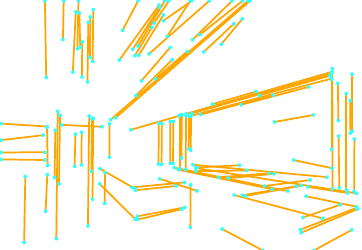} &
        \includegraphics[width=0.2\textwidth]{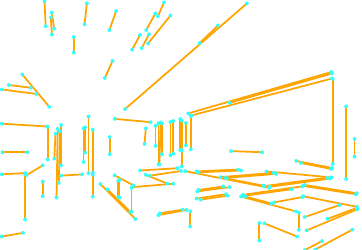} &
        \includegraphics[width=0.2\textwidth]{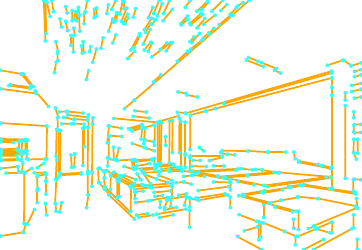} \\
        \scriptsize{AFM} &\scriptsize{WF-Parser} & \scriptsize{MCMLSD} & 
        \scriptsize{Linelet} & \scriptsize{LSD}\\
        
        \includegraphics[width=0.2\textwidth]{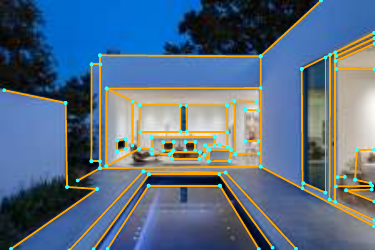} &
        \includegraphics[width=0.2\textwidth]{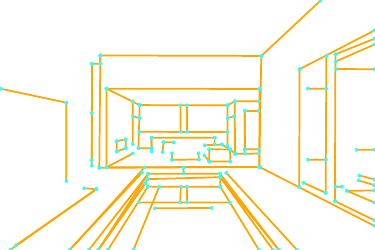} &
        \includegraphics[width=0.2\textwidth]{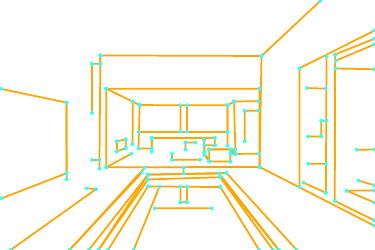} &
        \includegraphics[width=0.2\textwidth]{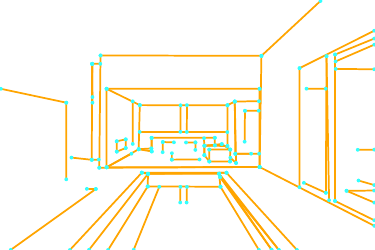} &
        \includegraphics[width=0.2\textwidth]{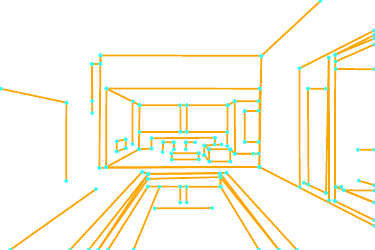} \\ 
        \scriptsize{Ground truth} & \scriptsize{HT-LCNN} & \scriptsize{LCNN} & \scriptsize{HT-HAWP} & \scriptsize{HAWP}\\
        \includegraphics[width=0.2\textwidth]{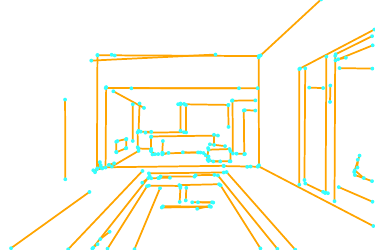}&
        \includegraphics[width=0.2\textwidth]{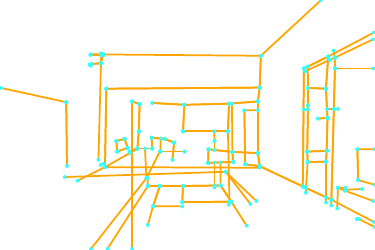}  &
        \includegraphics[width=0.2\textwidth]{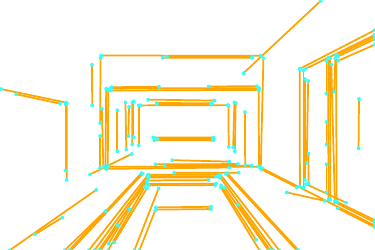} &
        \includegraphics[width=0.2\textwidth]{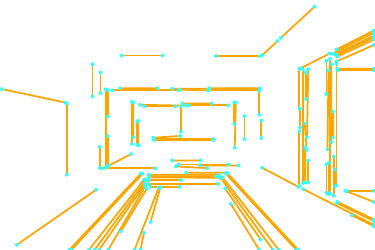} &
        \includegraphics[width=0.2\textwidth]{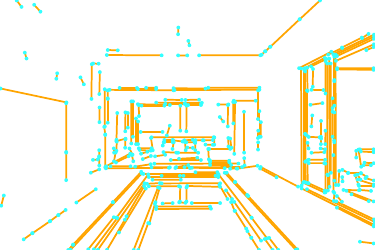} \\
         \scriptsize{AFM} &\scriptsize{WF-Parser} & \scriptsize{MCMLSD} & 
        \scriptsize{Linelet} & \scriptsize{LSD}\\
    \end{tabular}
    \caption{\textbf{Exp 3.(b):} Visualization of detected line segments on the Wireframe dataset \cite{huang2018learning}. We show predictions from our HT-HAWP, HT-LCNN, and seven other leading methods: HAWP\cite{xue2020holistically}, LCNN \cite{zhou2019end}, AFM \cite{xue2019learning}, WF-Parser \cite{huang2018learning}, MCMLSD \cite{almazan2017mcmlsd}, Linelet \cite{cho2017novel} and LSD \cite{von2008lsd}). 
    (Continued on the next page.)}
\end{figure}

